\documentclass{article}

\usepackage{PRIMEarxiv}
\usepackage{float} 

\usepackage[utf8]{inputenc} % allow utf-8 input
\usepackage[T1]{fontenc}    % use 8-bit T1 fonts
\usepackage{hyperref}       % hyperlinks
\usepackage{url}            % simple URL typesetting
\usepackage{booktabs}       % professional-quality tables
\usepackage{amsfonts}       % blackboard math symbols
\usepackage{nicefrac}       % compact symbols for 1/2, etc.
\usepackage{microtype}      % microtypography
\usepackage{lipsum}
\usepackage{graphicx}       % graphics
\usepackage{amsmath}       % graphics
\usepackage{listings}
\usepackage{xcolor}
\usepackage{fancyhdr}       % header
\usepackage{graphicx}       % graphics
\graphicspath{{media/}}     % organize your images and other figures under media/ folder
 
\usepackage{caption}
\usepackage{multirow}
\usepackage{soul}
\usepackage{tikz}
\usetikzlibrary{shapes.geometric, arrows, arrows.meta, positioning,fit,backgrounds}
 
\newcommand{\downarrowbetweenboxes}{
    \vspace{-0.4cm}
    \begin{center}
        \tikz \draw[-{Latex[length=3mm,width=2mm]}, thick] (0,-0) -- (0,-0.5);
    \end{center}
    \vspace{-.5cm}
}
 
\usetikzlibrary{decorations.pathreplacing}

\usepackage{sansmath} % To ensure math is also sans-serif inside flowchart

\tikzstyle{process} = [rectangle, rounded corners, minimum width=3cm, minimum height=1cm, text centered, draw=black, fill=orange!30]
\tikzstyle{start} = [ellipse, minimum width=3cm, minimum height=1cm, text centered, draw=black, fill=yellow!30]
\tikzstyle{decision} = [diamond, minimum width=3cm, minimum height=1cm, text centered, draw=black, fill=green!30]
\tikzstyle{arrow} = [thick,->,>=stealth]

\tikzstyle{phase} = [rectangle, rounded corners, minimum width=3cm, minimum height=1cm, text centered, draw=black, fill=orange!30]
\tikzstyle{result} = [ellipse, minimum width=3cm, minimum height=1cm, text centered, draw=black, fill=green!30]

\tikzstyle{dataset} = [rectangle, rounded corners, minimum width=2cm, minimum height=1cm, text centered, draw=black, fill=blue!30]
\tikzstyle{masking} = [rectangle, rounded corners, minimum width=2cm, minimum height=1cm, text centered, draw=black, fill=red!30]

\usepackage[most]{tcolorbox} % This loads most tcolorbox features, including advanced title positioning
\usepackage{fancyvrb}        % Enhances verbatim capabilities, which may be used within tcolorbox
\usepackage{xcolor}          % Essential for custom coloring and highlighting
\usepackage{soul}            % Provides highlighting commands like \hlred
\definecolor{aigold}{RGB}{244,210, 1} 
\definecolor{aigreen}{RGB}{210,244,211} 

\sethlcolor{aigreen}

\definecolor{aired}{RGB}{255,180,181}

\definecolor{aigold}{RGB}{255,180,181}

\definecolor{aiblue}{RGB}{173,216,230} % Light blue color (you can adjust the RGB values for a different shade)
 
\definecolor{lightred}{rgb}{1,0.9,0.9} % Light red color
\tcbset{
  customboxsmalll/.style={
    width=0.95\textwidth,  % 95% of page width
    fonttitle=\ttfamily\scriptsize, % Title font style
    colback=white,  % Box background color
    colframe=black,  % Box border color
    colbacktitle=blue!75!black, % Title background color
    coltitle=white,  % Title text color
    boxrule=0.5pt,  % Border thickness
    rounded corners,  % Rounded corners
    boxed title style={
        colback=blue!75!black, % Title background color
        colframe=black, % Title frame color
        boxrule=0.5pt  % Title border thickness
            before={\setlength{\parskip}{0pt} \setlength{\parindent}{-6pt}},  % Remove paragraph spacing
    after={\setlength{\parskip}{0pt} \setlength{\parindent}{-6pt}},   % Remove paragraph spacing
    halign=center,  % Center alignment
    before skip=-5pt,  % Space before the box
    after skip=-5pt,   % Space after the box
    innertop=-3pt,     % Inner top padding
    innerbottom=-3pt,  % Inner bottom padding
    valign=center,    % Center vertical alignment of content
    }
    
  }
}

\newtcolorbox{LLMboxSmall}[2][]{customboxsmalll, title=#2,#1}

\newtcolorbox{LLMbox}[2][]{custombox, title=#2,#1}

\newtcolorbox{OverallBox}[2][]{custombox,title=#2,#1}

\DefineVerbatimEnvironment{thinkingcolor}{Verbatim}{
  formatcom=\color{blue}, % Set color of the verbatim text
  fontsize=\scriptsize  % Optional: adjust font size if needed
}
\DefineVerbatimEnvironment{reflectcolor}{Verbatim}{
  formatcom=\color{orange}, % Set color of the verbatim text
  fontsize=\scriptsize  % Optional: adjust font size if needed
}

\newcommand{\hlred}[1]{{%
  \sethlcolor{lightred}% Set the highlight color to light red
  \hl{#1}% Highlight the argument text
}}

\tikzstyle{database} = [cylinder, cylinder uses custom fill, cylinder body fill=blue!20, cylinder end fill=blue!10, shape border rotate=90, aspect=0.5, draw]

\newcommand{\hllightgreen}[1]{{%
  \sethlcolor{aigreen}% Set the highlight color to light red
  \hl{#1}% Highlight the argument text
}}
\newcommand{\hlblue}[1]{{%
  \sethlcolor{aiblue}% Set the highlight color to light red
  \hl{#1}% Highlight the argument text
}}

\newcounter{textbox}
\renewcommand{\thetextbox}{\arabic{textbox}} % Adjust according to where you reset this

\newcommand{\textboxcaption}[1]{
  \refstepcounter{textbox}
  \noindent\small\textbf{Text Box \thetextbox:} #1\par\medskip
}

\tcbset{
  custombox/.style={
    width=512.18663pt,
    top=3pt,
    fonttitle=\bfseries\small\sffamily, % Bold, larger, and sans serif title font
    colback=white,
    colframe=black,
    colbacktitle=black,
    boxrule=0.5pt, % Border thickness
    colback=white, % Background color
    enhanced,
    rounded corners, % Rounded corners
    center,
    boxed title style={
            sharp corners,
            size=small,
            colframe=blue!75!black, % Title frame color
        },
    attach boxed title to top left={yshift=-0.1in,xshift=0.15in},
    boxed title style={boxrule=0pt,colframe=white,},
    before=\setlength{\baselineskip}{-2pt}, % Tighter line spacing
    before upper=\setlength{\baselineskip}{8pt}, % Tighter line spacing inside the box
    after=\setlength{\baselineskip}{8pt}, % Tighter line spacing after the box
  }
} 

\newtcbox{\mybox}[1][green]{on line,
arc=0pt,outer arc=0pt,colback=#1!10!white,colframe=#1!50!black,
boxsep=0pt,left=0pt,right=0pt,top=0pt,bottom=0pt,
boxrule=0pt,bottomrule=0pt,toprule=0pt}

\tcbset{
  customboxmultipage/.style={
    width=474.18663pt,
    top=3pt,%top=10pt,
    fonttitle=\bfseries\small\sffamily, % Bold, larger, and sans serif title font
    colback=white,
    breakable,
    colframe=black,
    colbacktitle=black,
    boxrule=0.5pt, % Border thickness
    colback=white, % Background color
    rounded corners, % Rounded corners
    center,
    boxed title style={
            sharp corners,
            size=small,
            colframe=blue!75!black, % Title frame color
        },
    attach boxed title to top left={yshift=-0.1in,xshift=0.15in},
    boxed title style={boxrule=0pt,colframe=white,},
  }
}
\newtcolorbox{LLMboxmultipage}[2][]{customboxmultipage,title=#2,#1}

\lstdefinelanguage{json}{
    basicstyle=\ttfamily\scriptsize, % Smaller font
    numbers=left,
    numberstyle=\tiny\color{gray}, % Line numbers color and style
    stepnumber=1,
    numbersep=10pt, % Increase separation between numbers and text
    showstringspaces=false,
    breaklines=true,
    frame=none,
    backgroundcolor=\color{white},
    literate=
     *{0}{{{\color{blue}0}}}{1}
      {1}{{{\color{blue}1}}}{1}
      {2}{{{\color{blue}2}}}{1}
      {3}{{{\color{blue}3}}}{1}
      {4}{{{\color{blue}4}}}{1}
      {5}{{{\color{blue}5}}}{1}
      {6}{{{\color{blue}6}}}{1}
      {7}{{{\color{blue}7}}}{1}
      {8}{{{\color{blue}8}}}{1}
      {9}{{{\color{blue}9}}}{1}
      {:}{{{\color{red}:}}}{1}
      {,}{{{\color{red},}}}{1}
      {\{}{{{\color{brown}\{}}}{1}
      {\}}{{{\color{brown}\}}}}{1}
      {[}{{{\color{brown}[}}}{1}
      {]}{{{\color{brown}]}}}{1},
}

\lstdefinelanguage{yaml}{
  basicstyle=\ttfamily\footnotesize, % Change \small to \footnotesize for a smaller font
  morekeywords={slices, sources, model, layer_range, merge_method, base_model, parameters, t, filter, value, dtype}, % list your keywords here
  keywordstyle=\color{blue},
  comment=[l]{\#},
  commentstyle=\color{gray},
  morestring=[b]',
  morestring=[b]",
  stringstyle=\color{orange},
  sensitive=true,
}

\title{PRefLexOR: Preference-based Recursive Language Modeling for Exploratory Optimization of Reasoning and Agentic Thinking
\thanks{\textit{\underline{Citation}}: 
\textbf{M.J. Buehler, et al., PRefLexOR: Preference-based Recursive Language Modeling for Exploratory Optimization of Reasoning and Agentic Thinking. Pages.... DOI:000000/11111.}} 
}

\author{  Markus J. Buehler \\
  Center for Computational Science and Engineering \\
  Schmarzman College of Computing \\  
  Laboratory for Atomistic and Molecular Mechanics (LAMM) \\
  Massachusetts Institute of Technology \\
  Cambridge, MA, USA\\ 
  \\
  \texttt{mbuehler@MIT.EDU} 
}

\begin{document}
\maketitle

%Preference-based Recursive Language Modeling Reasoning Optimization and Refinement
%Preference-based Refinement for Language Modeling Exploration and Optimization with Reasoning
%Preference-based Recursive Language Modeling for Exploratory Optimization of Reasoning
%Preference-based Recursive Language Modeling for Expansive Optimization of Reasoning

\begin{abstract}
\small
We introduce PRefLexOR (Preference-based Recursive Language Modeling for Exploratory Optimization of Reasoning), a framework that combines preference optimization with concepts from Reinforcement Learning (RL) to enable models to self-teach through iterative reasoning improvements, to create synthetic intelligence with enhanced scientific reasoning capabilities. Central to PRefLexOR is a recursive approach that engages the model in multi-step reasoning, revisiting, and refining intermediate steps before producing a final output in both training and inference phases. The foundation of PRefLexOR lies in multi-stage training, where the model first learns to align its reasoning with scientifically accurate decision paths by optimizing the log odds between preferred and non-preferred responses through a novel \textit{in-situ} dataset generation algorithm. For on-the-fly training data generation, PRefLexOR builds a dynamic knowledge graph by generating questions from random text chunks and utilizing retrieval-augmentation to contextualize relevant details from across the entire corpus, resulting in rigorous reasoning chains. In a second stage, preference optimization strategies further enhances model performance by using rejection sampling to fine-tune reasoning quality by continually producing \textit{in-situ} training data while masking the reasoning steps to focus on discovery of novel mechanisms to achieve correct answers. This hybrid approach mirrors key aspects of RL, where the model is continuously guided by feedback to improve decision-making and reasoning, and the adaptive process enables the model to self-teach as it continually improves through real-time feedback and recursive processing. Our method does not use pre-generated datasets and instead trains the model to continuously adapt and improve in real time. Recursive optimization within special thinking tokenization introduces iterative feedback loops, where the model refines its reasoning, much like policy refinement in RL, achieving deeper coherence, consistency, and adaptability. By recursively optimizing reasoning through feedback-driven learning, PRefLexOR achieves significant flexibility in its ability to handle complex tasks, learning and evolving its cognitive abilities autonomously. PRefLexOR’s recursive optimization mirrors how biological systems adapt and evolve. By using feedback loops to refine reasoning pathways during training and/or inference, it emulates nature’s resilience and adaptability, enhancing its decision-making capabilities. Implemented in very small language models with only 3 billion parameters, we showed that even tiny models can iteratively teach themselves to reason with greater depth and reflectivity, akin to an RL-based self-improving system capable of solving open-domain problems with superior reasoning depth and logic. Our implementation is straightforward and can be incorporated into any existing pretrained LLM. We focus our examples on applications in biological materials science, and demonstrate the method in a variety of case studies that range form in-domain to cross-domain applications. We explore several reasoning strategies that include both thinking and reflection modalities to construct a multi-agent recursive self-improving model that can successively improve responses via repeated sampling during inference, offering flexibility and integration into larger agentic systems.
\end{abstract}

% keywords can be removed
\keywords{Large language model \and Artificial intelligence \and Reinforcement Learning \and Materials science \and Reasoning}

\section{Introduction}

Generative artificial intelligence (AI) models, such as Large Language Models (LLMs) and many variants~\cite{Vaswani2017AttentionNeed,AlecRadfordImprovingPre-Training,Xue2021ByT5:Models,Jiang2023Mistral7Bb,Phi-2:Research,dubey2024llama3herdmodels} have not only impacted the landscape of natural language processing (NLP) but also unlocked the potential for scientifically-focused models that may ultimately be able to reason, think, and generate insight across an unparalleled range of disciplines. From general-purpose tasks to highly specialized domains like materials science and engineering~\cite{Buehler2023MeLMProblemsc, Singhal2023, Qu2024, Yu2024, Hu2023DeepScience, Buehler2024X-LoRA:Design,Buehler2023MechGPTModalities_updated, Luu2023BioinspiredLLM:Materials, lu2024finetuninglargelanguagemodels}, a grand challenge remains to develop strategies that yield more sophisticated scientific reasoning engines capable of performing tasks previously thought to be far beyond the reach of machines. 

Earlier work has resulted in attempts towards that goal, such as LLMs that were being taught to reason, not simply by brute force or through rote memorization, but by leveraging structured approaches that mimic human thought processes. Chain-of-thought prompting~\cite{wei2023chainofthoughtpromptingelicitsreasoning}, for instance, guides models to break complex problems into clear, manageable steps, mimicking the logical progression that human minds follow when faced with a challenging task. Similarly, few-shot learning methods~\cite{kojima2023largelanguagemodelszeroshot} give models the ability to handle new tasks with minimal examples, enabling them to generalize and adapt their reasoning capabilities to novel scenarios.

\begin{figure}
    \centering
    \includegraphics[width=1\linewidth]{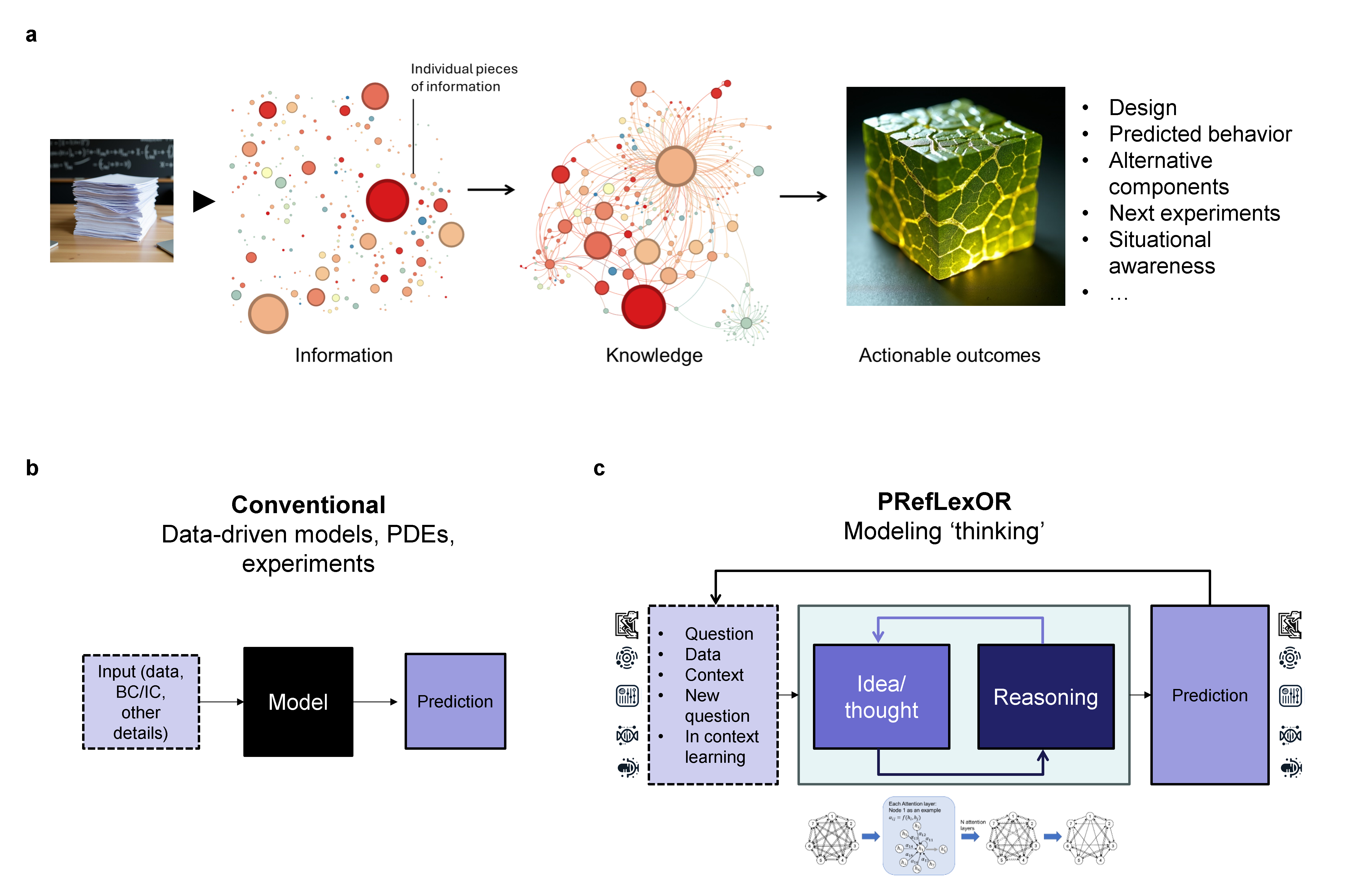}
    \caption{Illustration of the workflow and design principles behind generative materials informatics. Panel a: The process of transforming information into knowledge and actionable outcomes. Each individual piece of information (left) is synthesized into a network of interconnected knowledge, leading to informed decisions and innovative designs (right). Panel b: Conventional approaches in materials science rely on data-driven models, partial differential equations (PDEs), and experimental results, focusing on single-step predictions. Panel c: In contrast, generative materials informatics models built on the PRefLexOR framework proposed in this paper use ``thinking'' and ``reflection'' explicitly by incorporating iterative reasoning and contextual understanding, allowing for more complex, multi-step predictions. This approach expands from single inference steps, includes multiple modalities of data and responses, integrates real-world feedback and physics, and leverages self-assessment and self-learning. Using using reinforcement learning (RL) principles, the discovery of principles or the solution of specific tasks is further inspired by biological paradigms, using bio-inspired neural network designs. These advanced methods support continuous improvement in material predictions, enabling more adaptable and intelligent designs.}

    \label{fig:Fig_100}
\end{figure}

Yet, applying these powerful models in technical fields like biomateriomics~\cite{Cranford2010Materiomics:Macro,Cranford2012Biomateriomics} presents unique challenges. The intricacies of biomaterials design—where insights are drawn from multiscale, cross-disciplinary knowledge—require LLMs to go beyond surface-level understanding. In biomateriomics, researchers seek to explore and model biological systems at different scales, identifying how nature’s building blocks can inspire new materials~\cite{Cranford2012Biomateriomics,BuehlerMSME_BuildBlock_2023,Groen2011IntroducingMateriomics,Lee2022AnDiscovery,Arevalo2023LearningMaterials,Buehler2023MeLMProblemsc,ghafarollahi2024sciagentsautomatingscientificdiscovery}. Models of synthetic intelligence that capture scientific processes used in the analysis of such systems should offer a coherent and integrative strategy for solving cross-disciplinary problems, making them indispensable tools in fields like biomaterials research, where the ability to think, reason, and innovate is crucial. We posit that such advances can be achieved by developing models that can achieve several key objectives, including the ability to ingest rich, diverse and disparate information from varied sources by forming rigorous internal knowledge representations that can be used to predict actionable outcomes (Figure~\ref{fig:Fig_100}a). To reach this goal, models need to be developed that go beyond conventional predictions without situational awareness (Figure~\ref{fig:Fig_100}b) towards more sophisticated models that encompass a higher degree of situational awareness, realized through capabilities of self-reflection, error correction, and exploration of a wide space to predict novel solutions (Figure~\ref{fig:Fig_100}c). 

\subsection{Modeling reasoning, thinking and more}

As AI systems advance, the need for models capable of reasoning with greater depth, consistency, and adaptability has become increasingly critical. Traditional large language models (LLMs) have shown a certain level of proficiency in generating text, answering questions, and handling a wide range of natural language tasks. However, their reasoning capabilities--especially when it comes to reflecting on complex tasks or iterating over ideas to refine thought processes--remain limited, and are often only achieved in very large models.

In many current AI systems (especially in scientific applications), reasoning often follows a single-pass approach, where the model generates outputs without reflecting on the steps that led to its conclusions. This leads to challenges in solving open-domain or multi-step problems where deep cognitive engagement is required. Furthermore, the lack of flexibility in reasoning, adaptation to new challenges, and real-time learning means that these models struggle to handle tasks that require evolving, recursive reasoning strategies. 

To address these challenges, we propose PRefLexOR (Preference-based Recursive Optimization and Refinement), a framework that combines preference optimization with recursive reasoning inspired by Reinforcement Learning (RL) principles (Figure~\ref{fig:Fig_100}c). PRefLexOR enables models to self-teach by iterating over thought processes, refining reasoning, and continuously learning from both preferred and rejected outputs. This approach represents a shift towards a more reflective and flexible learning paradigm, where the model improves its decision-making in real time.

In PRefLexOR, the dynamic data generation process allows us to build a complex graph of interactions that facilitates recursive reasoning and refinement. For instance, when using a corpus of data sourced from scientific papers~\cite{Buehler2023MechGPTModalities_updated,Luu2023BioinspiredLLM:Materials,lu2024finetuninglargelanguagemodels}, the process begins by generating a question from a randomly selected piece of text, which acts as the initial node in the graph. To answer the question, we employ Retrieval-Augmented Generation (RAG), which queries the entire corpus, retrieving and integrating contextually relevant information from multiple sources.

This interaction between the question and the retrieved data forms a graph of knowledge, where nodes represent pieces of text, and edges represent the relationships between them. The embedding model plays a key role in this process by ensuring that similar pieces of information are mapped to adjacent nodes within the graph, facilitating efficient retrieval and reasoning. As the model continues to refine its reasoning across recursive cycles, this graph evolves, reflecting the complex interconnections between various pieces of knowledge and how they contribute to the model's final output.

\begin{figure}
    \centering
    \includegraphics[width=1\linewidth]{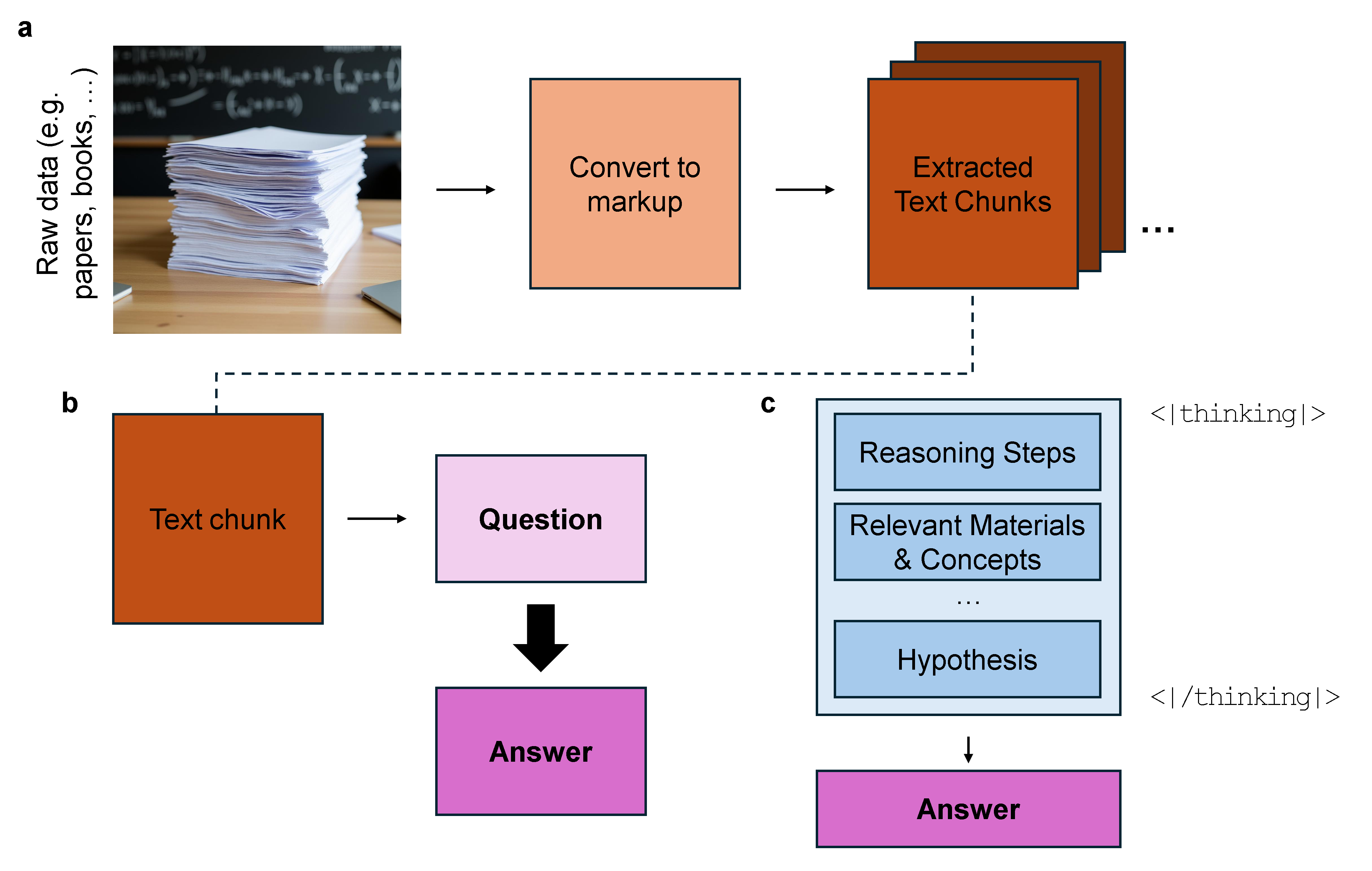}
    \caption{Strategic Dataset Generation Process with Structured Thought Integration.
    This figure illustrates a novel approach to generating datasets, where random text chunks are selected from raw data sources (e.g., papers, books, documents, notes, etc.) and used to develop question-answer pairs in a structured and strategic manner.
    Panel a: The process begins with raw data, such as research papers or books, which is converted into a markup format. This allows the data to be broken down into smaller, manageable text chunks. These chunks form the basis for generating questions in the subsequent steps.
    Panel b: A random selection of text chunks is used to generate question-answer pairs. This step involves creating a question from the text chunk and deriving an initial answer from the content. However, what distinguishes this approach is the next phase where a structured reasoning process is applied.
    Panel c: The system incorporates strategic reasoning and reflection, facilitated by the use of special thinking tokens (for instance: \texttt{<|thinking|>} and \texttt{<|/thinking|>}). Within this structured reasoning framework, the system iterates over several steps: Identifying relevant materials and concepts from the text, forming reasoning steps, and generating hypotheses. These processes are crucial to refining and validating the answer. Reflection, reasoning, and hypothesis generation are integrated to ensure that the answers are derived thoughtfully and are not merely surface-level extractions from the text. The thinking and reflection phases add depth to the question-answer generation, making the dataset richer and more valuable for subsequent learning tasks.
    }
    \label{fig:Fig_1}
\end{figure}

In this way, PRefLexOR constructs a dynamic, evolving knowledge graph that supports recursive reasoning, enabling the model to navigate, refine, and synthesize information across a vast corpus, improving the accuracy and coherence of its answers. Figure~\ref{fig:Fig_1} summarizes the process of strategic dataset generation with structured thought integration.

\subsection{Motivation and Challenges}
Traditional methods of training LLMs rely heavily on supervised fine-tuning, where models are trained on static datasets with fixed inputs and outputs. While this allows for the learning of broad patterns, it lacks the ability to dynamically adapt to new reasoning tasks. Furthermore, these models are limited in their capacity to engage in multi-step reasoning and reflection, often leading to outputs that lack coherence or depth when faced with complex, multi-faceted problems.

To overcome these limitations, recent advances have introduced preference optimization techniques, such as Odds Ratio Preference Optimization (ORPO)~\cite{hong2024orpomonolithicpreferenceoptimization} and Direct Preference Optimization (DPO)~\cite{rafailov2024directpreferenceoptimizationlanguage,ji2024efficientexactoptimizationlanguage_EXO}, or variants of these methods~\cite{saeidi2024insightsalignmentevaluatingdpo}. These methods guide the model to align its outputs with certain preferences (in the context of our particular application, scientific accuracy as identified using the raw corpus of data) by optimizing the log odds between preferred and rejected responses. However, existing implementations do not fully leverage the potential of recursive thinking and iterative refinement.

Additionally, the flexibility required to handle new tasks in real time, without relying on pre-constructed datasets, is often absent from these approaches. This necessitates the development of a model that can both learn autonomously and reflect on its own reasoning to improve continuously, a capability that can be framed in RL terms, where feedback loops and recursive processing drive learning improvements.

Recently proposed methods such as STaR and QuietSTaR frameworks~\cite{zelikman2022starbootstrappingreasoningreasoning,zelikman2024quietstarlanguagemodelsteach} introduce an innovative approach to enhancing the reasoning capabilities of language models through recursive thinking, reflection, and iterative refinement. Unlike traditional single-pass models that generate outputs in one step, Quiet-STaR emphasizes a multi-step process where models are encouraged to revisit, refine, and improve their reasoning before arriving at a final answer. This is achieved by integrating several key concepts that foster deeper cognitive engagement and reflection during the decision-making process. 
At the core of Quiet-STaR is the idea of recursive reasoning, where the model does not simply generate an output in a linear manner, but instead iteratively processes and refines its thoughts. This recursive process mirrors human thinking, where conclusions are often revisited, reassessed, and adjusted before a final decision is made. Quiet-STaR formalizes this by introducing intermediate steps that guide the model through this recursive process, allowing it to build upon its own reasoning in multiple stages.
In Quiet-STaR, the model engages in multi-step reasoning cycles, where each iteration produces a more refined version of the previous reasoning. These cycles enable the model to consider various aspects of a problem, explore different reasoning paths, and improve the coherence and depth of its output. This layered reasoning process leads to outputs that are more robust, structured, and better aligned with complex tasks that require detailed thought. 

Other methods, such as X-LoRA~\cite{Buehler2024X-LoRA:Design} have explored the use of `silent tokens' via the implementation of multiple forward passes, where training proceeded in two stages. First, the training focused on supervised fine-tuning that resulted in a set of distinct fine-tuned models, each realized via LoRA adapters and capable of solving particular tasks (e.g. protein property prediction, scientific methods, domain knowledge, etc.). Second, the X-LoRA model involves training of a additional layers in the model that utilize the first forward pass to create hidden states from which the relative contributions of all adapters, at every larger, is computed on a token-by-token level, forcing a state of self-reflection about its own configurational space. Because this strategy requires two forward passes for each token produced, the method utilizes silent thinking tokens that are used to configure itself for the actual prediction task. The  self-reflection tokens are never decoded, allowing this approach to invoke very rich contextual understanding during the thinking phase.  

A common theme behind these and related strategies is the use of increased compute during inference, to move away from autoregressive token predictions~\cite{Brown2020LanguageLearners} towards more sophisticated strategies where either more effort is spent per token, or where thinking and reflection strategies are employed that allow models to iterate through solutions and develop a higher level of self-awareness about their predictions. Many of the methods discussed above, however, require adaptation of new architectures and model structure changes.  As will be shown in this paper, we can utilize some of the ideas  by combining them with agentic modeling to create adversarial modeling strategies to ultimately arrive at  well-reasoned  responses to tasks (see, the flowchart in Figure~\ref{fig:Fig_100}).

\subsection{PRefLexOR Framework}
PRefLexOR addresses these challenges by integrating preference optimization with a recursive reasoning mechanism driven by thinking tokens, which explicitly mark phases of reasoning within the model’s output. This allows the model to:
\begin{enumerate}
    \item Generate initial reasoning steps.
    \item Revisit and refine those steps through recursive processing, ensuring that reasoning is consistent, coherent, and deeply aligned with scientifically accurate processes and resulting final answers.
    \item Adapt its decision-making by generating new tasks and feedback during training, enabling real-time learning.
\end{enumerate}
The algorithm features two major phases, complemented by agentic inference. We first focus on training strategies and move on to inference methods towards the end of the paper. The first phase is \textit{Structured Thought Integration Training}, followed by \textit{Independent Reasoning Development} and ultimately a \textit{Recursive Reasoning Algorithm}.

At the core of PRefLexOR’s approach is an initial alignment phase achieved using ORPO, which ensures that the model consistently aligns its reasoning with desired outcomes by directly optimizing preference odds. In a second phase, preference optimization strategies are then layered on to handle fine-tuning through rejection sampling, capturing more subtle distinctions in preference and further refining the model’s output. This layered approach, combined with recursive reasoning, makes the model capable of handling open-domain tasks with greater reasoning capacity and adaptability.

The recursive reasoning and iterative feedback loops in PRefLexOR closely resemble Reinforcement Learning (RL) methods, where models learn by refining policies based on rewards and feedback. In PRefLexOR, the model is continually provided with feedback in the form of preferred and rejected responses, which it uses to improve its thought process. This self-teaching mechanism is akin to the policy refinement seen in RL, where iterative feedback loops allow the model to explore, evaluate, and improve its decision-making in real-time.

The dynamic task generation in PRefLexOR introduces an active learning component, wherein the model generates tasks, reasoning steps, and negative examples on-the-fly during training. This method allows the model to handle more nuanced reasoning challenges, evolving its cognitive abilities without the need for extensive pre-curated datasets. The ability to recursively refine thoughts leads to a model that can continuously evolve and adapt to novel, complex problems, effectively teaching itself to reason more deeply and align its outputs with preferred outcomes that align with the ground truth data.

While Quiet-STaR focuses primarily on recursive reflection and iterative reasoning, it can be enhanced through preference optimization techniques like ORPO and preference optimization. By incorporating these techniques, the model can align its reflective reasoning with preferences rooted in training data, such as scientific papers or simulation results, ensuring that its refined thoughts and decisions meet desired outcomes. The recursive cycles in Quiet-STaR, for instance, can be viewed as a form of policy refinement in reinforcement learning, where the model’s reasoning policy is continually updated based on feedback. When combined with preference optimization, these cycles ensure that the model's internal reflections are not only coherent but also aligned with external preferences, further enhancing the model's performance in real-world tasks. 

Importantly, our method diverges from traditional approaches by not relying on pre-generated datasets; instead, it dynamically generates new tasks, reasoning steps, and feedback on the fly, allowing the model to continuously adapt and improve in real time and self-improve by comparing its own responses generated based on its current training state with ground truth answers extracted from the raw data using agentic prompting (details, see Materials and Methods).

\begin{figure}[ht]
\centering
\sffamily % Switch to sans-serif font
\sansmath % Ensure math in flowchart is also sans-serif
\begin{tikzpicture}[node distance=2.5cm and 2cm, every node/.style={align=center}]
    
    % Pretrained model
    \node (pretrained) [start] {Pretrained \\ Model};
    
    % Phase 1 - Structured Thought Integration
    \node (phase1) [phase, right of=pretrained, xshift=1.5cm] {Phase 1: Structured \\ Thought Integration \\ ORPO};
    \node (generateData1) [dataset, below of=phase1, yshift=0.5cm] {On-the-Fly Dataset \\ Generation for Phase 1};
    \node (rawdata1) [database, below of=generateData1, yshift=-4cm] {Corpus \\ of Raw Data};
    \node (image_papers) [left of=rawdata1, xshift=-.75cm, yshift=.35cm] {\includegraphics[width=3cm]{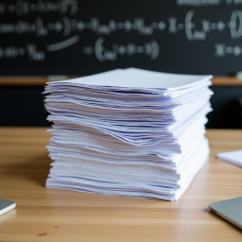}};

    % Phase 2 - Independent Reasoning Development
    \node (phase2) [phase, right of=phase1, xshift=2.cm] {Phase 2: Independent \\ Reasoning Development \\ EXO};
    \node (masking) [masking, below of=phase2,  yshift=0cm] {Apply Masking \\ to Thinking Tokens};
    \node (generateData2) [dataset, below of=masking, yshift=0cm, ] {On-the-Fly Dataset \\ Generation for Phase 2};
    
    % Final Aligned Model
    \node (result) [result, right of=phase2, xshift=2cm] {Aligned Model \\ with Reasoning \\ Capabilities};
    
    % Paths
    \draw [arrow] (pretrained) -- (phase1);
    \draw [arrow] (phase1) -- (phase2.west);
    \draw [arrow] (phase2) -- (result);
    
    % On-the-fly dataset generation input for Phase 1
    \draw [arrow] (generateData1) -- (phase1);
    
    % On-the-fly dataset generation and masking inputs for Phase 2
    \draw [arrow] (generateData2) -- (masking);
    \draw [arrow] (masking) -- (phase2);

    \draw [arrow] (rawdata1) -- (generateData1);
    \draw [arrow] (rawdata1) -- (generateData2);

\end{tikzpicture}
\sffamily % Restore default font
\caption{PreFLexOR: Model development and training strategy overview. The process starts with a pretrained model (here, \url{meta-llama/Llama-3.2-3B-Instruct}). Phase 1 focuses on structured thought integration, with on-the-fly dataset generation as input. Phase 2 develops independent reasoning capabilities by first generating a dataset, applying masking, and then proceeding with training. The final result is an aligned model with reasoning capabilities. }
\label{fig:PreFlexOR_training_strategy}
\end{figure}
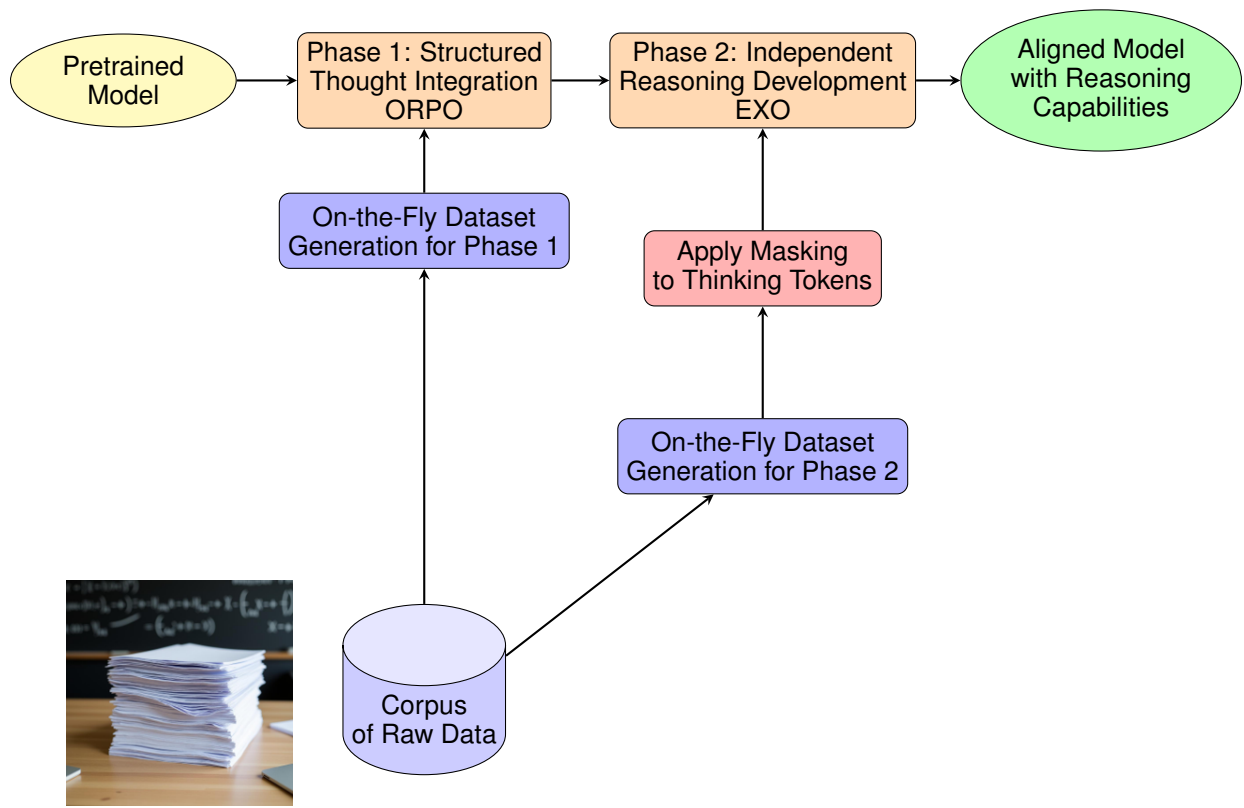

Figure~\ref{fig:PreFlexOR_training_strategy} shows an overview of the training strategy. Details of all aspects introduced therein will be covered in the remaining sections of the paper.

\subsection{Outline of this paper}
We first present the overall modeling strategy, focusing on the training phases and key aspects such as special tokens and other considerations. We present various inference examples with an in-depth technical analysis of the results. We then proceed to an experimental feature by incorporating multiple phases that feature both thinking and reflection. Using the reflection phase we implement a recursive algorithm that allows us to improve responses iteratively by scaling inference compute.  We conclude with a detailed discussion of strengths, weaknesses, and future results.

\section{Results and Discussion}

The training of the model consists of two distinct phases, each designed to progressively enhance its reasoning capabilities and ability to handle structured prompts and enhanced reasoning, here exemplified for domain-targeted structured thinking processes.

In the first phase, the model undergoes \textit{Structured Thought Integration Training}, where the primary focus is to teach the model how to handle new tokens specifically designed for reasoning, such as \texttt{<|thinking|>} and \texttt{<|/thinking|>}. This phase uses an algorithm that combines supervised fine-tuning and preference optimization to align the model’s outputs with high-quality responses that incorporate explicit reasoning steps, using ORPO. The objective here is twofold: 
\begin{itemize}
    \item To train the model to recognize and utilize structured prompts containing the new ``thinking" (and other) special tokens that delineate the reasoning process.
    \item To establish a preference framework that encourages the model to select and rank responses that demonstrate well-structured, step-by-step reasoning processes.
\end{itemize}
By the end of this phase, the model has learned how to generate outputs that adhere to explicit thought structures, preparing it to handle more complex tasks that involve reasoning elements.

The second phase, \textit{Independent Reasoning Development}, shifts the focus toward enabling the model to develop reasoning strategies autonomously. During this phase, tokens within the ``thinking" part of the training data are masked, which forces the model to reason independently without relying on explicit markers or structured prompts. The goal of this phase is to:
\begin{itemize}
    \item Encourage the model to generate coherent reasoning and decision-making strategies on its own, without explicitly being taught the thinking process, but rather to focus on the final correct answer.
    \item Enhance the model's ability to handle more challenging and ambiguous tasks by strengthening its internal reasoning mechanisms.
    \item Refine the model’s decision-making in cases where reasoning complexity increases, ensuring robustness even in extreme or difficult cases as the model sees new never-before-seen question-answer pairs with unknown reasoning steps (the model learns how to develop new reasoning strategies to arrive at the correct answers).
\end{itemize}
This phase not only improves the model's performance in handling reasoning tasks but also deepens its ability to make decisions without explicit guidance, preparing it to perform well in diverse, real-world scenarios.

We emphasize for all training phases, new question-answer data is generated on-the-fly by randomly selecting text chunks from the raw source data using the agentic framework to provide structured thinking mechanisms. 

\begin{figure}
    \centering
    \includegraphics[width=.66\linewidth]{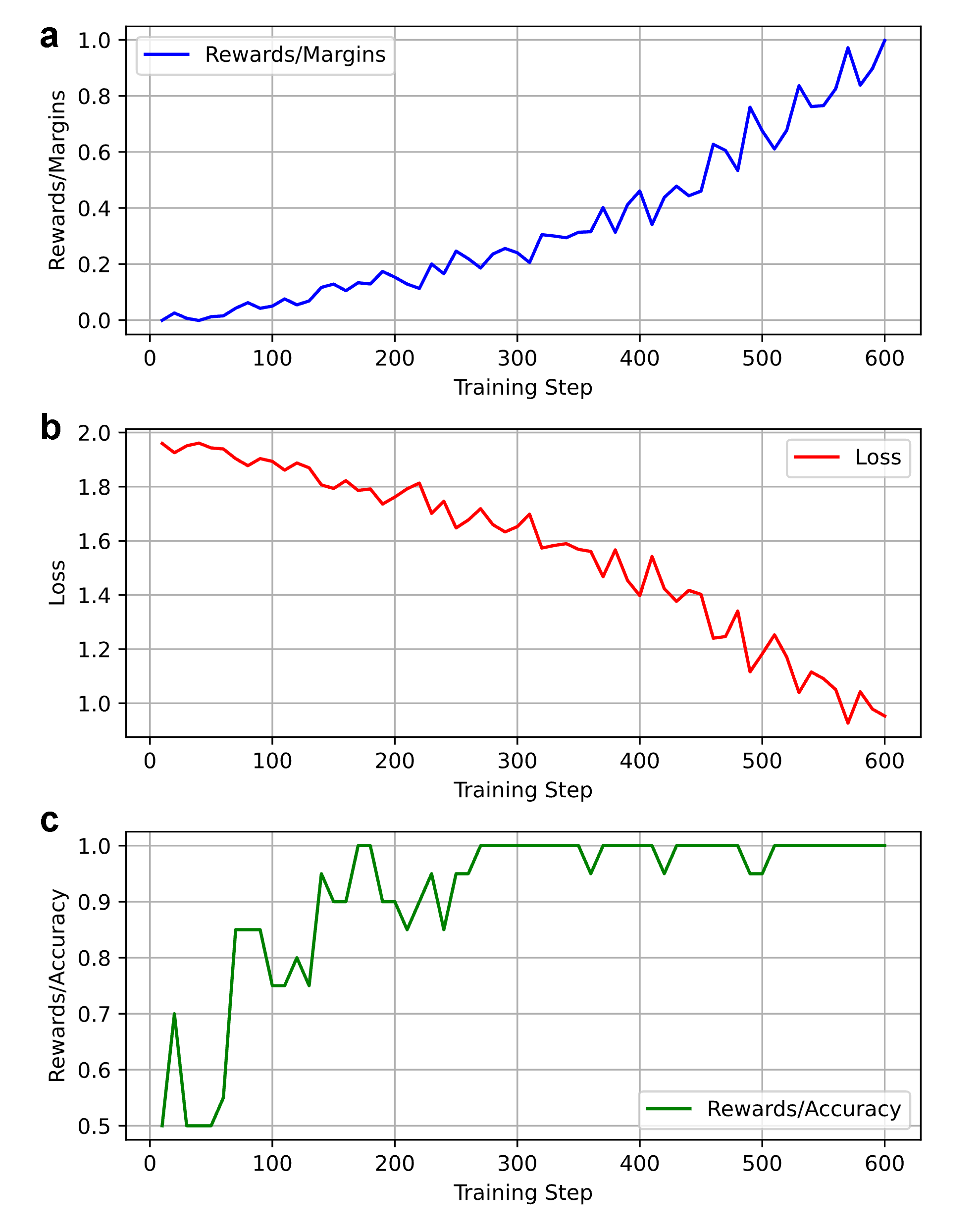}
    \caption{Training performance using the EXO method across three key metrics, during \textit{Independent Reasoning Development}. Panel a: The increase in rewards/margins over the course of training, indicating progressive improvement as the model learns. Panel b: The corresponding decrease in loss, showcasing successful convergence and optimization of the model, as reflected in a continuous decline in the loss function. Panel c: Rewards/accuracy during training, demonstrating rapid convergence toward high accuracy early in training, stabilizing after approximately 200 steps, with consistently high performance maintained throughout.  }
    \label{fig:Fig_5}
\end{figure}

We implemented an algorithm to generate domain-specific questions and their corresponding answers based on context retrieved from a pre-constructed index of data generated using \url{Llama-Index}~\cite{Run-llama/llama_index:Applications}. The algorithm extracts key information from the context, categorized into areas such as reasoning steps, relevant materials, and design principles (see Table~\ref{tab:categories}). This structured information is compiled into a ``Thinking Section for Reasoning'', which supports the generation of a well-reasoned, correct answer. Additionally, an incorrect answer is generated either by a trained model or via a prompt-based method, ensuring it lacks logical reasoning. The final output consists of the generated question, the correct answer with the Thinking Section, and the rejected answer, providing a robust framework for evaluating knowledge retention and reasoning skills. During preference-based alignment, we revise the generation process of the rejected answer by feeding it to the trained model in its current state to provide up-to-date answers. This challenges the model to develop improved reasoning strategies to obtain better answers by continually updating the rejected answers and thereby reducing the margin between chosen and rejected samples.  

The first phase of training uses Monolithic Odds Ratio Preference Optimization (ORPO)~\cite{hong2024orpomonolithicpreferenceoptimization}, a reference model-free method that simplifies preference alignment by leveraging the odds ratio to contrast favored and disfavored outputs. ORPO allows for effective supervised fine-tuning (SFT) with a minor penalty on disfavored outputs. This phase is used to teach the model the basic steps of thinking, including the introduction of the new thinking, reflection, and other, tokens. 

In the second phase, Efficient Exact Optimization (EXO)~\cite{ji2024efficientexactoptimizationlanguage_EXO} is applied to further refine the model's performance. EXO is a mode-seeking preference alignment approach that focuses on optimizing a model's final answers while masking intermediate reasoning (thinking tokens). Unlike Direct Preference Optimization (DPO), which uses forward KL divergence and can lead to diluted, mean-seeking behavior, EXO minimizes reverse KL divergence, allowing the model to concentrate on the most likely and effective answers. By aligning with the dominant modes in the preference data, EXO enables the model to infer the best reasoning patterns and produce more accurate final outputs, even when intermediate reasoning is hidden. This method results in better performance on tasks where final answer accuracy is prioritized. 

Once trained, the basic structure of the multi-stage process to generate the final answer is as follows:

\begin{LLMbox}{Basic structure of the reasoning strategy using a thinking phase before answering.}
\begin{lstlisting}
(*@\bf\hlblue{System:}@*) [System message]

(*@\bf\hlred{User:}@*) [User question or task]

(*@\bf\hllightgreen{Assistant:}@*) 
(*@\begin{thinkingcolor}
<|thinking|>
...
<|/thinking|>
(*@\end{thinkingcolor}
[Answer]
\end{lstlisting}
\end{LLMbox}

Further details are provided in the Materials and Methods section. 

\subsection{Sample Results}

We present a series of inference examples that cover a range of topics and tasks, from questions squarely in the training domain to questions at intersections to other areas, and tasks not included in the training data. These are meant to assess how well the model generalizes not only knowledge but specifically the reasoning steps, and whether it can translate its learned method of responding to tasks. The flexibility of hte prompting strategy allows us to trigger various phases of reasoning during inference. In a classic setting we simply provide the system message and user prompt, from which the model then completes the answer. In more nuanced approaches, we can provide the model with the system message, user prompt, and a draft thinking section. Variations of these can be used to scale inference compute, especially if we can dynamically adapt, improve and refine the thinking mechanics through recursive reasoning and reflection. 

\subsubsection{Properties of Hierarchical Structures}

In the first example we ask the model about why hierarchical structures work so well. No reference to ``materials'' is made to determine whether the model has been aligned with the domain of materials science. As shown in  Text Box~\ref{TexBox:thinking_and_reflection_Box_10}, in the ``thinking'' section, the model first dissects the concept of hierarchical structures by identifying multiple advantages, such as energy dissipation, size adaptation, and material utilization. It demonstrates a sophisticated understanding of how hierarchical structures work across different scales, benefiting mechanical strength, thermal insulation, and impact resistance. The reasoning steps not only delve into the abstract principles behind these structures but also connect them to specific material properties and their applications, notably using examples like nacre.

This process shows that the model can move from theoretical concepts to practical insights, offering clear explanations of how hierarchical designs optimize performance in various domains. The ability to articulate both general principles and specific examples (e.g., aragonite tablets in nacre) shows a nuanced grasp of material science. By synthesizing this information into a structured, coherent final answer, the assistant demonstrates advanced reasoning skills, capable of understanding complex systems and distilling that understanding into a succinct yet thorough explanation.

\subsubsection{Biological Materials Failure Mechanism}

In Text Box~\ref{TexBox:thinking_and_reflection_Box_11} we show an example where the user asks about how biological materials fail gracefully. The ``thinking'' section elaborates on the underlying mechanisms by which biological materials fail gracefully. It highlights concepts such as viscoelasticity, crack bridging, and fiber sliding, among others, which contribute to energy dissipation and gradual failure rather than sudden collapse. The thinking section also points out the hierarchical structure of biological materials and its role in stress redistribution and energy absorption. This part serves as a deep, reflective reasoning process aimed at comprehensively outlining the failure mechanisms of biological materials.

After the ``thinking'' section, the model synthesizes the reasoning into a coherent answer that integrates the identified mechanisms—such as viscoelasticity, crack propagation, and fiber pullout—into a concise explanation. Interestingly, the answer directly builds on the detailed exploration in the thinking section but presents it in a more refined and structured manner. The hierarchical structure is again emphasized as a key feature, and practical examples (e.g., bone and nacre) are introduced, demonstrating how the abstract reasoning leads to a well-grounded and practical explanation. 

The assistant not only identifies key mechanisms of graceful failure in biological materials but also provides a well-organized, step-by-step breakdown of these complex processes. This reflects a deep conceptual grasp of material science principles, such as viscoelasticity, crack propagation, and hierarchical structures. The assistant shows the ability to connect abstract concepts with practical examples (e.g., bone and nacre), further enhancing the clarity of the explanation. The structured reflection in the ``thinking'' section suggests an advanced reasoning capability. The model's ability to analyze the topic comprehensively and then distill the information into a succinct, coherent answer demonstrates intelligence comparable to that of a subject-matter expert. This combination of deep theoretical knowledge and the skill to communicate it effectively highlights a sophisticated level of cognitive processing, which goes beyond mere fact retrieval to active synthesis and application of knowledge.

\subsubsection{Intersection between Literature, Philosophy and Materials Science}

In Text Box~\ref{TexBox:thinking_and_reflection_Box_12}, the user asks a challenging and interdisciplinary question, requesting an explanation of the conceptual connections between Hermann Hesse's Glass Bead Game (German: Das Glasperlenspiel)~\cite{hesse2000glass} and proteins. This query was chosen as an example of a task that had not been included in the training set, to see how well the model can generalize its reasoning capabilities to areas outside of materials science. We find that the model's response in the  ``thinking'' section demonstrates a high level of reasoning and synthesis, drawing parallels between a work of philosophical fiction and the scientific realm of proteins. The response explores both concepts in depth, highlighting key themes such as structural complexity, hierarchical organization, dynamic nature, interconnectedness, and evolutionary adaptation, which are common to both Hesse’s Glass Bead Game and biological proteins.

The reasoning steps begin by reviewing the main ideas in Hesse's Glass Bead Game, a metaphor for the interconnectedness of knowledge across art, science, and philosophy. The assistant then draws an analogy with proteins, which are structurally and functionally complex biological molecules. Proteins, like Hesse’s symbolic glass beads, operate on multiple hierarchical levels (primary, secondary, tertiary, and quaternary), and the assistant recognizes that this hierarchical structure is a key feature in both domains. For example, proteins rely on molecular interactions and bonding to form stable structures, much like how Hesse's game symbolizes the interaction between intellectual domains.

The dynamic nature of both systems is another strong connection. Proteins, with their ability to change conformation and function depending on their environment, reflect the flexibility and adaptability of Hesse's philosophical game. Proteins are not static entities but evolve and adapt, much like how knowledge and ideas in The Glass Bead Game evolve over time.

The assistant also highlights interconnectedness as a major theme, noting that proteins play roles in various biological processes through interactions with other proteins and molecules, which mirrors how Hesse’s game illustrates the interdependence of art, science, and philosophy. This interconnectedness is a crucial insight, showing how the assistant is able to bridge these abstract and scientific concepts.

The hypothesis put forth \textit{Proteins, with their intricate structures and hierarchical organization, and their dynamic nature, are analogous to the interconnected, hierarchical, and dynamic elements of Herrman Hesse's Glass Bead Game}, reflects advanced synthesis. The model positions proteins and the game as analogous systems that, despite being from entirely different domains (biology and philosophy), share deep structural and functional parallels. This hypothesis is a strong foundation for the analysis, suggesting that the underlying principles governing biological materials can also apply to conceptual frameworks in philosophical thought.

The ability to develop such a hypothesis demonstrates sophisticated interdisciplinary thinking, requiring the assistant to understand not only Hesse’s complex metaphysical ideas but also the scientific details of protein function. It combines abstract thinking with scientific rigor, drawing out the universal patterns that apply across both domains. By framing proteins as analogous to Hesse's symbolic representation of reality, the assistant adds depth to the interpretation of both concepts, demonstrating a holistic understanding of interconnected systems.

The level of discourse required to connect The Glass Bead Game and materials science is remarkably high. Hermann Hesse’s novel is deeply philosophical, exploring the abstract and often metaphysical connections between human intellectual pursuits. On the other hand, proteins, as biological macromolecules, are grounded in the concrete, molecular world of biology and material science. Successfully combining these two requires a profound understanding of both philosophical and scientific concepts.

The assistant’s ability to merge these distinct disciplines involves cognitive flexibility and the capacity to think abstractly about systems. This level of reasoning is typically seen in advanced interdisciplinary studies, where individuals are not confined to a single field but draw on multiple domains of knowledge to generate novel insights. The fact that the assistant effectively bridges philosophical narrative with molecular biology demonstrates the application of high-order thinking skills such as synthesis, analogy, and abstraction.

The parallels between proteins and The Glass Bead Game involve not only shared structural elements (e.g., hierarchical complexity) but also shared functional characteristics (e.g., adaptability, interconnectedness), which are universal principles that transcend disciplines. This demonstrates the assistant’s ability to identify and articulate abstract patterns that apply across seemingly unrelated fields, a hallmark of advanced intellectual discourse.

This inference example showcases an in-depth analysis and synthesis of two very different domains: Hermann Hesse's Glass Bead Game and the science of proteins. The assistant draws on the structural, hierarchical, and dynamic aspects of both, weaving them together into a coherent and insightful hypothesis. The level of discourse reflects advanced interdisciplinary thinking, requiring both abstract philosophical interpretation and detailed scientific knowledge. This analysis highlights the universality of certain principles, such as complexity, interconnectedness, and adaptation, that apply across both biological and philosophical systems. 

The result is particularly remarkable given that the base model is a tiny LLM with only around 3 billion parameters; yet, our algorithm endows it with superior reasoning capabilities.

\subsubsection{Analysis of Research Abstract and Proposal of new Hypotheses}

In this task, the user presents an abstract from a a recently published paper that was not included in the training data~\cite{D3MH01277H} focused on the development of a novel platform for manufacturing structural myco-composites. The model is asked to summarize the results and propose future research directions, prompting the use of a structured reasoning process. The response in the ``thinking'' section showcases a comprehensive analysis and well-developed research proposal, reflecting an intelligent, high-level discourse typical in materials science and biocomposites research. Notably, the model has succesfully applied its reasoning strategy to this new task. 

During the reasoning phase, the model starts by summarizing the core findings of the abstract, focusing on key innovations such as high-resolution biocomposite additive manufacturing, robust mycelium colonization, and the scalability and tunability of the resulting myco-composites. By highlighting the mechanical improvements—namely, a 15-fold increase in strength and modulus—it captures the most significant results of the study. The assistant emphasizes the hierarchical composite design and selective nutritional provision as central principles that contribute to the improved mechanical and surface properties. Additionally, it notes the versatility of the platform, demonstrated through applications like foldable bio-welded containers and flexible mycelium textiles, illustrating the study's practical implications.

The model then proposes a well-framed hypothesis: \textit{The novel platform for manufacturing structural myco-composites, leveraging high-resolution biocomposite additive manufacturing and robust mycelium colonization, can create scalable, tunable, and complex-geometry compatible myco-composites with superior mechanical and surface properties.} This hypothesis effectively captures the innovative aspects of the research and highlights the key attributes of the platform—scalability, tunability, and mechanical superiority. It reflects a clear understanding of the study’s goals and the broader implications for biocomposite and hybrid-living materials research.

The assistant then goes beyond summarization by offering eight well-articulated proposals for future research, each building on the foundation laid by the original study. These include:
\begin{enumerate}
    \item Scaling Up and Down: Investigating the scalability of the manufacturing process to produce both larger and smaller structures, a logical next step for practical applications.
    \item Material Properties Enhancement: Exploring ways to further improve the composite's properties, potentially by altering the colonization process or integrating new materials, which reflects a forward-looking approach to optimizing performance.
    \item Multifunctional Composites: Proposing research into creating composites with integrated functionalities, such as self-healing or conductivity, a suggestion that opens the door to entirely new applications.
    \item Biodegradability and Sustainability: Addressing the environmental aspect of the materials, which aligns with the growing focus on sustainable material science.
    \item Hybrid-Living Materials: Continuing the integration of living organisms with synthetic materials, advancing the frontier of hybrid-living material research.
    \item Complex Geometry and Topology: Exploring the mechanical effects of more intricate geometries, further leveraging the platform’s compatibility with complex structures.
    \item Inoculation Strategies: Optimizing the colonization process by experimenting with different strains or nutrients, which could lead to further improvements in material performance.
    \item Biocomposite Additive Manufacturing: Developing new manufacturing techniques to improve the resolution and speed of production, which is essential for industrial-scale adoption.
\end{enumerate}

The level of discourse required to effectively summarize and propose research directions based on this abstract is advanced, both in terms of scientific understanding and strategic vision. The assistant demonstrates a solid grasp of materials science principles, particularly regarding the use of hierarchical composite design, additive manufacturing, and biocomposites. The proposed research ideas reflect a deep understanding of the field’s current state and potential future trajectories, showing intellectual maturity and the ability to think creatively about how the field can progress.

This task demands interdisciplinary knowledge, as it touches on materials science, biology, engineering, and sustainability. The assistant successfully bridges these areas, synthesizing the information in a coherent manner and proposing forward-thinking, practical research directions. The result is a well-rounded, intelligent response that addresses both the technical details of the research and broader implications for the field.

The assistant's analysis of the paper’s abstract showcases an intelligent approach to summarization and research proposal development. The reasoning steps clearly capture the essential findings of the study, and the proposed hypothesis is well-aligned with the research goals. Furthermore, the future research directions demonstrate a high level of strategic thinking, covering a range of potential innovations, from scaling and material optimization to multifunctional and biodegradable composites. Overall, the assistant’s response reflects a high level of expertise, interdisciplinary knowledge, and creative thinking in the realm of structural myco-composites and hybrid-living materials.

\subsubsection{Overall Analysis of Inference Examples}

In these examples, the model displays a sophisticated level of interdisciplinary reasoning, effectively synthesizing knowledge from areas such as philosophy, materials science, biology, and literary analysis. The model’s ability to handle abstract concepts like Hermann Hesse’s Glass Bead Game alongside detailed scientific inquiries into protein structures and myco-composites highlights its versatility. A notable strength of the model’s performance is its capacity to make high-level connections between seemingly disparate domains, such as drawing parallels between the hierarchical structure of proteins and the interconnected elements of Hesse's philosophical game. This ability to fluidly shift between abstract and applied reasoning showcases a nuanced understanding of both conceptual and practical frameworks.

In its analysis of the myco-composites abstract, the model excels at summarizing the study’s core findings—highlighting innovations like high-resolution biocomposite additive manufacturing and robust mycelium colonization. Its prediction also captures specific technical details of importance of achieving a modulus of 160 MPa and tensile strength of 0.72 MPa, emphasizing the significance of a 15-fold improvement in material properties. Furthermore, the model’s forward-thinking research proposals demonstrate not only a grasp of current scientific advancements but also the potential for future innovation, such as exploring multifunctional composites with integrated self-healing or optical properties, or addressing sustainability through biodegradable materials.

What stands out is its ability to extrapolate insightful research directions from initial findings. For instance, the suggestions to further optimize inoculation strategies for mycelium colonization or to investigate complex geometry impacts on material properties display an understanding of cutting-edge scientific trends. Similarly, when connecting The Glass Bead Game to protein structures, the model provides a compelling well-reasoned hypothesis that demonstrates the structural and dynamic analogies between the two fields, underscoring a deep conceptual link between philosophy and biology.

The implementation of training strategies inspired by reinforcement learning methods, where thinking tokens are masked but the model is trained on the final answer, is integral to these high-level insights. This training method emphasizes clarity and precision in the final response, ensuring that the model can produce coherent, well-reasoned conclusions without relying on explicit intermediate reasoning steps. By focusing on outcome-driven learning, the model is able to internalize the reasoning process and deliver sophisticated answers efficiently. This approach is particularly evident in the model’s ability to articulate complex research proposals and cross-disciplinary analogies with minimal overt reasoning, yet delivering accurate and innovative insights. The particular training approach enhances the model’s ability to present answers that are both contextually rich and technically sound, resulting in a streamlined yet insightful final output.

\subsection{Expanding the Analysis to incorporate Thinking and Reflection for Recursive Improvement in Agentic Modeling}

\begin{figure}
    \centering
    \includegraphics[width=.45\linewidth]{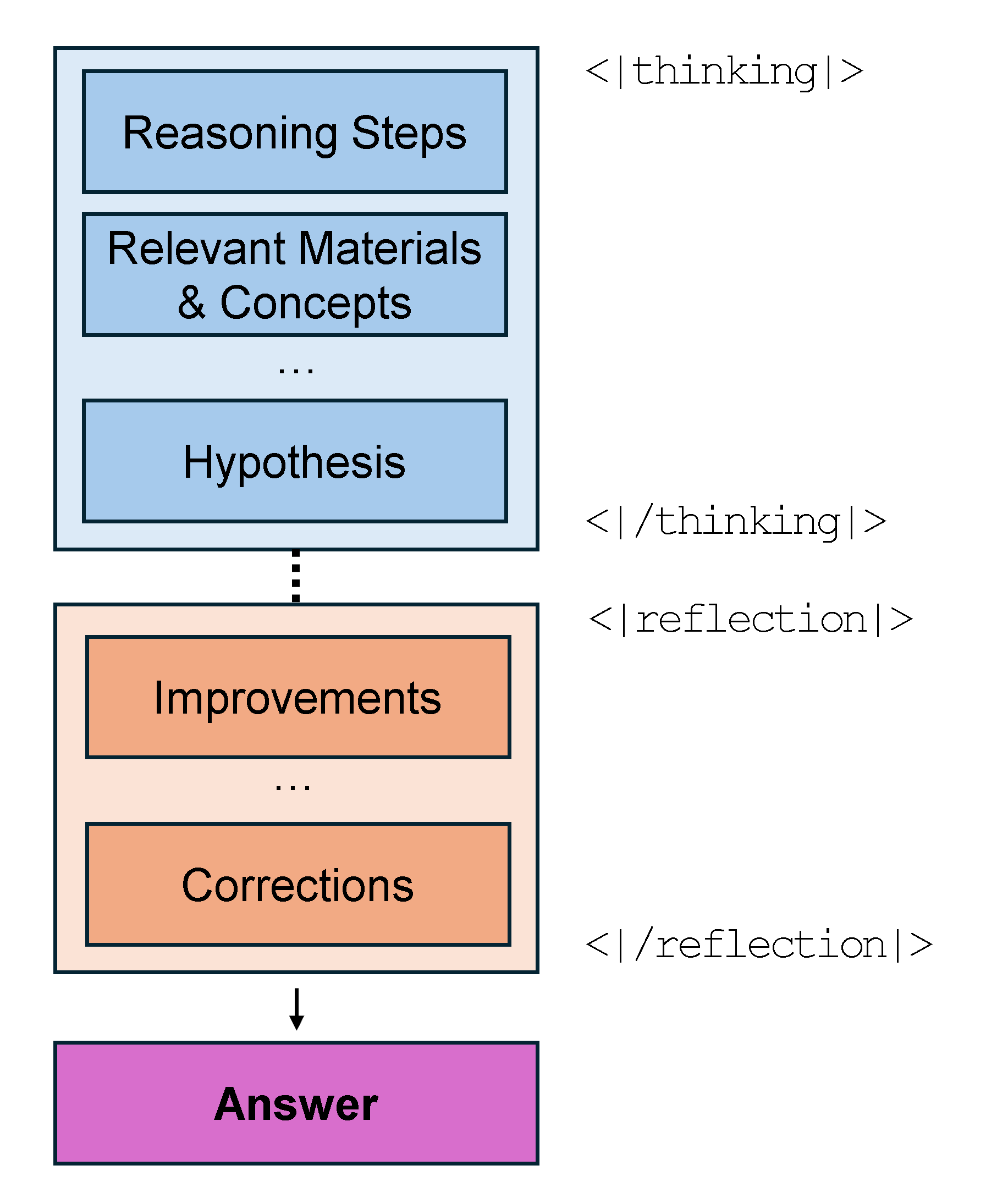}
    \caption{Structured Thought and Reflection in Answer Generation.
This diagram illustrates the multi-step process of answer generation, incorporating both structured thinking and reflection phases to ensure thoroughness and accuracy. 
As in the original approach, the process begins with the \texttt{<|thinking|>} phase, where key reasoning steps are identified. This phase involves the following steps: 
(i) outlining the reasoning steps based on the available data, 
(ii) referencing relevant materials and concepts that support the reasoning process, 
(iii) forming hypotheses to guide the conclusion.  After the initial thinking process, the system moves to the \texttt{<|reflection|>} phase, where the generated answer is refined. During this phase, improvements and corrections are made to ensure that the final output is accurate and relevant. The combination of these two phases—structured thinking and reflection—results in a robust and refined final answer, which is shown at the bottom of the diagram.}
    \label{fig:Fig_10}
\end{figure}

To show the flexibility of the method, we experiment also with other reasoning mechanisms such as combining \texttt{<|thinking|>}, \texttt{<|/thinking|>} with a second stage of reflection, triggered by \texttt{<|reflection|>} and \texttt{<|/reflection|>}. In this phase the model reviews earlier responses and is encouraged to critique, improve and otherwise enhance the responses before the final answer is produced. Figure~\ref{fig:Fig_10} depicts an overview of this approach.

The basic structure of this multi-stage process is as follows:

\begin{LLMbox}{Basic structure of the reasoning strategy with thinking and reflection tokens.}
\begin{lstlisting}
(*@\bf\hlblue{System:}@*) [System message]

(*@\bf\hlred{User:}@*) [User question or task]

(*@\bf\hllightgreen{Assistant:}@*) 
(*@\begin{thinkingcolor}
<|thinking|>
...
<|/thinking|>
(*@\end{thinkingcolor}
\begin{reflectcolor}
<|reflect|>
...
<|/reflect|>
(*@\end{reflectcolor}
[Answer]
\end{lstlisting}
\end{LLMbox}

Text Box~\ref{TexBox:thinking_and_reflection_Box_14} shows a sample conversation answering the question ~\texttt{Tell me why hierarchical structures work so well.}. The model follows a two-step process involving \textit{thinking} and \textit{reflection} to infer and refine the final answer. During the \textit{thinking phase}, the model generates reasoning steps related to hierarchical structures, focusing on concepts such as mechanical properties, anisotropic behavior, and functional adaptation. This phase is driven by inference, where the model explores possible answers through logical reasoning and relevant details.

The second step, \textit{reflection}, serves to refine the initial ideas. In this phase, the model critically evaluates its reasoning and proposes specific improvements, such as clarifying the role of hierarchical organization and recognizing the interplay of multiple factors like structure and composition. This reflection process helps the model fine-tune its inferences, leading to a more accurate and complete final answer.

By separating the process into thinking and reflection, the model ensures that its inference mechanism is both exploratory and self-correcting. The result is a well-balanced answer that combines reasoning with critical evaluation.

In the thinking phase, the model generates the following reasoning steps:

\begin{itemize}
    \item Mechanical Properties: Hierarchical structures exhibit unique properties across different length scales, enabling material flexibility and strength.
    \item Material Organization: These properties result from organized changes in material composition at various scales.
    \item Anisotropic Nature: These structures behave anisotropically, adapting their mechanical properties based on directional requirements.
    \item Functional Adaptation: This anisotropic behavior allows materials to efficiently perform different functions.
\end{itemize}

In the reflection phase, the model revisits its previous reasoning and suggests the following improvements:

\begin{itemize}
    \item Clarify Hierarchical Organization: Emphasize that the properties of hierarchical structures result from changes in material properties and structure at different scales.
    \item Interplay of Factors: Recognize that hierarchical structures are influenced by a combination of factors—structure, composition, and architecture—rather than a single dominant feature.
    \item Cost Considerations: Introduce the complexity and cost associated with designing and manufacturing hierarchical structures.
    \item Anisotropic Nature: Clarify that anisotropic behavior is a product of organized material changes, not an independent principle.
\end{itemize}

This structured inference process allows the model to generate reasoning, reflect on its accuracy, and refine the answer. The \textit{thinking phase} handles the exploration of possible answers, while the \textit{reflection phase} corrects and refines them, resulting in a more informed and optimized final answer.

\subsubsection{Recursive Reasoning Algorithm}

The existence of reflection allows us to implement a Recursive Reasoning Algorithm, a method designed to enhance the quality and depth of responses generated by the reasoning model by iteratively improving its reasoning steps in the thinking phase based on the reflection feedback.  

The algorithm utilizes a multi-agent format, and exploits a synergistic interaction between two distinct models: The fine-tuned reasoning model and a general-purpose critic model. The reasoning model, specialized through careful fine-tuning, excels in generating structured, logical responses to given prompts. It not only produces initial responses but also demonstrates the capability to iteratively improve its outputs based on feedback. Complementing this, the critic model serves as an evaluator and improved, analyzing the reasoning model's outputs and improving it based on the feedback received. 

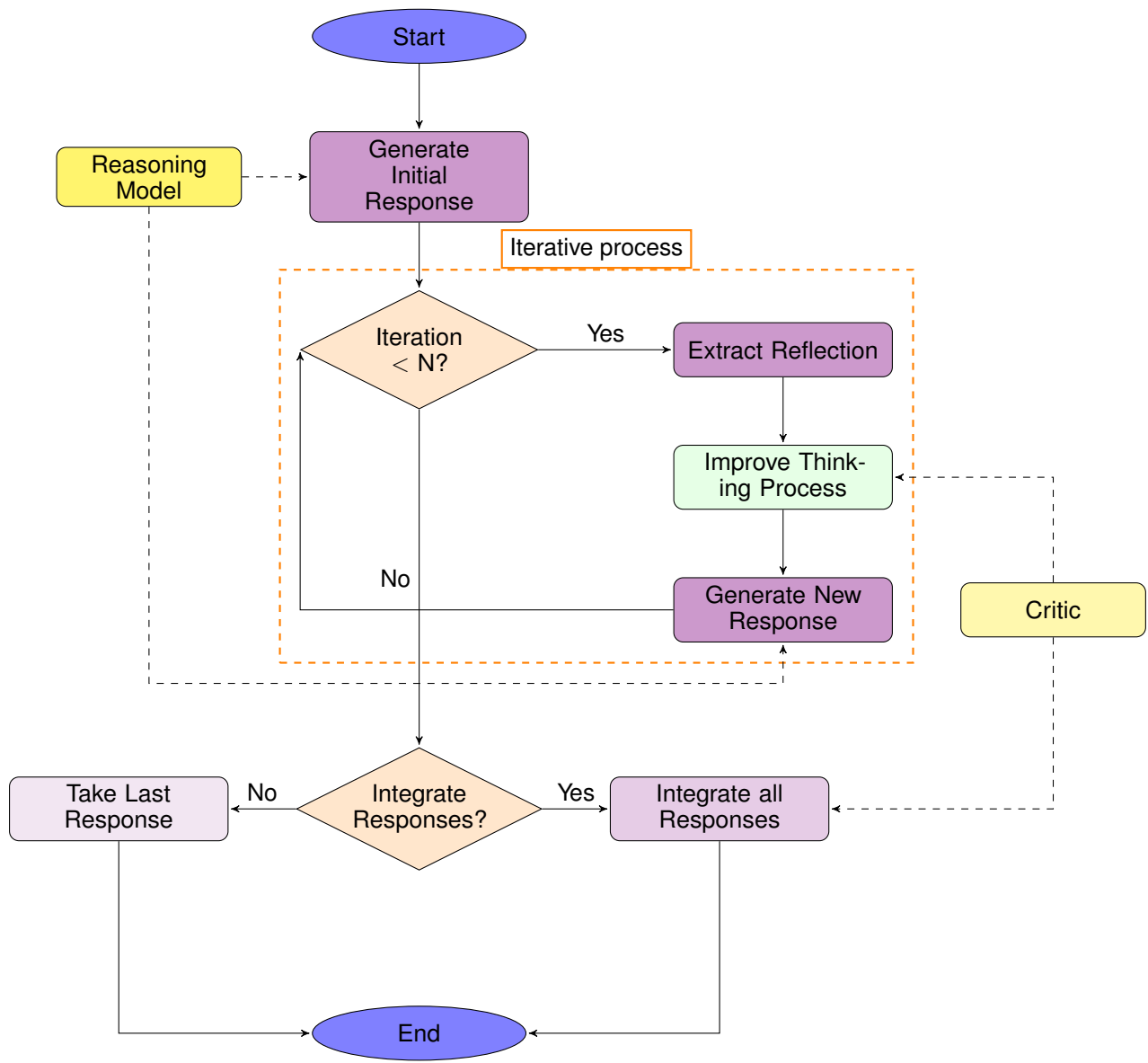
\begin{figure}
\sffamily % Switch to sans-serif font
\sansmath % Ensure math in flowchart is also sans-serif

\begin{tikzpicture}[node distance=1cm and 2cm, auto,
    block/.style={rectangle, draw, fill=violet!40, text width=3cm, text centered, rounded corners, minimum height=0.8cm},
    block2/.style={rectangle, draw, fill=violet!20, text width=3cm, text centered, rounded corners, minimum height=0.8cm},
    block3/.style={rectangle, draw, fill=violet!10, text width=3cm, text centered, rounded corners, minimum height=0.8cm},
    block4/.style={rectangle, draw, fill=green!10, text width=3cm, text centered, rounded corners, minimum height=0.8cm},
    decision/.style={diamond, draw, fill=orange!20, text width=2cm, text centered, inner sep=1pt, aspect=2},
    cloud/.style={draw, ellipse, fill=blue!50, text width=2cm, text centered, minimum height=0.8cm},
    model/.style={rectangle, draw, fill=yellow!40, text width=2.5cm, text centered, rounded corners, minimum height=0.8cm},
    model2/.style={rectangle, draw, fill=yellow!70, text width=2.5cm, text centered, rounded corners, minimum height=0.8cm},
    line/.style={draw, -stealth', shorten >=1pt},
    dashed line/.style={draw, dashed, -stealth', shorten >=1pt}]
    % Define nodes
    \node [cloud] (start) {Start};
    \node [block, below=of start] (init) {Generate \\ Initial \\ Response};
    \node [decision, below=of init] (iterate) {Iteration \\ $<$ N?};
    \node [block, right=of iterate] (extract) {Extract Reflection};
    \node [block4, below=of extract] (improve) {Improve Thinking Process};
    \node [block, below=of improve] (generate) {Generate New Response};
    \node [decision, below=of iterate, yshift=-4cm] (final_decision) {Integrate \\ Responses?};
    \node [block3, left=of final_decision, xshift=1cm] (last_response) {Take Last Response};
    \node [block2, right=of final_decision, xshift=-1cm] (integrate) {Integrate all Responses};
    \node [cloud, below=of final_decision, yshift=-1cm] (end) {End};
    % Model nodes
    \node [model2, left=of init, xshift=1cm] (reasoning) {Reasoning Model};
    \node [model, right=of generate, xshift=-1cm] (critic) {Critic};
    % Connect nodes
    \path [line] (start) -- (init);
    \path [line] (init) -- (iterate);
    \path [line] (iterate) -- node [above] {Yes} (extract);
    \path [line] (extract) -- (improve);
    \path [line] (improve) -- (generate);
    \path [line] (generate) -| (iterate.west);
    \path [line] (iterate) -- node [left] {No} (final_decision);
    \path [line] (final_decision) -- node [above] {No} (last_response);
    \path [line] (final_decision) -- node [above] {Yes} (integrate);
    \path [line] (last_response) |- (end);
    \path [line] (integrate) |- (end);
    % Connect models
    \path [dashed line] (reasoning) -- (init);
    \path [dashed line] (reasoning) |- ([yshift=-7.5cm]reasoning -| generate.south) -- (generate.south);
    \path [dashed line] (critic) |- (improve.east);
    \path [dashed line] (critic) |- (integrate);
    % Add a frame around the iteration process
    \begin{pgfonlayer}{background}
        \node [draw=orange, thick, dashed, fit=(iterate) (extract) (improve) (generate), inner sep=0.3cm, 
               label={[fill=white, draw=orange, thick]above:Iterative process}] {};
    \end{pgfonlayer}
\end{tikzpicture}
 \caption{PRefLexOR Recursive Reasoning Algorithm: An iterative approach leveraging a fine-tuned Reasoning Model and a general-purpose Critic Model to generate, refine, and optionally integrate responses. The process involves generating initial responses, extracting reflections, improving thinking processes, and creating new responses based on refined thinking, with an optional final integration step. The algorithm relies on extracting thinking processes (indicated via \texttt{<|thinking|>}..\texttt{<|/thinking|>}) and reflection processes  (indicated via \texttt{<|reflect|>}..\texttt{<|/reflect|>}). The use of special tokens allows us to easily construct such agentic modeling as it facilitates pausing inference, improving the strategy, and re-generating improved answers. The sampled responses can either be used in their final state or integrated into an amalgamated response that shows very rich facets in the scientific process.  
}
\label{fig:PreFlexOR_recursive}
\sffamily % Restore default font
\end{figure}

At the heart of the algorithm lies an iterative process shown in Figure~\ref{fig:PreFlexOR_recursive}, forming the core mechanism for continuous improvement of responses. This process begins with the Reasoning Model generating an initial response to a given prompt. Subsequently, this response undergoes a cycle of refinement. Each iteration involves a thorough analysis of the current response, from which a reflection is extracted (indicated via \texttt{<|reflect|>}..\texttt{<|/reflect|>}). The critic model then utilizes this reflection to suggest improvements to the thinking process indicated via \texttt{<|thinking|>}..\texttt{<|/thinking|>}. Based on these suggestions, the model generates a new, improved response, incorporating the insights gained from the critic's analysis. This is done by feeding the improved thinking process to the model, which then uses that to generate a new reflection mechanism (to be used in the next iteration, if another one is done) and then the next answer. 

This cycle of generation, evaluation, and refinement continues for a predetermined number of iterations or until the algorithm achieves a response that meets specified quality criteria. The iterative nature of this process allows for the progressive enhancement of responses, with each cycle building upon the improvements of the previous one.

Upon completion of the iterative process, the algorithm presents two options for final output selection. The first option is to select the response from the final iteration as the definitive output. Alternatively, the algorithm offers the capability to integrate all generated responses into a comprehensive final answer, potentially capturing a broader range of insights and perspectives developed throughout the iterative process.

This approach combines the structured thinking of the specialized reasoning model with the broader perspective of the critic model. The result is a system capable of producing responses that are not only logically sound but also nuanced and comprehensive. By incorporating elements of self-reflection and iterative improvement, the recursive response algorithm strives to emulate human-like reasoning and problem-solving processes, potentially leading to higher-quality outputs in various natural language processing tasks. 

\begin{figure}
    \centering
    \includegraphics[width=.8\linewidth]{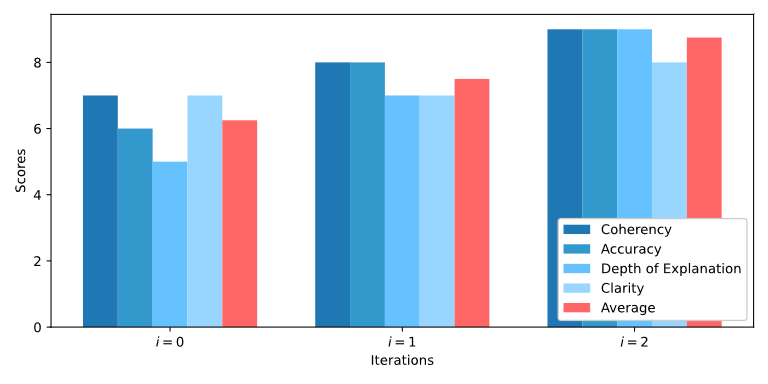}
   \caption{Scores of model responses to the question ``How do biological materials fail gracefully'' across three iterations (\textit{i=0}, \textit{i=1}, and \textit{i=2}). Each bar represents the score for one of the evaluated criteria: Coherency, accuracy, depth of explanation, and clarity, with the fifth bar showing the average score for each iteration. The final iteration (\textit{i=2}) exhibits the highest overall performance, reflecting improvements in the depth and technical accuracy of the explanation. The color scheme differentiates individual criteria (in shades of blue) from the average score (in red).}

    \label{fig:Fig_11}
\end{figure}

We show an example result based on this algorithm in Text Box~\ref{TexBox:thinking_and_reflection_Box_30}.  Table~\ref{tab:biological-materials-comparison} shows an analysis of the text produced over three iterations, clearly showing how the responses are improved successively. Figure~\ref{fig:Fig_11} depicts a quantitative analysis of the writing quality over the iterations, conducted using \url{gpt-4o} (details on prompting, see Materials and Methods). The three iterative responses were analyzed based on four key criteria: coherency, accuracy, depth of explanation, and clarity. The first version (\textit{i=0}) presented a concise overview, introducing the concepts of hierarchical structures and periodic hierarchies but lacking specific mechanisms and supporting details. This iteration scored lower in depth of explanation (5/10) and accuracy (6/10), as it provided only a high-level summary of biological material failure. The second iteration (\textit{i=1}) improved by introducing specific mechanisms such as brittle fracture, sacrificial bonds, and helical fibers, contributing to a better understanding of energy dissipation mechanisms. However, some redundancy in the structure affected coherency (8/10), and transitions between ideas could be smoother. The final version (\textit{i=2}) provided the most detailed and accurate response, offering a comprehensive explanation of hierarchical structures, periodic hierarchies, and material properties such as nacre’s strength-to-weight ratio. This version also addressed the role of loading conditions and how they influence energy dissipation at various scales, scoring highest in accuracy (9/10) and depth (9/10). We find that overall, \textit{i=2} demonstrated the clearest and most coherent explanation, making it the strongest iteration of the three, with an average score of 8.75/10.

The final answer correctly identifies that biological materials fail gracefully due to a combination of hierarchical structures and periodic hierarchies that operate at multiple scales, from the molecular to the macroscopic level. These hierarchical structures help redistribute stress and dissipate energy, while periodic hierarchies—repeating patterns at various scales—enhance toughness and prevent sudden failure. Mechanisms such as helical fibers, sacrificial bonds, and mineral bridges contribute to energy dissipation, making the material more resistant to catastrophic failure. The specific effects of these features can vary depending on loading conditions, such as impact, tensile, or compressive stress, allowing biological materials to maintain functionality under diverse mechanical demands. 

We note that further research should be done to examine this mechanism and perhaps including iterative refinement in the reinforcement training process. There are many important directions to be explored, such as how to best train for improved thinking and reflection processes using masking, and/or which particular reinforcement learning approach may work best.

\begin{table}[h]
\small
\centering
\begin{tabular}{|p{7cm}|c|c|c|}
\hline
\textbf{Feature} & \textbf{$i$=0} & \textbf{$i$=1} & \textbf{$i$=2} \\
\hline
Basic concept explanation & \checkmark & \checkmark & \checkmark \\
\hline
Hierarchical structures mentioned & \checkmark & \checkmark & \checkmark \\
\hline
Periodic hierarchies mentioned & \checkmark & \checkmark & \checkmark \\
\hline
Detailed explanation of structures & $\times$ & \checkmark & \checkmark$+$ \\
\hline
Energy dissipation mechanisms & $\times$ & \checkmark & \checkmark \\
\hline
Specific examples (e.g., nacre) & $\times$ & \checkmark & \checkmark$+$ \\
\hline
Quantitative information & $\times$ & $\times$ & \checkmark \\
\hline
Discussion of loading conditions & $\times$ & $\times$ & \checkmark \\
\hline
Well-structured response & $\times$ & \checkmark & \checkmark$+$ \\
\hline
Comprehensive summary & $\times$ & $\times$ & \checkmark \\
\hline
\end{tabular}
\vspace{0.25cm}
\caption{Comparison of responses for the question around biological failure mechanisms, as shown in Text Box~\ref{TexBox:thinking_and_reflection_Box_30}.  This table compares three responses obtained via iterative  using the algorithm depicted in Figure~\ref{fig:PreFlexOR_recursive}, for three steps ($i$=0, $i$=1, $i$=2), explaining how biological materials fail gracefully. A checkmark (\checkmark) indicates the presence of a feature, a cross ($\times$) indicates its absence, and a checkmark with a plus (\checkmark$+$) indicates the feature is present and more extensively covered. 
Response i=2 is the most comprehensive, covering all aspects with greater depth and including additional 
elements like quantitative information and loading condition effects.}
\label{tab:biological-materials-comparison}
\end{table}

\subsubsection{Comparison with Non-fine-tuned Model}

When inferencing the questions as covered above using the non-fine tuned model, we find that the responses are not aligned to the application domain (here: biological materials) and do not feature thinking sections. Text Box~\ref{TexBox:thinking_and_reflection_Box_40_refmodel} shows an example that illustrates the different, non-domain specific and more generic response without any thinking and reflection section, which we contrast to the results shown in Text Box~\ref{TexBox:thinking_and_reflection_Box_14}. 
 
The two responses analyzed provide different but complementary perspectives on the effectiveness of hierarchical structures. The response by the non-fine-tuned model takes a broad, organizational view, focusing on the practical benefits of hierarchies in fields like business and government. It highlights key principles such as clear lines of authority, specialization, and accountability. The response emphasizes how hierarchies facilitate efficient communication, structured decision-making, and scalability, making them highly effective in large, complex organizations. Additionally, it addresses the role of motivation and incentives, noting that hierarchical structures provide clear career paths and promote accountability, which in turn drive productivity and organizational success. However, while Response 1 offers a comprehensive overview of the managerial benefits of hierarchies, it lacks a deeper exploration of the fundamental principles that extend to other fields, such as material science or biology.

In contrast, the response from the reasoning model presents a more specialized, technical analysis, focusing on hierarchical structures within the context of materials science, particularly biological materials. This response delves into the multi-scale organization of hierarchical systems, explaining how they enable the efficient absorption and distribution of energy, often through anisotropic behavior. It highlights the superior mechanical properties that arise from hierarchical designs, such as enhanced strength, toughness, and adaptability. A key concept introduced is progressive damage, a mechanism that allows hierarchical materials to fail gracefully rather than catastrophically, contributing to their durability in both natural and engineered systems. This response provides a detailed explanation of the structural advantages of hierarchical designs, particularly their ability to maintain functionality under stress through organized changes in material properties at different length scales.

The ``thinking'' and ``reflection'' components, prominent in the response from the reasoning model, are missing from the non-fine-tuned model. The inclusion of a structured ``thinking'' phase allows for a detailed breakdown of the reasoning behind hierarchical structures, focusing on mechanical properties, material organization, and functional adaptation. This explicit reasoning process helps build a logical argument, grounding the response in scientific principles. Furthermore, as discussed above, the ``reflection'' phase offers an opportunity to refine the explanation by addressing potential improvements, clarifying assumptions, and considering broader implications, such as the costs or complexities of hierarchical systems. This iterative approach to reasoning—thinking followed by reflection—enhances the depth and rigor of the analysis, particularly in technical contexts. In contrast,the response from the non-fine-tuned model lacks such a reflective element, offering a more static presentation of ideas without delving into the nuances or reconsidering the assumptions behind its claims.  

While the response from the non-fine-tuned model offers a broad, functional perspective applicable to various fields, the response from the reasoning model provides a more rigorous, scientific analysis with a focus on the underlying mechanical principles. It offers a much deeper understanding of the structural advantages, particularly in biological and materials science applications. This and earlier inference examples show that the iterative, on-the-fly training method not only produces thinking and reflection sections but also provides deep domain knowledge. The iterative nature of the training strategy allows users to iteratively improve, enhance and refine training objectives.

\section{Conclusions}
This study addressed the challenge of fine-tuning generative models of synthetic intelligence, such as LLMs, to a specific domain, while endowing it with particular reasoning capabilities for enhanced modeling of scientific thinking (Figure~\ref{fig:Fig_100}c). Inspired by biological systems' adaptability and evolution, PRefLexOR’s recursive optimization approach mimics the processes through which natural materials achieve resilience and complexity. Just as biological systems self-organize and adapt to achieve optimal performance~\cite{Cranford2012Biomateriomics}, PRefLexOR uses iterative feedback loops to refine and evolve its reasoning pathways. This bioinspired approach allows the model to autonomously enhance its decision-making abilities, achieving coherence and adaptability reminiscent of nature's design principles, particularly in applications involving biological materials and cross-domain scientific discovery.

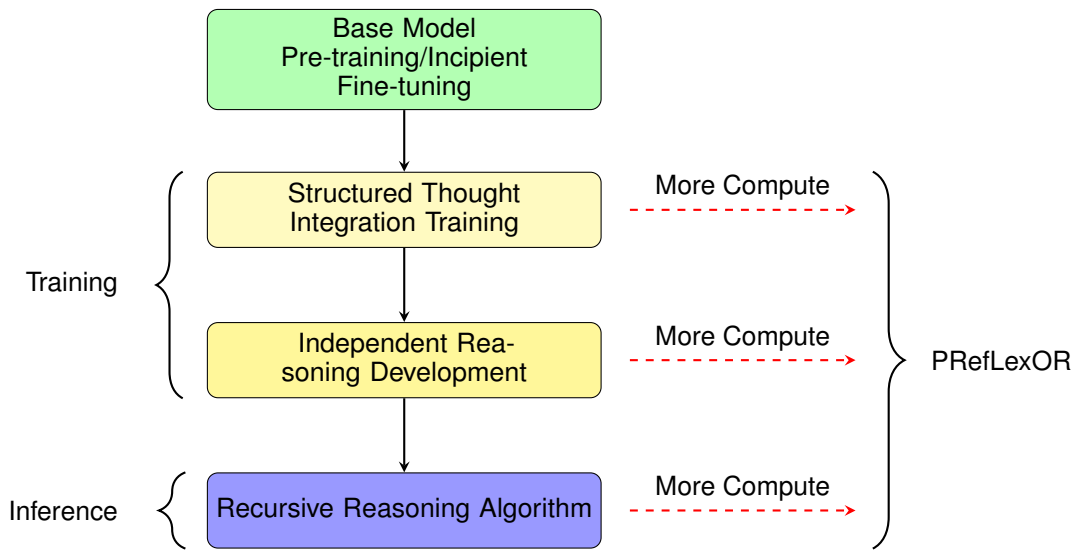
\begin{figure}[ht]
\centering
\sffamily % Switch to sans-serif font
\sansmath % Ensure math in flowchart is also sans-serif
\begin{tikzpicture}[node distance=2cm, every node/.style={align=center},
    startstop/.style={ellipse, draw, fill=blue!50, text width=5cm, text centered, minimum height=1cm},
    box0/.style={rectangle, draw, fill=green!30, text width=5cm, text centered, rounded corners, minimum height=1cm},
    box1/.style={rectangle, draw, fill=yellow!30, text width=5cm, text centered, rounded corners, minimum height=1cm},
    box2/.style={rectangle, draw, fill=yellow!50, text width=5cm, text centered, rounded corners, minimum height=1cm},
    box3/.style={rectangle, draw, fill=blue!40, text width=5cm, text centered, rounded corners, minimum height=1cm},
    scaling/.style={draw=red, thick, ->, >=stealth, dashed},] 
    % Nodes
    \node (phase0) [box0] {Base Model\\Pre-training/Incipient Fine-tuning};
    \node (phase1) [box1, below of=phase0] {Structured Thought Integration Training};
    \node (phase2) [box2, below of=phase1] {Independent Reasoning Development};
    \node (inference) [box3, below of=phase2] {Recursive Reasoning Algorithm};

    % Brace to indicate training phases
    \draw [decorate,decoration={brace,amplitude=10pt,mirror,raise=2pt}, thick]
        ([xshift=-0.25cm]phase1.north west) -- ([xshift=-0.25cm]phase2.south west) node[midway,xshift=-0.8cm,anchor=east] {Training};

    % Scaling arrows
    \draw [scaling] (3, -2) -- ++(3,0) node[midway, above] {More Compute};
    \draw [scaling] (3, -4) -- ++(3,0) node[midway, above] {More Compute};
    \draw [scaling] (3, -6) -- ++(3,0) node[midway, above] {More Compute};

    % Paths
    \draw [arrow] (phase0) -- (phase1);
    \draw [arrow] (phase1) -- (phase2);
    \draw [arrow] (phase2) -- (inference);
    
    % Brace for inference phase
    \draw [decorate,decoration={brace,amplitude=10pt,raise=1pt}, thick]
        ([xshift=-0.25cm]inference.south west) -- ([xshift=-0.25cm] inference.north west) node[midway,xshift=-0.8cm,anchor=east] {Inference};

    % Outer brace for the entire PRefLexOR algorithm
    % Outer brace for the entire PRefLexOR algorithm
    \draw [decorate,decoration={brace,amplitude=12pt,raise=3pt}, thick]
        ([xshift=3.5cm] phase1.north east) -- ([xshift=3.5cm] inference.south east) node[midway,xshift=0.75cm,anchor=west] {PRefLexOR};
    
\end{tikzpicture}
\sffamily % Restore default font
\caption{Overview of the PRefLexOR algorithm, consisting of \textit{Base Model Pre-training/Incipient Fine-tuning}, \textit{Structured Thought Integration Training}, \textit{Independent Reasoning Development}, and the \textit{Recursive Reasoning Algorithm}. Each phase can be scaled independently with additional compute to improve performance.}
\label{fig:three_phases_overview}
\end{figure}

We view this as an extension of more conventional physics or data-driven models that typically feature only forward capabilities without situational awareness. In other words, conventional models cannot assess the quality of their own predictions. For example, a Partial Differential Equation (PDE) will confidently predict solutions to boundary value problems whether or not the model actually captures the underlying physics), true for both physics-based or data-driven models (Figure~\ref{fig:Fig_100}b). In conventional scientific methods, humans will assess the quality of predictions using a host of methodologies, specifically logical assessment, additional data collection, comparison with literature, and more. Our quest to expand the reference to a model to include not only its forward capabilities but much broader situational awareness is, in our opinion, an important area of research that can benefit greatly from synthetic, or artificial intelligence~\cite{ghafarollahi2024sciagentsautomatingscientificdiscovery}, especially in applications to solving inverse materials design problems~\cite{BuehlerMSME_BuildBlock_2023,Ni2020AElasticity,Maurizi2022InverseResistance,Buehler2023MeLMProblemsc}. PRefLexOR offers one possible avenue to overcome these limitations through a multi-stage training and inference strategy, as visualized in Figure~\ref{fig:three_phases_overview}.

In our examples we aimed to develop a model with capabilities in the bio-inspired materials domain while following a structured thinking approach that elucidates great levels of detail. We developed a multi-stage dynamic training approach that uses on-the-fly dataset generation based on continually generated training data, reflecting not only an ability for the model to self-learn, but also a general strategy to efficiently develop domain-tuned models. The introduction of special thinking and reflection tokens provided us with a structured strategy to organize distinct reasoning tasks. 

The key contributions of PRefLexOR are:
\begin{itemize}
    \item A new integration of preference optimization with recursive reasoning to allow models to engage in multi-step thought refinement.
    \item A framework that uses thinking tokens to explicitly define and guide recursive reasoning within the model’s outputs.
    \item The incorporation of ORPO and preference optimization (e.g. DPO/EXO) to align model reasoning with human preferences through direct and fine-tuned optimization.
    \item An active learning mechanism that enables real-time task generation, ensuring flexibility and adaptability to new reasoning challenges.
    \item The application of recursive optimization that mirrors Reinforcement Learning feedback loops, allowing the model to self-teach and iteratively improve its cognitive capacities.
    \item Highly structured approaches to solve problems, especially relevant for science, can be used to endow the model with specific reasoning strategies relevant for particular domains.
    \item Since our training was conducted with LoRA adapters, it can be done efficiently on local GPU hardware, and easily extended to cover a wider range of adaptations (and it can be utilized, for instance, in mixture-of-expert strategies such as X-LoRA~\cite{Buehler2024X-LoRA:Design}).
\end{itemize}

Looking to a few specific examples of results, one of the most compelling highlights is the model’s ability to draw meaningful connections between seemingly disparate fields, such as its analogy between Hermann Hesse’s Glass Bead Game~\cite{hesse2000glass} and the hierarchical structure of proteins. This comparison underscores the model’s capacity for interdisciplinary reasoning, demonstrating how abstract philosophical concepts about interconnectedness and dynamic systems can be mapped onto concrete scientific phenomena, such as the layered complexity and functionality of biological systems. This synthesis of ideas illustrates the model’s potential to not only operate across diverse domains but also generate novel insights by bridging the gap between abstract thought and applied science. Another demonstration of interest was the transfer of the reasoning capability to new tasks, such as summarization and research proposal development. 

\subsection{Enhancing the Algorithm by Invoking Multidisciplinary Concepts from the Glass Bead Game}

The invocation of the  Glass Bead Game~\cite{hesse2000glass} goes beyond the use as a test case to probe the model's generalization capabilities, but forms also an analogy to what advanced reasoning models can do. In his novel, Hermann Hesse presents a game that synthesizes knowledge from various fields—such as mathematics, music, and philosophy—into a higher-order conceptual framework, with players combining ideas in ways that reveal deeper patterns and insights. Within the scope of PRefLexOR, this game becomes a metaphor for how thinking and reflection processes in reasoning models, operate. Just as players of the game engage in an iterative exploration of connections between disparate disciplines, LLMs with thinking and reflection phases mimic this recursive synthesis. The ``thinking'' and ``reflection'' phases, along with recursive agentic self-improvement, allows the model to explore multiple layers of reasoning and refinement for cohesive responses (see, e.g., Figure~\ref{fig:PreFlexOR_recursive}, much like the Glass Bead Game connects concepts across domains. The structured interplay of thought and reflection in reasoning models echoes the intellectual depth and complexity of Hesse's game, suggesting that, like the Glass Bead Game, such models may be capable of uncovering rich, interdisciplinary insights when guided by sophisticated reasoning strategies. This capacity to connect and reflect upon diverse ideas highlights the potential of LLMs to act as powerful tools for understanding, much like the characters in The Glass Bead Game use their symbolic play to explore the essence of knowledge itself as it resembles connections between bits of information, as shown in Figure~\ref{fig:Fig_100}a.

Specifically, the Glass Bead Game as proposed in the novel~\cite{hesse2000glass}, is a symbolic system that serves as ``\textit{a kind of synthesis of human learning}.'' The game represents a means of integrating and refining knowledge from diverse disciplines, such as mathematics, music, and philosophy. Players engage in an iterative process, continuously refining and revisiting concepts to discover deeper relationships between them. Similarly, the algorithm in this method employs a recursive approach where a fine-tuned Reasoning Model generates an initial response, which is then subjected to reflection and improvement through multiple iterations.

The process begins with the generation of an initial response, analogous to the first move in the Glass Bead Game, where the players begin with basic knowledge. As in the game, where players continually refine their moves through reflective thought, the algorithm extracts reflections from the initial response, enhancing the reasoning behind it. The Critic Model plays a role much like the intellectual rigor imposed by the rules of the Glass Bead Game, providing an evaluative framework that helps guide the refinement of responses. Through this iterative process, the model improves its output, cycling between generating new responses and reflecting on previous iterations until an optimal or integrated solution is reached.

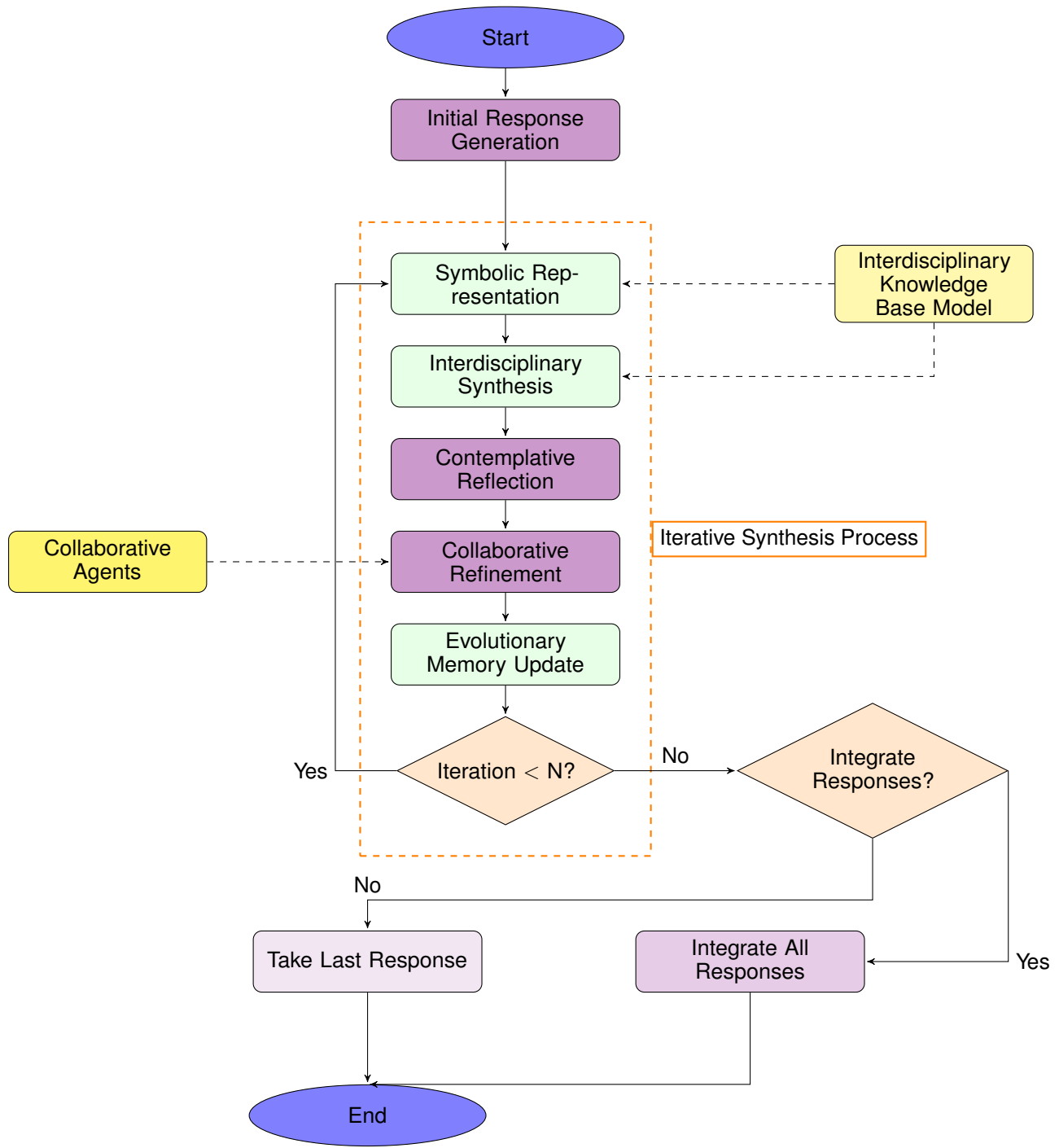
\begin{figure}
\sffamily % Switch to sans-serif font
\sansmath % Ensure math in flowchart is also sans-serif

\begin{tikzpicture}[node distance=1.5cm and 2.5cm, auto,
    % Define styles
    block/.style={rectangle, draw, fill=violet!40, text width=3.5cm, text centered, rounded corners, minimum height=1cm},
    block2/.style={rectangle, draw, fill=violet!20, text width=3.5cm, text centered, rounded corners, minimum height=1cm},
    block3/.style={rectangle, draw, fill=violet!10, text width=3.5cm, text centered, rounded corners, minimum height=1cm},
    block4/.style={rectangle, draw, fill=green!10, text width=3.5cm, text centered, rounded corners, minimum height=1cm},
    decision/.style={diamond, draw, fill=orange!20, text width=3cm, text centered, inner sep=0pt, aspect=2},
    startstop/.style={ellipse, draw, fill=blue!50, text width=2.5cm, text centered, minimum height=1cm},
    model/.style={rectangle, draw, fill=yellow!40, text width=3cm, text centered, rounded corners, minimum height=1cm},
    model2/.style={rectangle, draw, fill=yellow!70, text width=3cm, text centered, rounded corners, minimum height=1cm},
    line/.style={draw, -stealth', shorten >=1pt},
    dashed line/.style={draw, dashed, -stealth', shorten >=1pt}]
    % Define nodes
    \node [startstop] (start) {Start};
    \node [block, below=0.5cm of start] (init) {Initial Response Generation};
    \node [block4, below=1.5cm of init] (symbolic) {Symbolic Representation};
    \node [block4, below=0.5cm of symbolic] (synthesis) {Interdisciplinary Synthesis};
    \node [block, below=0.5cm of synthesis] (contemplation) {Contemplative Reflection};
    \node [block, below=0.5cm of contemplation] (collaborative) {Collaborative Refinement};
    \node [block4, below=0.5cm of collaborative] (evolution) {Evolutionary Memory Update};
    \node [decision, below=0.5cm of evolution] (iterate) {Iteration $<$ N?};
    \node [decision, right=2cm of iterate] (final_decision) {Integrate \\ Responses?};
    \node [block2, below=1.5cm of final_decision, xshift=-2cm] (integrate) {Integrate All Responses};
    \node [block3, left=of integrate] (last_response) {Take Last Response};
    \node [startstop, below=of last_response] (end) {End};
    % Model nodes
    \node [model2, left=of collaborative, xshift=-0.5cm] (agents) {Collaborative Agents};
    \node [model, right=of symbolic, xshift=1cm] (knowledge_base) {Interdisciplinary \\ Knowledge Base Model};
    % Connect nodes
    \path [line] (start.south) -- (init.north);
    \path [line] (init.south) -- (symbolic.north);
    \path [line] (symbolic.south) -- (synthesis.north);
    \path [line] (synthesis.south) -- (contemplation.north);
    \path [line] (contemplation.south) -- (collaborative.north);
    \path [line] (collaborative.south) -- (evolution.north);
    \path [line] (evolution.south) -- (iterate.north);
    % Iteration decision
    \path [line] (iterate.west) -- ++(-1cm,0) node [pos=1, left] {Yes} |- (symbolic.west);
    \path [line] (iterate.east) -- node [above] {No} (final_decision.west);
    % Final decision branches
    \path [line] (final_decision.south) -- ++(0,-1) -| node [midway, above] {No} (last_response.north);
    \path [line] (final_decision.east) |- node [right] {Yes} (integrate.east);
    \path [line] (last_response.south) -- (end.north);

    \path [line] (integrate.south) |- (end.north);
    % Connect models
    \path [dashed line] (knowledge_base.west) -- (symbolic.east);
    \path [dashed line] (knowledge_base.south) |- (synthesis.east);
    \path [dashed line] (agents.east) -- (collaborative.west);
    % Add a frame around the iterative process
    \begin{pgfonlayer}{background}
        \node [draw=orange, thick, dashed, fit=(symbolic) (synthesis) (contemplation) (collaborative) (evolution) (iterate), inner sep=0.5cm, 
               label={[fill=white, draw=orange, thick]right:Iterative Synthesis Process}] {};
    \end{pgfonlayer}
\end{tikzpicture}
 \caption{Flowchart of an expanded PRefLexOR algorithm forming a new approach that expands the original Recursive Reasoning Algorithm depicted in Figure~\ref{fig:PreFlexOR_recursive}. The diagram of this proposed approach illustrates how the algorithm incorporates symbolic representation, interdisciplinary synthesis, and collaborative refinement to enhance its reasoning capabilities. This integration aligns the algorithm with the game's emphasis on the unity of knowledge and deep contemplation across domains, knowledge fields, and modalities. A key shift, comparing this to Figure~\ref{fig:PreFlexOR_recursive}, is that a singularly focused Reasoning Model was replaced with a set of Collaborative Agents that have develop particular capabilities to examine logical steps towards solving a problem, and the Critic with an Interdisciplinary Knowledge Base Model that transcends across boundaries of fields.  An additional feature of emphasis is the utilization of symbolic representation of knowledge; a feat that may resemble our incipient attempt to formulate certain categories of thinking (see, Table~\ref{tab:categories}) that yield highly structured thought processes.}
\label{fig:PreFlexOR_recursive_GlassBead}
\sffamily % Switch to sans-serif font
\end{figure}

This recursive thinking and reflection model mirrors the way the Glass Bead Game synthesizes diverse strands of knowledge into a cohesive whole. Just as the game is meant to model a kind of synthesis of human learning, the algorithm integrates reasoning and reflection to create responses that combine multiple iterations of thought into a more comprehensive final answer. In this way, the recursive algorithm not only produces more refined outputs but also illustrates how generative AI can emulate deep, interdisciplinary reasoning, much like the intellectual pursuit portrayed in Hesse’s game. Figure~\ref{fig:PreFlexOR_recursive_GlassBead} depicts a possible flowchart of such an algorithm that merges ideas proposed in the PRefLexOR framework with the process introduced in the Glass Bead Game. 

In the integrated framework, a simple Reasoning Model is replaced with a set of Collaborative Agents, each acting as an individual reasoning engine with specialized expertise or perspectives (or a single model with distinct sets of special tokens to induce a particular type of reasoning specialty). This transformation allows the algorithm to simulate a community of thinkers, reflecting the collective intellectual exploration emphasized in the Glass Bead Game~\cite{hesse2000glass}. By incorporating multiple reasoning models as collaborative agents, the algorithm harnesses diverse viewpoints and methodologies, enhancing creativity, depth, and robustness in problem-solving. Each agent contributes unique insights, challenges others' ideas, and collaboratively refines responses through iterative dialogue, much like the scholars in the Glass Bead Game who engage in symbolic synthesis across disciplines.

Similarly, the Critic is replaced with an Interdisciplinary Knowledge Base Model, serving as a rich repository of information from various fields that all agents can access and utilize. This shift moves the focus from evaluation to synthesis, aligning the algorithm with the game's emphasis on the unity of knowledge and deep contemplation by finding new connections~\cite{BuehlerGraphReasoning2024}. The knowledge base enables agents to draw connections across different domains, fostering holistic understanding and allowing for more profound insights. By integrating this shared resource, the algorithm encourages collaborative synthesis rather than hierarchical critique, mirroring the Glass Bead Game's practice of unifying arts and sciences through collective intellectual endeavor.

These revisions emulate collective intellectual endeavors at high levels of integrated societal scales, simulating a community of thinkers enhances the algorithm's ability to explore complex problems from multiple angles. It incorporates diverse expertise via agents with specialized knowledge contribute to a more comprehensive and nuanced understanding. This is believed to improve problem-solving as collaborative refinement leads to innovative solutions and deeper insights. Replacing the Critic with the Interdisciplinary Knowledge Base Model~\cite{BuehlerGraphReasoning2024,ghafarollahi2024sciagentsautomatingscientificdiscovery} improves the algorithm by facilitating knowledge synthesis, providing agents with access to a broad spectrum of information promotes interdisciplinary connections. Emphasizing synthesis over critique fosters a holistic approach to reasoning by aligning with universal concepts of structures in knowledge representations, e.g. identified via isomorphic mappings. A shared knowledge base serves as common ground for agents to collaboratively build upon ideas, as was demonstrated already in earlier work using graph reasoning~\cite{BuehlerGraphReasoning2024}. This reconfiguration aligns the algorithm with the philosophical foundations of the Glass Bead Game, enhancing its capacity for profound, interconnected, and innovative reasoning. It transforms the algorithm into a more powerful system that mirrors the game's emphasis on collaborative exploration, symbolic synthesis, and the unity of knowledge, ultimately leading to richer and more insightful responses.

A feature of importance is the use of symbolic representation of knowledge. This may resemble our incipient attempt to formulate certain categories of thinking as already conducted in the PRefLexOR algorithm implemented in this study. For instance, we refer to Table~\ref{tab:categories}) that yield highly structured thought processes here tailored to the field of biological design. We anticipate that we can structure these inherently unique to meet a more generalistic logical progression of ideas. Alternatively, we may be able to develop concepts that utilize reasoning before decoding hidden states into tokens (as done in X-LoRA) to yield highly complex, abstract, thought processes. Alternatively, one may utilize a finite set of algebra to force a bottleneck of expressing reasoning in a narrow vocabulary of relationships.   
 
As shown in Figure~\ref{fig:PreFlexOR_recursive_GlassBead}, the integration of these components transforms the response generation process into a multidimensional, reflective system. First, symbolic representation abstracts initial responses into a universal form, facilitating manipulation across disciplines. This feeds into interdisciplinary synthesis, where knowledge from diverse fields enriches the response, promoting unity of understanding. Through contemplative reflection, deeper insights are uncovered, while collaborative refinement allows multiple agents to contribute diverse perspectives, enhancing the intellectual depth of the process. The system undergoes an evolutionary memory update, incorporating new insights for continuous learning. The entire process is driven by an iterative synthesis, looping through these stages until a refined and comprehensive understanding is achieved, mirroring the intellectual rigor of the Glass Bead Game. This revised PRefLexOR algorithm may thereby further enhance its capacity for profound, interconnected reasoning, aligning with the philosophical foundations of the Glass Bead Game. This results in responses that are richer, more innovative, and deeply reflective of a unified knowledge framework.

\begin{table}[H]
\centering
\small % Reduce font size within the table
\renewcommand{\arraystretch}{1.} % Adjust row height
\begin{tabular}{|p{2.8cm}|p{4.2cm}|p{4.2cm}|p{3.8cm}|}
\hline
\textbf{Component} & \textbf{Description} & \textbf{Integration with the Modeling Framework} & \textbf{Relation to Glass Bead Game Concepts} \\
\hline
\textbf{Symbolic Representation} &
Converts the initial response into symbolic form using a universal language or encoding. &
Uses \textbf{Thinking Tokens} to guide the LLM in creating symbolic abstractions, facilitating abstraction and generalization of ideas, and enabling manipulation and combination of concepts. &
Emphasizes the use of a \textbf{symbolic language} to connect ideas, reflecting the game's core activity of manipulating symbols. \\
\hline
\textbf{Interdisciplinary Synthesis} &
Combines symbolic representations with knowledge from multiple disciplines to enrich the response. &
The LLM accesses an \textbf{Interdisciplinary Knowledge Base}, uses \textbf{Thinking Tokens} to integrate diverse insights, enhancing creativity and innovation. &
Mirrors the game's synthesis of \textbf{arts and sciences}, promoting the \textbf{unity of knowledge}. \\
\hline
\textbf{Contemplative Reflection} &
Engages in deep, meditative reflection on the synthesized knowledge to uncover hidden insights. &
Utilizes \textbf{Reflection Tokens} for introspection; the LLM performs deep analysis of implications and principles. &
Captures the game's \textbf{meditative and introspective} aspects, encouraging profound \textbf{contemplation}. \\
\hline
\textbf{Collaborative Refinement} &
Multiple collaborative agents contribute diverse perspectives to refine the response. &
Implements \textbf{Multi-Agent Interaction} among LLMs; agents use \textbf{Thinking and Reflection Tokens} to contribute and evaluate ideas, simulating a \textbf{community of thinkers}. &
Emulates the game's \textbf{collective intellectual endeavor}, enhancing responses through \textbf{collaborative intelligence}. \\
\hline
\textbf{Evolutionary Memory Update} &
Updates the system's memory with new insights, enabling evolution over iterations. &
The LLM stores successful patterns and connections; uses \textbf{Thinking Tokens} for learning and \textbf{Reflection Tokens} for evaluation, improving future reasoning strategies. &
Reflects the game's \textbf{evolutionary iteration}, supporting continuous \textbf{growth and learning}. \\
\hline
\textbf{Explicit Abstraction} &
Makes abstraction an intentional and directed process within the framework. &
Guides the LLM to align with specific goals; enhances the quality and depth of reasoning; uses explicit prompts for abstraction. &
Aligns with the game's emphasis on \textbf{symbolism and abstraction}, facilitating the creation of \textbf{harmonious connections}. \\
\hline
\textbf{Bridging Implicit and Explicit Abstraction} &
Connects the LLM's inherent abstraction with explicit symbolic reasoning. &
Combines the LLM's strengths with guided processes; enhances explainability and control; leverages both implicit and explicit reasoning. &
Enhances the game's practice of connecting \textbf{visible and underlying patterns}, enriching the \textbf{intellectual depth} of the process. \\
\hline
\end{tabular}
\vspace{0.25cm}
\caption{This table summarizes how key concepts from the Glass Bead Game are integrated into a future multi-agent reasoning framework. By incorporating components such as symbolic representation, interdisciplinary synthesis, and collaborative refinement, the algorithm enhances its capacity for profound, interconnected responses, aligning with the game's emphasis on the unity of knowledge and deep contemplation.}
\label{table:glass_bead_game_integration}
\end{table}

For further delineation of key analogies and processes, Table~\ref{table:glass_bead_game_integration} provides a comprehensive overview of how key concepts are integrated into the multi-agent reasoning framework. Each component represents a crucial enhancement to your algorithm, enabling it to generate more profound, interconnected, and innovative responses, and we provide a direct delineation with the existing algorithm.
The task of symbolic representation seeks to convert the initial response generated by the model into symbolic form using a universal language or encoding system. This facilitates manipulation and combination of concepts across different domains. For example, if the LLM provides an initial explanation of a biological process, this step translates key concepts like ``cell division'' or ``DNA replication'' into symbols or diagrams, emphasizing the use of a symbolic language to connect ideas. This mirrors the game's core activity of manipulating symbols to reveal deep connections between disciplines. This could be accomplished by introducing a special token for this particular purpose, to teach models to achieve such abstraction. 
Next, interdisciplinary synthesis combines symbolic representations with knowledge from multiple disciplines to enrich the response. The LLM accesses a diverse knowledge base spanning various fields and uses thinking tokens to integrate insights from different domains. For instance, integrating mathematical models with philosophical theories to address complex problems like ethical considerations in artificial intelligence. This mirrors the game's synthesis of arts and sciences, promoting the unity of knowledge. This can be accomplished by introducing yet another special token for this particular purpose. 
During contemplative reflection, the process engages in deep, meditative reflection on the synthesized knowledge to uncover hidden insights. The LLM uses reflection tokens to introspect and perform deep analysis. For example, after synthesizing information, the LLM reflects on the ethical implications of its conclusions, considering long-term impacts. This captures the game's meditative and introspective aspects, encouraging profound contemplation.
In collaborative refinement, multiple collaborative agents contribute diverse perspectives to refine the response. Implements multi-agent interaction among LLMs, where agents contribute ideas and evaluate each other's inputs. For example, agents specializing in different fields collectively refine the response: For instance, one focusing on technical accuracy, another on ethical considerations, and a third on societal impact, physical soundness, experimental feasibility, and so on. This emulates the game's collective intellectual endeavor, enhancing responses through collaborative intelligence.
The step evolutionary memory update updates the system's memory with new insights, enabling evolution over iterations. The model stores successful patterns and connections for future use (e.g. via category graph representations), using thinking and reflection tokens to guide learning and evaluate effectiveness. This reflects the game's evolutionary iteration, supporting continuous growth and learning. 
During explicit abstraction, the model renders an abstraction an intentional and directed process within the framework. Uses explicit prompts to direct model's abstraction efforts toward specific goals, improving the depth and coherence of reasoning. For instance, instructing the model to focus on abstract principles underlying data rather than just summarizing it, perhaps using symbolic mechanisms, or using similar special tokens as used above during the initial symbolic representation. This aligns with the game's emphasis on symbolism and abstraction, facilitating the creation of harmonious connections.

We can further seek to endow the model with capabilities to conduct implicit and explicit abstraction, where we connect the inherent abstraction capabilities with explicit symbolic reasoning. Combines the model's strengths with guided processes, enhancing transparency in the reasoning process. While the LLM may naturally abstract concepts, the framework ensures these abstractions are represented in alignment with the overall reasoning strategy. This enhances the game's practice of connecting visible and underlying patterns, enriching the intellectual depth of the process.

As the algorithm proceeds, the flowchart itself may be optimized by planning agents, akin to what has been reported in other multi-agent systems, for instance using concepts from graph reasoning~\cite{ghafarollahi2024sciagentsautomatingscientificdiscovery}. This can include suggestions for deepening interdisciplinary connections by expanding the knowledge base, enhancing reflective depth through recursive self-improvement loops, optimizing collaboration with dynamic agent roles, and leveraging human-AI synergy for oversight and input.

Integrating these components may ultimately align the algorithm with the philosophical and methodological foundations of the Glass Bead Game and thereby enhance its capacity to generate responses rich in insight, creativity, and interconnected understanding, encouraging the algorithm to transcend traditional problem-solving approaches and embrace a holistic, integrative perspective.

\subsection{Future Work, Challenges and Opportunities}

Several avenues for future work offer exciting opportunities to enhance the capabilities of our model. Key directions include exploring agentic reasoning strategies, such as AutoGen~\cite{Wu2023AutoGen:Conversation} and high degrees of agentic modeling via swarm-based approaches, and scaling to larger models for increased performance. Additionally, testing the model’s generalizability across diverse domains and incorporating multiple thinking sections with partial masking are promising methods for improving reasoning efficiency.

While PRefLexOR demonstrates promising results in enhancing AI reasoning capabilities, particularly in biological materials science, several limitations warrant further investigation. The framework's increased computational cost, especially in its recursive phases, may limit real-time applications, necessitating optimization strategies. However, in cases where compute is not an issue, such as scientific discovery, this may not present a significant burden. Its current focus on specific domains suggests that future work should explore other areas of applications including broader, multi-disciplinary training. 

Future work may also focus on refining reasoning strategies, including more structured outputs (e.g. additional steps to discovery reasoning categories from data) and integrating other methods, potentially mixing various approaches for optimal outcomes. One direction is to trigger different reasoning strategies based on task type or allow the model to autonomously detect the best approach. For example, logic-based questions might follow a distinct reasoning pathway compared to materials design or regression tasks. The use of symbolic reasoning may further enhance generalization capabilities, perhaps combined with graph theoretic concepts such as isomorphic analysis as was suggested in other work~\cite{BuehlerGraphReasoning2024}. This adaptability offers remarkable flexibility and precision in addressing diverse challenges.

More sophisticated agentic modeling can be another promising next step, where reasoning or reflection stages are critiqued or assessed for feasibility, particularly in areas such as physical design or materials science. By incorporating reflective critique, the model can continuously refine its reasoning processes. For example, reasoning steps could be critiqued based on real-world constraints, such as physical feasibility or design limitations, to ensure solutions are not only theoretically sound but practically viable.

Models can also benefit from improved reasoning feedback loops, where the reasoning steps are continuously refined based on the input obtained from the reflection phase, ultimately leading to higher-quality outputs. For instance, if an initial reasoning process lacks key considerations about material properties or environmental factors, the reflection process can identify these gaps, leading to a more complete and accurate solution in the final output. Naturally, the method can be expanded also to offer a variety of reasoning strategies during the initial \textit{Structured Thought Integration Training} phase, so that a greater variety of thinking mechanisms can be utilized in the second phase. 

This iterative enhancement of reasoning will result in models that are not only more intelligent but also capable of producing outputs that are better aligned with complex, real-world challenges.

\section{Materials and Methods}

\subsection{Special Tokens for Reasoning}

In this work, several special tokens were introduced to improve the structured reasoning and reflection capabilities of the model. These tokens are integrated into the tokenizer of the Llama 3.2 model~\cite{dubey2024llama3herdmodels,llama3_website} and help guide the model in generating specific types of outputs, such as thinking steps, reflective improvements, and final answers, while providing structured reasoning pathways during the training process.

The following special tokens were added:

\begin{itemize}
    \item \texttt{<|response|>} and \texttt{<|/response|>} - Used to demarcate the boundaries of the final answer or response provided by the model.
    \item \texttt{<|reflect|>} and \texttt{<|/reflect|>} - Used to mark the reflection phase, where the model evaluates and improves upon its initial reasoning.
    \item \texttt{<|thinking|>} and \texttt{<|/thinking|>} - Used to denote the thinking phase, where the model generates its reasoning steps.
    \item \texttt{<|scratchpad|>} and \texttt{<|/scratchpad|>} - Optionally used to provide a scratchpad for interim steps, allowing the model to store intermediary calculations or thoughts during inference.
\end{itemize}

These tokens allow for a clear delineation of different reasoning processes and phases within the model’s output, enabling it to engage in reflective and structured thinking. Below is a summary of these tokens and their properties, including their token IDs in the customized tokenizer:

\begin{table}[h!]
\small
\centering
\begin{tabular}{|p{5cm}|p{5cm}|}
\hline
\textbf{Token ID} & \textbf{Token}   \\
\hline
128252 & \texttt{<|thinking|>}  \\
\hline
128253 & \texttt{<|/thinking|>}   \\
\hline
128250 & \texttt{<|reflect|>}  \\
\hline
128251 & \texttt{<|/reflect|>}   \\
\hline
128254 & \texttt{<|scratchpad|>}  \\
\hline
128255 & \texttt{<|/scratchpad|>}   \\
\hline
\end{tabular}
\vspace{0.25cm}
\caption{List of special tokens used during model training with the updated Llama 3.2 tokenizer. Only the \texttt{<|thinking|>}/\texttt{<|/thinking|>} and \texttt{<|reflect|>}/\texttt{<|/reflect|>} tokens are used in this work, but the approach can be extended to other concepts, such as scratchpads, sections for symbolic representation of reasoning, and other tokens.}
\label{tab:special_tokens}
\end{table}

These tokens are instrumental in organizing and structuring the model's reasoning and reflection capabilities, allowing for more precise control over the model's inference and answer generation process. The tokenizer is available as part of the models developed in this work, or separately at \url{lamm-mit/meta-llama-Meta-Llama-3.2-3B-Instruct-Reasoning-Tokenizer}.

\subsection{On-the-fly dataset generation via \textit{in-situ} knowledge extraction}

The algorithm is designed to questions from a given context and provide both correct and incorrect answers. The process is conducted  \textit{in-situ} during training and consists of several key steps, which are described below.

\subsubsection{Context Enhancement with Retrieval-Augmented Generation during Dataset Generation}
The context is enriched using Retrieval-Augmented Generation (RAG). This process involves querying the index with the generated question to retrieve additional relevant information and reasoning, which is appended to the original context. 

We build an index of text embeddings to facilitate efficient retrieval-augmented generation (RAG). It transforms each text chunk \( T_i \) from a corpus of original raw data into a dense vector representation \( \mathbf{v}_i \) using the embedding model: 
\begin{equation}
\mathbf{v}_i = f_{\text{embed}}(T_i)
\end{equation}
where \( f_{\text{embed}} \) is the embedding function. When a query \( Q \) is generated, it is similarly encoded into a vector \( \mathbf{v}_q \):
\begin{equation}
\mathbf{v}_q = f_{\text{embed}}(Q)
\end{equation}
Llama Index then computes the cosine similarity between \( \mathbf{v}_q \) and each \( \mathbf{v}_i \) in the index:
\begin{equation}
\text{similarity}(\mathbf{v}_q, \mathbf{v}_i) = \frac{\mathbf{v}_q \cdot \mathbf{v}_i}{\|\mathbf{v}_q\| \|\mathbf{v}_i\|}
\end{equation}
The most relevant vectors are selected based on this similarity measure, retrieving the corresponding text chunks, \( T_j \), which are then appended to the original query context. This expanded context allows the LLM to generate a response that incorporates both the retrieved information and the pre-existing knowledge, improving the depth and relevance of the output.

We use the~\url{BAAI/bge-large-en-v1.5} text embedding model in RAG implemented in Llama Index~\cite{Run-llama/llama_index:Applications}. 

\subsubsection{Raw data used for training}

We use 500 scientific papers from the domain of biological and bio-inspired materials as the training data, as reported in earlier work~\cite{Luu2023BioinspiredLLM:Materials}. To construct the raw corpus of text, we convert all PDFs into Markup language and then create text chunks.  We use the LlamaIndex SentenceSplitter function  with chunk size of 1024 tokens with chunk overlap of 20 tokens.

\subsubsection{Context Retrieval}
The algorithm first retrieves relevant context information from a pre-constructed index of nodes. When a specific topic is provided, it selects nodes related to that topic; otherwise, it retrieves a random set of nodes ($n=3$ in the work reported here). The text from the selected nodes is concatenated into a single context, which serves as the basis for question generation. The token length of the concatenated context is computed using a tokenizer.

\subsubsection{Question Generation}
A domain-specific question is generated based on the provided context using a text generation model. The question is formulated to capture an important aspect of the context, without referring to specific studies, papers, or authors. The question is intended to be challenging, requiring expert-level knowledge to answer. The prompt used is:

\begin{LLMbox}{Generation of question} 
\begin{lstlisting} 
You are a Teacher/Professor. Your task is to set up a quiz/examination. Using information in the provided context, formulate a single question that captures an important fact from the context.

Restrict the question to the context information provided, and make sure this is a question that a highly trained domain expert can answer without seeing the context.

Just return the question, nothing else. Do not refer to the context, a paper, names, or authors, just ask the question.

The question must be challenging, deep, and stand on its own and query facts and expert domain knowledge. The question must NOT refer to a study, paper, or a specific author. 
\end{lstlisting} 
\end{LLMbox}

\subsubsection{Category-Based Information Extraction}
The algorithm extracts structured information from the context based on several predefined categories~\cite{Giesa2012CategoryDesign}. These categories include reasoning steps, relevant materials, and design principles, among others. For each category, the model generates a well-reasoned, concise explanation, which contributes to a deeper understanding of the question. The predefined categories are listed in Table~\ref{tab:categories}.

\begin{table}[h!]
\small
\centering
\begin{tabular}{|p{5cm}|p{9cm}|}
\hline
\textbf{Category} & \textbf{Description} \\ \hline
Reasoning Steps & Logical steps that explain the reasoning behind the answer. \\ \hline
Relevant Materials or Concepts & Key materials or scientific concepts related to the context. \\ \hline
Design Principles & Design-related considerations from the context. \\ \hline
Material Properties & Important properties of materials discussed in the context. \\ \hline
Hypothesis & A proposed explanation based on the context. \\ \hline
\end{tabular}
\vspace{0.25cm}
\caption{Categories used for extracting structured information from the context.}
\label{tab:categories}
\end{table}

For each of the categories shown in Table~\ref{tab:categories}, we use this prompting strategy:

\begin{LLMbox}{Extraction of information by category} 
\begin{lstlisting} 
Based on the context, extract the "{category}" relevant to the question. Keep it brief and avoid lists.

Question: {question}

Context: {context}

Provide only the "{category}" without additional explanations.

If you cannot find any, respond with an empty string. Keep the answer brief, but use step-by-step reasoning and a clear explanation.

Just provide the answer. Do not use lists; rather, develop contents written out in logical ideas.

Do not refer to the context or specific figures, text, sections, or others. 
\end{lstlisting} 
\end{LLMbox}

This approach ensures a highly structured strategy to thinking through a particular problem space or domain. It can be modified, e.g. via the use of a set of special tokens and/or specially trained LoRA adapters, to obtain specific thought processes that align with a particular aspect of reasoning. For instance, we can focus one reasoning process on design, another on manufacturing, another on biology, and so on. In the scope of symbolic reasoning, this can also be used to create a higher abstraction of the reasoning process.

In the work reported in this paper, we limit the scope to a single thinking section but with multiple categories embedded withn to represent multiple streams of analysis. 

\subsubsection{Thinking Section for Reasoning}
The extracted information from each category is assembled into a ``Thinking Section for Reasoning'' This section is designed to aid in the reasoning process by providing structured, logical insights. The Thinking Section includes key pieces of information from each category, which help guide the construction of the correct answer. It serves as a structured reasoning framework for answering the question.

\subsubsection{Correct and Incorrect Answer Generation}
The correct answer is generated using the context and the Thinking Section. The reasoning included in the Thinking Section helps to formulate a well-structured and comprehensive response. Additionally, an incorrect (rejected) answer is generated either by a trained model or through a prompt-based approach. The rejected answer lacks logical reasoning and does not reference the correct context.

The correct answer is generated as follows:

\begin{LLMbox}{Generation of correct response} 
\begin{lstlisting} 
Using the context provided, answer the following question:

Question: {question}

Context: {context}

Provide a comprehensive and accurate answer. 
\end{lstlisting} 
\end{LLMbox}

In the first training stage, the rejected answer is generated by requesting the model to create an incorrect answer, as follows:

\begin{LLMbox}{Generation of rejected (incorrect) response} 
\begin{lstlisting} 
You are to provide an incorrect answer to the question below.

Question: {question}

Do not include any reasoning or refer back to the question.

Just provide the incorrect answer. 
\end{lstlisting} 
\end{LLMbox}

It is noted optionally, and used always in the second stage of training, the current trained model can be used to generate an answer using simply the question. 

\subsubsection{Final Output}
The algorithm outputs three elements:
\begin{itemize}
    \item The generated question with an instruction to include the Thinking Section for Reasoning.
    \item The correct answer, which is enhanced with the structured Thinking Section for Reasoning.
    \item The rejected answer, which is designed to be incorrect and devoid of proper reasoning (in ORPO phase) or an answer generated based on the current trained state of the model (in DPO/EXO phase).
\end{itemize}

\begin{figure}[h!]
    \centering
    \small
    \begin{LLMbox}[colback=gray!15!white, colframe=black, title=Algorithm Overview]
    
    \vspace{0.25cm}
    1. Retrieve context from the index by randomly selecting a number of text chunks (here, we use $n=3$); whereas the text chunks are concatenated. \\
    2. Generate a domain-specific question based on the context. \\
    3. Enhance the context with RAG, which allows retrieval of information from the entire corpus of data. \\
    4. Extract structured information based on key categories. \\
    5. Assemble the Thinking Section for Reasoning. \\
    6. Generate the correct and incorrect answers.
    \end{LLMbox}
    \caption{Overview of the algorithm used for question generation and answering, leading to a prompt, chosen and rejected responses.}
    \label{fig:algorithm}
\end{figure}

\subsubsection{Reflection Section}
 When we use an additional reflection section, we introduce an introspective step in the algorithm that critiques the reasoning process used to generate an answer. This function asks the model to evaluate the thinking behind the generated answer and suggest improvements. The reflection process is guided by the following prompt:
 
\begin{LLMbox}{Generation of reflection section}
\begin{lstlisting}
Analyze the strategy to answer the question and suggest improvements or corrections.

Question: {question}

Strategy: {thinking}

Do not answer the question, just suggest improvements or corrections, such as but not limited to missing facts, or other considerations of relevance.

Do not refer to the context. Keep it short.
\end{lstlisting}
\end{LLMbox}

\subsubsection{Models used for Dataset Generation}

We use the \url{mistralai/Mistral-Nemo-Instruct-2407} model for dataset generation. We also experimented with \url{meta-llama/Llama-3.1-8B-Instruct} for some training runs, which works well also. Alternatively, a host of other models can be used including more sophisticated models (e.g. \url{o1}, \url{gpt-4o}, \url{Claude Sonnett 3.5}, etc.) but we deliberately focused on small-scale open-source models for this  study. 

\subsection{Handling Reasoning Tokens in Preference Alignment Loss Computation}

In our algorithm we revise conventional preference optimization frameworks to handle learning intermediate reasoning steps, referred to as ``thinking tokens.'' These tokens represent the model's internal reasoning processes and are enclosed by special tokens: \texttt{<|thinking|>} and \texttt{<|/thinking|>}. Our primary objective is to exclude inner thinking tokens from contributing to the loss computation while ensuring that the model learns to generate the \texttt{<|thinking|>} and \texttt{<|/thinking|>} tokens correctly. We introduce two distinct approaches to achieve this: \textit{masking of thinking tokens} (suitable for multiple sections of thinking sections) and \textit{dynamic final answer detection} (where all tokens during the thinking section(s) up to the the last \texttt{<|/thinking|>} token are masked).

\subsubsection{Masking of Thinking Tokens in Multiple Sections}
In this approach, all tokens between \texttt{<|thinking|>} and \texttt{<|/thinking|>} are masked, meaning they are excluded from the log-probability computation and the subsequent loss calculation. However, the \texttt{<|thinking|>} and \texttt{<|/thinking|>} tokens themselves are included in the loss calculation to ensure that the model learns to produce these tokens correctly.

For a sequence of token IDs \(\mathbf{t} = [t_1, t_2, \dots, t_n]\) and log-probabilities \(\mathbf{p} = [p_1, p_2, \dots, p_n]\), a boolean mask \(\mathbf{m} = [m_1, m_2, \dots, m_n]\) is applied, where:

\begin{equation}
m_i = 
\begin{cases}
0 & \text{if } t_i \text{ is an inner thinking token}, \\
1 & \text{otherwise (including \texttt{<|thinking|>} and \texttt{<|/thinking|>})}.
\end{cases}
\end{equation}
The masked log-probabilities \(\mathbf{p}_{\text{masked}}\) are computed as:
\begin{equation}
\mathbf{p}_{\text{masked}} = \mathbf{p} \odot \mathbf{m},
\end{equation}
where \(\odot\) denotes element-wise multiplication.

This approach ensures that the inner thinking tokens are ignored during the DPO loss computation, while the model is still incentivized to generate the correct reasoning markers. The DPO loss is then calculated as:
\begin{equation}
L_{\text{DPO}} = -\log \sigma\left(\beta \cdot (\mathbf{p}_{\text{masked, chosen}} - \mathbf{p}_{\text{masked, rejected}})\right),
\end{equation}
where \(\beta\) is a temperature parameter, \(\mathbf{p}_{\text{masked, chosen}}\) are the masked log-probabilities for the chosen response, and \(\mathbf{p}_{\text{masked, rejected}}\) are those for the rejected response.

Additionally, we introduce flexibility by allowing a fraction of the inner thinking tokens to be masked, controlled by a parameter \(\alpha\). For a sequence of \(n\) thinking tokens, \(\lfloor \alpha \cdot n \rfloor\) tokens are randomly selected for masking, where \(0 \leq \alpha \leq 1\). Setting \(\alpha = 0\) results in no masking, while \(\alpha = 1\) masks all inner thinking tokens.

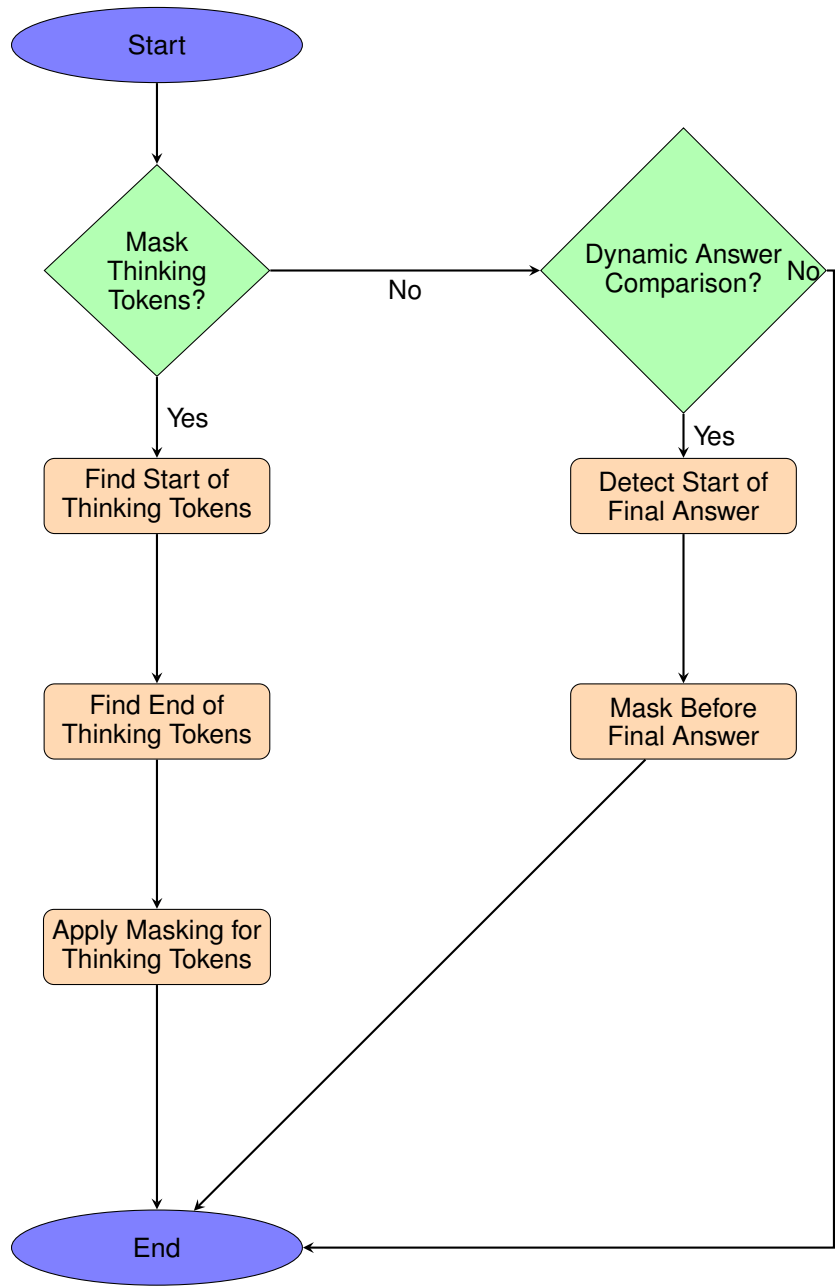
\begin{figure}[ht]
\sffamily % Switch to sans-serif font
\sansmath % Ensure math in flowchart is also sans-serif
\centering
\begin{tikzpicture}[node distance=3cm, every node/.style={align=center},
    startstop/.style={ellipse, draw, fill=blue!50, text width=2.5cm, text centered, minimum height=1cm},]
    % Nodes

    \node (start) [startstop] {Start};
    \node (maskThinking) [decision, below of=start] {Mask \\ Thinking \\ Tokens?};
    \node (dynamicAnswerComp) [decision, right of=maskThinking, xshift=4cm] {Dynamic Answer \\ Comparison?};
    
    % Thinking Tokens Masking Logic
    \node (startThink) [process, below of=maskThinking] {Find Start of \\ Thinking Tokens};
    \node (endThink) [process, below of=startThink] {Find End of \\ Thinking Tokens};
    \node (applyMask) [process, below of=endThink] {Apply Masking for \\ Thinking Tokens};
    
    % Dynamic Answer Comparison
    \node (finalAnswerStart) [process, below of=dynamicAnswerComp] {Detect Start of \\ Final Answer};
    \node (applyAnswerMask) [process, below of=finalAnswerStart] {Mask Before \\ Final Answer};
    
    % End node
    \node (end) [startstop, below of=applyMask, yshift=-1cm] {End};
    
    % Paths
    \draw [arrow] (start) -- (maskThinking);
    \draw [arrow] (maskThinking) -- node[anchor=west] {Yes} (startThink);
    \draw [arrow] (maskThinking) -- node[anchor=north] {No} (dynamicAnswerComp);
    
    \draw [arrow] (startThink) -- (endThink);
    \draw [arrow] (endThink) -- (applyMask);
    \draw [arrow] (applyMask) -- (end);
    
    \draw [arrow] (dynamicAnswerComp) -- node[anchor=west] {Yes} (finalAnswerStart);
    \draw [arrow] (finalAnswerStart) -- (applyAnswerMask);
    \draw [arrow] (applyAnswerMask) -- (end);
    
    % No paths
    \draw [arrow] (dynamicAnswerComp) -- node[anchor=east] {No} ++(2,0) |- (end);
\end{tikzpicture}
\caption{Flowchart showing how thinking tokens are masked and dynamic answer comparison is applied. If thinking tokens are masked, the start and end are identified and masked accordingly. If dynamic answer comparison is enabled, masking occurs before the final answer. All experiments done as part of the study reported in this paper used only one thinking section, which is equivalent to the ``Dynamic Answer Comparison'' approach where we mask all content before the final answer. All tokens within the thinking start/end tokens are masked, whereas the thinking start/end tokens are provided to the model to trigger it to use that particular process.}
\label{fig:masking_thinking_tokens}
\end{figure}

Figure~\ref{fig:masking_thinking_tokens} shows a flowchart that explains the process when all thinking tokens are masked, either in multiple thinking sections or all thinking tokens before the answer is developed.

\begin{figure}[ht]
\centering
\sffamily % Switch to sans-serif font
\sansmath % Ensure math in flowchart is also sans-serif
\begin{tikzpicture}[node distance=2cm, every node/.style={align=center},
    startstop/.style={ellipse, draw, fill=blue!50, text width=2.5cm, text centered, minimum height=1cm},]
    % Nodes
    \node (start) [startstop] {Start};
    \node (findTokens) [process, below of=start] {Find Start and End \\ of Thinking Tokens};
    \node (unmatchedStarts) [process, below of=findTokens] {Track Unmatched \\ Start Tokens};
    \node (validRange) [decision, below of=unmatchedStarts] {Valid Range \\ for Masking?};
    \node (segmentLength) [process, below of=validRange, yshift=-0.3cm] {Calculate Segment \\ Length};
    \node (randomMask) [process, below of=segmentLength, yshift=-0.3cm] {Randomly Select Tokens \\ to Mask};
    \node (applyMask) [process, below of=randomMask, yshift=-0.3cm] {Apply Masking \\ to Selected Tokens};
    
    % End node
    \node (end) [startstop, below of=applyMask, yshift=-0.5cm] {End};
    
    % Paths
    \draw [arrow] (start) -- (findTokens);
    \draw [arrow] (findTokens) -- (unmatchedStarts);
    \draw [arrow] (unmatchedStarts) -- (validRange);
    \draw [arrow] (validRange) -- node[anchor=west] {Yes} (segmentLength);
    \draw [arrow] (segmentLength) -- (randomMask);
    \draw [arrow] (randomMask) -- (applyMask);
    \draw [arrow] (applyMask) -- (end);
    
    % No path from valid range
    \draw [arrow] (validRange.east) -- ++(2,0) |- (end.east) node[anchor=north] {No};
\end{tikzpicture}
\sffamily % Restore default font
\caption{Flowchart showing the process of applying partial masking to thinking tokens. In this algorithm, the model identifies the start and end of thinking tokens and randomly selects a subset of tokens within that range for masking.}
\label{fig:random_partial_masking}
\end{figure}
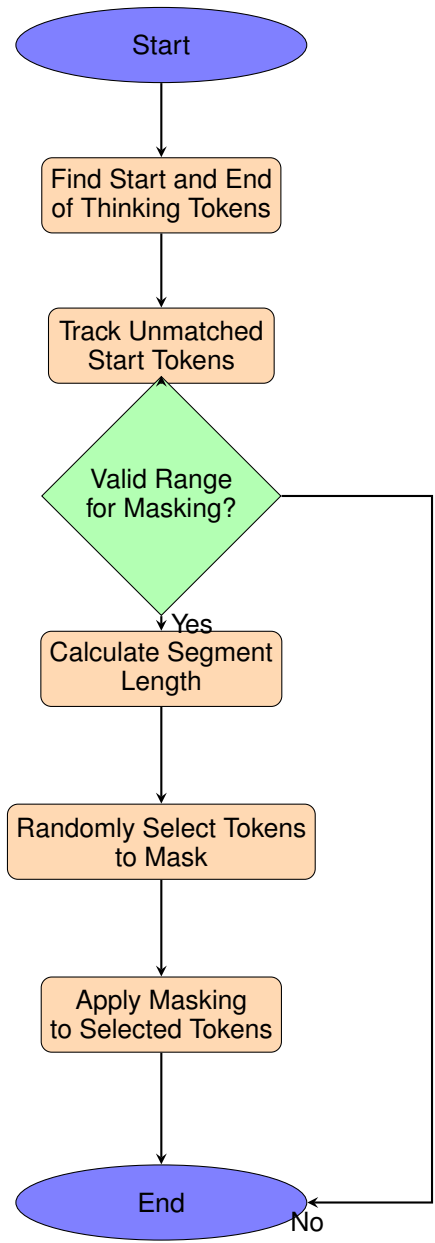

We also implemented an option where we mask only a fraction of the thinking tokens. Figure~\ref{fig:random_partial_masking} depicts the corresponding flowchart. The model identifies the start and end of thinking tokens, determines the valid range, and randomly selects a subset of tokens within that range for masking. In the masking logic, the \textit{valid range for masking} refers to the sequence of tokens that lies between the \texttt{<|thinking|>} and \texttt{<|/thinking|>} tokens. This range is where the model applies partial masking, optionally excluding the actual \texttt{<|thinking|>} and \texttt{<|/thinking|>} tokens themselves. By defining this valid range, the model can selectively mask a subset of tokens within the reasoning process, ensuring that only parts of the reasoning are obscured while the overall structure remains intact. This helps the model learn to handle incomplete reasoning segments. 
Additionally, the algorithm tracks \textit{unmatched start tokens}, which ensures that every \texttt{<|/thinking|>} token has a corresponding \texttt{<|thinking|>} token. If an \texttt{<|/thinking|>} token is found without a matching \texttt{<|thinking|>} token, the masking operation is not applied for that segment, preventing incorrect masking. This tracking mechanism guarantees that masking only occurs within valid reasoning sequences, ensuring consistency in handling incomplete or invalid token sequences.

\subsubsection{Dynamic Final Answer Detection via Masked Thinking Tokens}
To provide the model with more flexibility in producing variable-length thinking periods, we introduce a \textit{dynamic final answer detection} approach. Instead of masking tokens, this approach dynamically identifies the final answer by detecting the last occurrence of the \texttt{<|/thinking|>} token in each sequence. Let \(\mathbf{t}_{\text{end}}\) denote the position of the last \texttt{<|/thinking|>} token in the sequence. The final answer is defined as the sequence of tokens starting immediately after this position:
\begin{equation}
\mathbf{t}_{\text{final}} = [t_{\mathbf{t}_{\text{end}} + 1}, t_{\mathbf{t}_{\text{end}} + 2}, \dots, t_n].
\end{equation}
Log-probabilities for the final answer are computed as:
\begin{equation}
\mathbf{p}_{\text{final}} = [p_{\mathbf{t}_{\text{end}} + 1}, p_{\mathbf{t}_{\text{end}} + 2}, \dots, p_n].
\end{equation}
The DPO loss is then calculated based only on the log-probabilities for the final answers, as:
\begin{equation}
L_{\text{DPO}} = -\log \sigma\left(\beta \cdot (\mathbf{p}_{\text{final, chosen}} - \mathbf{p}_{\text{final, rejected}})\right).
\end{equation}
This approach allows the model to produce reasoning steps of variable lengths while ensuring that the comparison focuses solely on the final answer. As in the other method, the last end thinking token is optionally excluded from the comparison.

\subsubsection{Comparison of Masking Approaches}
The two approaches—masking of thinking tokens and dynamic final answer detection—provide complementary mechanisms for handling reasoning steps during training. Masking focuses on excluding inner thinking tokens while ensuring the model learns to produce the start and end reasoning markers, whereas dynamic detection provides more flexibility by ignoring the intermediate reasoning tokens altogether and focusing on the final output. These methods are toggled via the \texttt{dynamic\_answer\_comparison} flag in our implementation, offering the flexibility to experiment with different strategies for handling reasoning steps and penalizing incomplete responses.

\subsection{Optimizing Final Answers with Masked Reasoning Using Mode-Seeking Preference Alignment}

We employed the Efficient Exact Optimization (EXO) method~\cite{ji2024efficientexactoptimizationlanguage_EXO} in the special case where \( K = 2 \) optimizing the model to align with the ground truth answers while masking intermediate reasoning tokens. While masking these tokens, we observed significant improvements performance when using EXO compared to Direct Preference Optimization (DPO). We additionally applied \textit{label smoothing} with a smoothing factor of \( 5 \times 10^{-3} \) and set the scaling parameter \( \beta \) to 0.1. We use a learning rate of \( 5 \times 10^{-7} \) and a maximum grad norm of 0.3. We run one epoch of training each time a new on-the-fly dataset is generated, each featuring a set of 50 samples.  

The EXO method minimizes the reverse Kullback-Leibler (KL) divergence, which promotes a mode-seeking behavior in the model. Mode-seeking ensures that the model focuses on generating the most likely and preferred answers based on the training data, prioritizing reasoning paths that consistently lead to correct outcomes. This is in contrast to DPO’s mean-seeking behavior, which balances between multiple potential reasoning paths, often resulting in more generalized and less effective answers. 

In the special case of \( K = 2 \), the EXO loss is expressed as:
\begin{equation}
L_{\text{EXO}}(\pi_\theta) = \mathbb{E}_{(x, y_w, y_l) \sim D_{\text{pref}}} \left[ D_{\text{KL}}\left( p_\theta \| p_{\text{r}} \right) \right]
\end{equation}
where \( p_\theta \) and \( p_r \) represent the empirical distributions from the learned policy and the preference model, respectively, over the two completions: \( y_w \) (the ground truth answer) and \( y_l \) (the rejected answer predicted by the current model being trained). The scaling parameter \( \beta \) (set to 0.1) controls the strength of the preference alignment.

We mask the reasoning tokens during training, which prevents the model from directly observing how to reason through problems. However, the model still learns to infer the likely reasoning pathways based on the surrounding context and the final answer. EXO’s mode-seeking behavior plays a crucial role here, as it ensures that the model fills in the masked thinking tokens by inferring the most effective reasoning patterns that align with successful final answers. This is crucial because even though the thinking tokens are hidden, EXO thereby encourages the model to predict the reasoning that most often leads to the correct solution.

By focusing on the final answer during training, EXO effectively optimizes the entire reasoning process by indirectly learning which reasoning paths are most effective. It does so by leveraging the preference between the ground truth answer (\( y_w \)) and the rejected answer (\( y_l \)). The model learns to generate answers that are more likely to match the preferred reasoning processes by inferring and prioritizing the key modes in the training data, even when reasoning is masked. The use of label smoothing further regularizes the model by preventing it from being overly confident in its predictions, encouraging smoother distributional outputs.

\subsubsection{EXO vs. DPO: Impact on Final Answers}

In comparison to DPO, which minimizes the forward KL divergence and thus adopts a mean-seeking approach, EXO ensures that the model focuses on the most dominant reasoning patterns that lead to correct answers. DPO's mean-seeking behavior leads the model to spread probability mass across multiple reasoning paths, which can dilute its performance. In contrast, EXO concentrates on the most likely and correct answers, ignoring less effective or outlier reasoning paths, thus leading to more precise and higher-quality answers.

This is especially relevant in our case, where the final answer is prioritized in training. While the thinking tokens are masked, the final answer provides the key signal for optimizing the model’s reasoning capabilities. EXO helps the model learn to predict the most likely and effective reasoning paths that lead to that final answer, ensuring strong alignment with ground truth reasoning patterns, even without direct observation.

The use of EXO in our training, combined with label smoothing and the chosen \(\beta\) value, led to superior alignment with the ground truth, particularly in scenarios where final answer accuracy was prioritized over intermediate reasoning visibility with overall better long-term training performance and less overfitting. 

\subsection{Iterative Response Improvement using Thinking and Reflection model}

We implemented a recursive response algorithm that leverages the reasoning model to generate, improve, and integrate responses to a given question. This approach aims to produce more refined, comprehensive, and well-reasoned answers through iterative improvement and integration of multiple perspectives, using two LLM agents (see Figure~\ref{fig:PreFlexOR_recursive} for a detailed flowchart).

\begin{LLMbox}{Basic structure of the reasoning strategy with thinking and reflection tokens.}
\begin{lstlisting}
function recursive_response(question, N, reasoning_model, critic_model):
    initial_response = generate_response(reasoning_model, question)
    responses = [initial_response]
    
    for i in range(N):
        thinking = extract_thinking(responses[i])
        reflection = extract_reflection(responses[i])
        
        improved_thinking = improve_thinking(critic_model, thinking, reflection)
        new_response = generate_response(reasoning_model, question, improved_thinking)
        responses.append(new_response)
    
    final_answer = integrate_responses(critic_model, responses)
    return final_answer

where:
- generate_response(): Produces a response with thinking and reflection
- extract_thinking(), extract_reflection(): Parse the response
- improve_thinking(): Enhances the thinking process based on reflection
- integrate_responses(): Combines all responses into a final answer
\end{lstlisting}
\end{LLMbox}

The algorithm operates using two distinct LLM agents:

\begin{enumerate}
    \item \textbf{Reasoning Model (Agent 1):} This is a fine-tuned model specifically designed for advanced reasoning capabilities. It generates responses that include both a thinking process and a reflection on that process. The thinking process represents the model's step-by-step reasoning approach to answering the question, while the reflection component critically evaluates this thinking process.

    \item \textbf{Critic Model (Agent 2):} This is a non-fine-tuned, general-purpose language model. Its role is twofold:
    \begin{itemize}
        \item To improve the thinking process based on the reflections provided by the reasoning model.
        \item To integrate all generated responses into a final, comprehensive answer.
    \end{itemize}
\end{enumerate}

The reasoning model is the trained model, and the critic model was delineated as \url{meta-llama/Llama-3.2-3B-Instruct}. 

The algorithm proceeds as follows:

\begin{enumerate}
    \item The reasoning model generates an initial response to the question, including both a thinking process and a reflection on that process.
    
    \item For a specified number of iterations (N):
    \begin{itemize}
        \item The thinking process and reflection are extracted from the most recent response.
        \item The critic model analyzes the reflection and generates an improved version of the thinking process.
        \item The reasoning model then generates a new response using this improved thinking process as a guide.
        \item A new reflection is extracted from this response for use in the next iteration.
    \end{itemize}
    
    \item After all iterations are complete, the critic model performs a final integration step. It combines all generated responses into a comprehensive answer, leveraging the diverse perspectives and improvements made throughout the iterative process.
\end{enumerate}

This iterative approach, leveraging the strengths of both a specialized reasoning model and a general-purpose critic model, allows for continuous refinement of the reasoning process. The fine-tuned reasoning model provides structured, thoughtful responses, while the critic model offers a fresh perspective for improvement and integration. This combination potentially leads to more nuanced, well-considered, and comprehensive final responses.

The use of thinking and reflection tokens (<|thinking|>, <|/thinking|>, <|reflect|>, <|/reflect|>) allows for clear delineation and extraction of different components of the reasoning process. This structured approach facilitates the targeted improvement of the thinking process in each iteration.

By separating the roles of response generation (reasoning model) and response improvement/integration (critic model), the algorithm benefits from both specialized reasoning capabilities and general language understanding. This separation also allows for potential future improvements or replacements of either model independently, enhancing the flexibility and scalability of the approach.

\subsubsection{Analysis of Iterative Result Quality}

We use \url{gpt-4o} to analyze the quality of the text provided in each of the iterations, here $i=0$..$i=2$. The prompt used was:

\begin{LLMbox}{Basic structure of the reasoning strategy with thinking and reflection tokens.}
\begin{lstlisting}
[RAW OUTPUT FOR EACH ITERATION, MARKED WITH i=0, i=1, ...]

The three versions I provided above are iterative refinements of a LLM response due to iterative reasoning. Can you conduct a careful analysis of the three versions i=0. 1 and 2, assign a score, and give me a python code to plot the results in a bar chart. save as SVG file with time stamp.

The question asked was: [QUESTION]

Analyze the quality and other measures of writing, such as coherency or accuracy. Analyze each response on its own as it is not clear which is best.
\end{lstlisting}
\end{LLMbox}

The categories identified by \url{gpt-4o} were:
\begin{enumerate}
    \item Coherency: Measures the logical flow and structure of the response.
    \item Accuracy: How well the response aligns with established scientific facts regarding biological materials and failure mechanisms.
    \item Depth of Explanation: The level of detail and technical understanding conveyed in the response.
    \item Clarity: How easily understandable the explanation is, especially in communicating complex scientific ideas.
\end{enumerate}

\subsection{Training Stages}

The first training stage, using ORPO, is conducted using a Low-Rank Adaptation (LoRA)~\cite{Hu2021LoRA:Models} mechanism with a rank of 64 and an $\alpha$ value of 64. This design is applied to the following layers of the transformer architecture: \texttt{q\_proj}, \texttt{k\_proj}, \texttt{v\_proj}, \texttt{o\_proj}, \texttt{gate\_proj}, \texttt{up\_proj}, \texttt{down\_proj}, \texttt{embed\_tokens}, and \texttt{lm\_head}. The LoRA adapters enable efficient fine-tuning by introducing trainable low-rank matrices, allowing us to adjust the model parameters while significantly reducing the number of trainable parameters compared to full model fine-tuning. This approach enhances the model's ability to capture task-specific knowledge while preserving its overall structure. Once the initial training stage is completed, the adapter is merged into the base model. 

In the second training stage, we create a new adapter (built on top of the resulting model obtained after the first stage). We use a  rank of 64 and an $\alpha$ value of 64, applied to the following layers of the transformer architecture: \texttt{q\_proj}, \texttt{k\_proj}, \texttt{v\_proj}, \texttt{o\_proj}, \texttt{gate\_proj}, \texttt{up\_proj}, and \texttt{down\_proj}  (in this phase, the embedding and head layers are excluded). 

The second phase focuses less on the intermediate reasoning steps and more on producing accurate final answers. Instead of explicitly guiding the model through the reasoning process, this stage lets the model figure out the best reasoning paths on its own, using the knowledge and structures learned during the first phase. The EXO-based preference alignment optimizes the final answer, allowing the model to concentrate on mode-seeking behavior, where it prioritizes the most likely and correct answer based on learned reasoning strategies. By focusing on the outcome, this phase ensures the model produces high-quality answers while letting it autonomously handle reasoning complexity.

Thus, the combination of ORPO in the first stage for learning basic reasoning, and preference optimization via EXO in the second stage for refining answer accuracy, results in a robust, multi-phase training process that improves both the model’s reasoning abilities and its final outputs.

The training algorithms are implemented on top of the Hugging Face Transformer Reinforcement Learning (TRL) library~\url{https://huggingface.co/docs/trl/en/index}, specifically using inherited classes from the ORPO and DPO trainers. 

\subsection{Brief recap of Key Concepts in Hesse's novel The Glass Bead Game}

Hermann Hesse's ``The Glass Bead Game''~\cite{hesse2000glass,Ziolkowski1965} is a novel that envisions a future society where intellectual pursuit and the synthesis of knowledge are paramount. Central to the story is the province of \textit{Castalia}, a secluded community dedicated to scholarly excellence and the mastery of the Glass Bead Game. The Game itself is an abstract, highly intellectual practice that symbolically combines elements from various disciplines—such as music, mathematics, literature, and philosophy—into a cohesive and harmonious whole. 

Players use a sophisticated symbolic language to explore and reveal the underlying connections between disparate fields of knowledge, embodying ideals of interdisciplinary synthesis and deep contemplation. The novel reflects philosophical traditions that emphasize the unity of knowledge and the interconnectedness of all things, resonating with concepts from Neoplatonism, which posits a single underlying reality, and Eastern philosophies like Taoism and Zen Buddhism, which highlight the harmony and balance inherent in the universe. Hesse's work invites reflection on the role of intellectualism in society, the pursuit of enlightenment, and the balance between contemplation and engagement with the material world. 

While the precise mechanics of the Glass Bead Game are intentionally left abstract by Hesse, within the novel it is portrayed as a highly formalized and complex practice. The Game involves the creation and manipulation of symbols that represent ideas from various disciplines, structured in a way that reveals their underlying relationships and harmonies. Players, who have devoted years to study and preparation, engage in elaborate sessions where they compose and interpret these symbolic sequences, often accompanied by meditative reflection. The highest authority in the Game is the \textit{Magister Ludi} (Latin for ``Master of the Game''), who oversees the advancement and integrity of the Game within Castalia. The Magister Ludi is responsible for guiding the community of players, fostering the development of the Game's symbolic language, and ensuring that the practice remains aligned with the ideals of wisdom, intellectual purity, and the unification of knowledge that the Game embodies.

By paralleling the vision in the Glass Bead Game, where the synthesis of all knowledge transforms human understanding, the rise of AI and complex reasoning engines is dramatically expanding the scope of human intellect, shifting our role from mere information processors to interpreters, ethical stewards, and collaborators with intelligent systems. This evolution challenges us to redefine our intellectual boundaries and embrace the profound impact of AI on the future of human thought, creativity, and the collective pursuit of wisdom.

\subsection{Relation between Preference Optimization and Reinforcement Learning Strategies}

In the Direct Preference Optimization (DPO) setup, we typically optimize based on preference comparisons of chosen versus rejected outputs. However, by generating training data randomly at every step and dynamically constructing reference reasoning, the model's training environment begins to incorporate elements of meta-learning. This randomization introduces exploration, adaptability, and generalization, making the process more akin to meta-learning, where the model is trained not just to solve a specific task, but to learn how to solve tasks in general.

Meta-learning (``learning to learn'') aims to develop models that can quickly adapt to new tasks by leveraging prior knowledge. In this context, by generating random training data and reference reasoning at each training step, we are effectively training the model on a distribution of tasks. Each task (random reasoning structure) represents a variation in reasoning, and the model learns not just how to answer specific reasoning problems, but how to reason in general.

One of the key principles of meta-learning is task distribution. Rather than a fixed, static dataset, the model encounters a wide variety of reasoning challenges, each randomly generated. This encourages the model to develop a generalized reasoning framework that can adapt quickly to new tasks, mirroring the concept of few-shot learning in meta-learning.

Another significant aspect is exploration of reasoning paths. Since the reference reasoning is constructed dynamically, the model must learn to explore different reasoning strategies each time. This exploration forces the model to generalize better because it cannot rely on memorizing patterns from repeated examples. This is similar to reinforcement learning (RL) where the agent explores different actions and learns from the consequences of those actions. In this DPO setup, however, the exploration happens in the space of reasoning paths rather than state-action sequences, and the reward signal is in the form of preference alignment rather than cumulative rewards.

By incorporating partial masking in the reasoning process, the model is trained to handle incomplete information. This is similar to RL agents making decisions with partial observability. The masking challenges the model to optimize for the final answer despite missing parts of the reasoning. It also encourages the model to focus more on the outcome (the preferred answer) rather than the exact steps taken to reach that answer. This aligns well with the preference optimization framework, which optimizes based on output preference rather than the reasoning steps themselves.

Moreover, the randomization of data and reasoning structures encourages the model to develop a meta-strategy for reasoning. Instead of learning a fixed method for solving a specific task, the model learns to adapt to new reasoning patterns dynamically. This meta-strategy helps the model generalize across a variety of reasoning tasks, much like how a meta-learning model generalizes across different tasks by learning higher-level patterns and strategies.

By generating data and reference reasoning on-the-fly, we are also introducing a form of exploration that is typically seen in reinforcement learning. In RL, exploration is necessary for the agent to discover optimal policies, while in this DPO setup, the exploration arises from the random selection of reasoning data and paths, requiring the model to adapt its reasoning strategy each time. The difference, however, is that DPO focuses on preference optimization rather than cumulative rewards over time.

Furthermore, this dynamic and random training process allows for fast adaptation, a key element in meta-learning. The model is forced to adapt quickly to new tasks (reasoning paths) and make decisions based on incomplete or masked information. This is akin to the few-shot learning scenario in meta-learning, where models must learn to perform tasks with minimal data. In this case, the model learns to reason effectively even when part of the reasoning process is hidden.

Finally, the overall structure of this training setup, with dynamic data generation and reasoning masking, creates a meta-learning-like environment where the model develops the ability to handle random and diverse reasoning tasks without overfitting to any specific pattern. The model is learning not just to perform well on a single task, but to adapt quickly and generalize across a wide range of reasoning challenges, which is the core goal of meta-learning. Our experiments have confirmed that this was indeed the case. 

\subsubsection{Modification to Further Align with Reinforcement Learning}

To modify this method further and make it more similar to RL, the process of generating random training data could be formalized as a sequential decision-making problem, where each step in the reasoning process is treated as an action, and the sequence of reasoning steps forms a trajectory. In this scenario, new random training data would be constructed as a result of decisions made by the model in a feedback loop, where each reasoning step influences the next.

For instance, rather than randomly selecting data for each training step, the model could be given the ability to actively select which reasoning path or data point to explore based on a learned policy. This would involve assigning a reward signal based on how well the reasoning aligns with human preferences at the end of the trajectory. The model would be rewarded not only for generating the correct final output, but also for choosing intermediate reasoning steps that lead to better outcomes. This reward could be delayed, encouraging the model to think strategically about the long-term effects of its reasoning choices, much like in RL, where the agent learns to maximize cumulative rewards over time.

Moreover, the random generation of data could be made contingent on past reasoning steps, transforming the reasoning process into an interactive environment where the model’s actions (reasoning decisions) impact the data it encounters next. In this way, the reasoning process becomes more dynamic, with the model continually adjusting its path based on feedback from previous steps, which aligns closely with the exploration and exploitation trade-off seen in RL.

\section*{Author contributions}
MJB designed the overall research, developed algorithms, codes and conducted the training, assessments and analysis. MJB developed and executed the distillation strategies to generate structured and adversarial data.  

\section*{Conflicts of interest}
The author declares no conflicts of interest.

\section*{Code, data and model weight availability}

Training scripts are available at \url{https://github.com/lamm-mit/PRefLexOR}. Model weights can be found at \url{https://huggingface.co/lamm-mit}.

\section*{Acknowledgments}
This work was supported in part by Google, the MIT Generative AI Initiative, with additional support from NIH. 

%Bibliography
%\bibliographystyle{unsrt}  
\bibliographystyle{naturemag}
\bibliography{references,references-Mendeley}

\clearpage

\textboxcaption{Inference example, thinking and answer, for a question related to the role of hierarchical structures. }
\label{TexBox:thinking_and_reflection_Box_10}

\begin{LLMboxmultipage} {Assessment of reasoning and response with thinking tokens.}
\begin{lstlisting}

(*@\bf\hlred{User:}@*) Tell me why hierarchical structures work so well. Use <|thinking|>.

(*@\bf\hllightgreen{Assistant:}@*) <|thinking|>
**Reasoning Steps**:

Hierarchical structures work well due to their ability to optimize properties at multiple length scales. This is achieved by combining simpler, lower-level components in a hierarchical manner, allowing for:
1. **Size and Shape Adaptation**: Hierarchical structures can adapt to different environments and tasks by changing their size and shape, enhancing performance.
2. **Material Utilization**: Hierarchical structures enable the use of materials with different properties at different levels, maximizing overall performance.
3. **Energy Dissipation**: Hierarchical structures can effectively dissipate energy through mechanisms like crack bridging and deflection, improving toughness and fracture resistance.
4. **Structural Reinforcement**: Hierarchical structures can reinforce the material by providing additional support and stability, enhancing mechanical properties.
5. **An
6. **Energy Absorption**: Hierarchical structures can absorb energy through mechanisms like plastic deformation and fracture, improving impact resistance.
7. **Thermal Insulation**: Hierarchical structures can provide better thermal insulation by trapping air and minimizing heat transfer, enhancing thermal performance.
8. **Mechanical Strength**: Hierarchical structures can enhance mechanical strength by distributing stress and improving toughness, making them suitable for applications requiring high strength and toughness.

**Relevant Materials or Concepts**:

Hierarchical structures work well due to their ability to optimize properties at multiple length scales by combining simpler, lower-level components in a hierarchical manner. This allows for size and shape adaptation, material utilization, energy dissipation, structural reinforcement, and thermal insulation, enhancing overall performance.

**Design Principles**:

Hierarchical structures work well due to their ability to optimize properties at multiple length scales by combining simpler, lower-level components in a hierarchical manner. This allows for size and shape adaptation, material utilization, energy dissipation, structural reinforcement, and thermal insulation, enhancing overall performance.

**Material Properties**:

Hierarchical structures work well due to their ability to optimize properties at multiple length scales by combining simpler, lower-level components in a hierarchical manner. This allows for size and shape adaptation, material utilization, energy dissipation, structural reinforcement, and thermal insulation, enhancing overall performance.

**Hypothesis**:

"Hierarchical structures work well due to their ability to optimize properties at multiple length scales by combining simpler, lower-level components in a hierarchical manner."

**Additional Background**:

Hierarchical structures work well due to their ability to optimize properties at multiple length scales by combining simpler, lower-level components in a hierarchical manner. This approach allows for size and shape adaptation, which is crucial for applications requiring flexibility and adaptability. Hierarchical structures can also enhance material properties by utilizing different materials at various levels, which improves mechanical strength and toughness. Additionally, they can provide better thermal insulation by trapping air and minimizing heat transfer, which is beneficial for energy efficiency. This hierarchical design enables the creation of complex structures that mimic natural systems, which often exhibit superior performance due to their ability to adapt and respond to environmental changes.

<|/thinking|>
Hierarchical structures work exceptionally well due to their ability to optimize properties at multiple length scales by combining simpler, lower-level components in a hierarchical manner. This hierarchical design allows for several key advantages that contribute to their superior performance:

1. **Size and Shape Adaptation**: Hierarchical structures can adapt to different environments and tasks by changing their size and shape. This is crucial for applications requiring flexibility and adaptability, such as in biological systems and soft materials.

2. **Material Utilization**: Hierarchical structures enable the use of materials with different properties at different levels. This maximizes the overall performance of the material by leveraging the unique properties of each level. For instance, in nacre, the aragonite tablets are stiff and hard, while the protein matrix is soft and flexible, allowing for excellent toughness and fracture resistance.

3. **Energy Dissipation**: Hierarchical structures can effectively dissipate energy through mechanisms like crack bridging and deflection. This improves toughness and fracture resistance, as seen in the case of nacre, where the aragonite tablets help bridge cracks and prevent their propagation.

4. **Structural Reinforcement**: Hierarchical structures can reinforce the material by providing additional support and stability. This enhances mechanical properties, as observed in the case of nacre, where the aragonite tablets provide structural reinforcement.

5. **Thermal Insulation**: Hierarchical structures can provide better thermal insulation by trapping air and minimizing heat transfer. This is beneficial for applications requiring high thermal performance, such as in thermal insulation materials.

6. **Mechanical Strength**: Hierarchical structures can enhance mechanical strength by distributing stress and improving toughness. This makes them suitable for applications requiring high strength and toughness, such as in bone and nacre.

In the context of nacre, these hierarchical structures work exceptionally well due to the combination of stiff aragonite tablets and soft protein matrix. This hierarchical design allows nacre to exhibit excellent toughness and fracture resistance, making it a model for designing materials with high strength and toughness.

In conclusion, hierarchical structures work well due to their ability to optimize properties at multiple length scales by combining simpler, lower-level components in a hierarchical manner.
\end{lstlisting}
\end{LLMboxmultipage}

\clearpage

\textboxcaption{Example of failure analysis. }
\label{TexBox:thinking_and_reflection_Box_11}

\begin{LLMboxmultipage}{Assessment of reasoning and response with thinking tokens.}
\begin{lstlisting}

(*@\bf\hlred{User:}@*) How do biological materials fail gracefully? Use <|thinking|>.

(*@\bf\hllightgreen{Assistant:}@*) <|thinking|>
**Reasoning Steps**:

1. Biological materials fail gracefully by dissipating energy through plasticity and crack propagation, rather than catastrophic failure.
2. This is achieved by mechanisms like viscoelasticity, crack bridging, fiber pullout, and fiber sliding, which occur at the microstructural level.
3. These mechanisms allow materials to absorb energy and delay fracture, preventing sudden collapse.
4. This gradual failure process enables biological materials to maintain structural integrity and functionality even under stress.
5. The hierarchical structure of biological materials, with features like mineralized collagen fibers and bone lamellae, enhances their ability to dissipate energy and resist fracture.
6. This hierarchical structure allows for localized energy absorption and stress redistribution, further contributing to the materials' ability to fail gracefully.

**Relevant Materials or Concepts**:

The relevant materials or concepts for understanding how biological materials fail gracefully are:
- **Plasticity and Crack Propagation**: These mechanisms allow biological materials to absorb energy and delay fracture, enabling them to fail gradually.
- **Viscoelasticity**: This property enables biological materials to exhibit time-dependent viscoelastic behavior, further contributing to their ability to dissipate energy and resist fracture.
- **Crack Bridging**: This mechanism involves the bridging of cracks by fibers or other structural elements, preventing crack propagation and allowing materials to absorb more energy before failing.
- **Fiber Pullout**: This occurs when fibers are pulled out of their matrix, dissipating energy and delaying fracture.
- **Fiber Sliding**: This involves the sliding of fibers relative to each other, which also contributes to energy dissipation and gradual failure.
- **Hierarchical Structure**: The hierarchical structure of biological materials, with features like mineralized collagen fibers and bone lamellae, enhances their ability to dissipate energy and resist fracture by allowing for localized energy absorption and stress redistribution.

**Design Principles**:

Biological materials fail gracefully by dissipating energy through plasticity and crack propagation, rather than catastrophic failure. This is achieved by mechanisms like viscoelasticity, crack bridging, fiber pullout, and fiber sliding, which occur at the microstructural level. The hierarchical structure of biological materials, with features like mineralized collagen fibers and bone lamellae, enhances their ability to dissipate energy and resist fracture by allowing for localized energy absorption and stress redistribution.

**Material Properties**:

Biological materials fail gracefully by dissipating energy through mechanisms such as viscoelasticity, crack bridging, fiber pullout, and fiber sliding, which occur at the microstructural level. These mechanisms allow for gradual energy absorption and stress redistribution, enhancing the materials' ability to resist fracture and maintain structural integrity.

**Hypothesis**:

"Biological materials fail gracefully by dissipating energy through plasticity and crack propagation, rather than catastrophic failure, through mechanisms like viscoelasticity, crack bridging, fiber pullout, and fiber sliding."

**Additional Background**:

Biological materials fail gracefully by utilizing mechanisms such as viscoelasticity, crack bridging, and fiber pullout, which are designed to dissipate energy gradually rather than leading to sudden failure. These mechanisms are often influenced by the hierarchical structure of biological materials, which allows for localized energy absorption and stress redistribution. This approach enables biological materials to maintain their structural integrity and functionality even under stress. The ability of these materials to fail gradually is also attributed to their intrinsic viscoelastic properties, which are time-dependent and allow for energy dissipation over time. This gradual failure process is crucial for the survival and function of biological systems, as it prevents sudden collapse and maintains structural integrity.

<|/thinking|>
Biological materials fail gracefully through a combination of intrinsic and extrinsic mechanisms that dissipate energy and delay fracture, rather than undergoing catastrophic failure. This is achieved through the following strategies:

1. **Viscoelasticity**: Many biological materials, like bone and cartilage, exhibit viscoelastic behavior. This means they can absorb and dissipate energy over time, similar to viscoelastic materials like rubber. This property allows them to gradually absorb stress and energy, preventing sudden collapse.

2. **Crack Propagation and Bridging**: Cracks in biological materials are often bridged by fibers or other structural elements. This mechanism prevents crack propagation and allows the material to absorb more energy before failing. For instance, in bone, cracks are bridged by mineralized collagen fibers.

3. **Fiber Pullout**: When fibers are pulled out of their matrix, energy is dissipated, and the material can continue to withstand stress. This mechanism is seen in materials like bone and nacre.

4. **Fiber Sliding**: The sliding of fibers relative to each other also contributes to energy dissipation. This is observed in materials like nacre and bone.

5. **Hierarchical Structure**: The hierarchical structure of biological materials, with features like mineralized collagen fibers and bone lamellae, enhances their ability to dissipate energy and resist fracture. This structure allows for localized energy absorption and stress redistribution, further contributing to the materials' ability to fail gracefully.

6. **Time-Dependent Properties**: Many biological materials exhibit time-dependent properties, such as viscoelasticity, which allow them to gradually absorb energy and resist fracture over time.

7. **Plasticity**: Biological materials often exhibit plasticity, which allows them to deform under stress without immediately failing. This plasticity helps to dissipate energy and delay fracture.

These mechanisms work together to enable biological materials to fail gradually, maintaining their structural integrity and functionality even under stress. This is crucial for the survival and function of biological systems, as sudden collapse would lead to catastrophic failure.
\end{lstlisting}
\end{LLMboxmultipage}

\clearpage

\textboxcaption{Example of combining thinking and reflection tokens, for a question at the intersection between philosophy and biological materials science. }
\label{TexBox:thinking_and_reflection_Box_12}

\begin{LLMboxmultipage}{Assessment of reasoning and response with thinking tokens.}
\begin{lstlisting}

(*@\bf\hlred{User:}@*) Explain the ideas that can connect Herrman Hesse's Glass Bead Game with proteins. First, review the concept and second, discuss relatiionships with proteins. Use <|thinking|>.

(*@\bf\hllightgreen{Assistant:}@*) <|thinking|>
**Reasoning Steps**:

1. **Hesse's Glass Bead Game**: A philosophical and artistic exploration of the nature of reality, focusing on the interconnectedness of art, science, and philosophy.
2. **Proteins**: Biological molecules with complex structures, crucial for various functions in living organisms.
3. **Connection Ideas**:
   - **Structural Complexity**: Proteins' intricate structures, like Hesse's glass beads, are influenced by chemical bonding and molecular interactions.
   - **Hierarchical Organization**: Proteins' hierarchical structures, from primary to quaternary levels, are similar to the hierarchical organization of Hesse's glass beads.
   - **Dynamic Nature**: Proteins' ability to change conformation and function, like the dynamic nature of Hesse's game, reflects the dynamic nature of reality.
   - **Interconnectedness**: Proteins' roles in various biological processes, like protein-protein interactions, mirror the interconnectedness of art, science, and philosophy in Hesse's game.
   - **Evolutionary Adaptation**: Proteins' ability to evolve and adapt, like the evolution of art and science, reflects the dynamic and adaptive nature of reality.

**Relevant Materials or Concepts**:

The relevant materials or concepts that connect Herrman Hesse's Glass Bead Game with proteins are:
- The structural complexity and hierarchical organization of proteins, which are similar to the intricate structures and hierarchical organization of Hesse's glass beads.
- The dynamic nature of proteins, which can change conformation and function, reflecting the dynamic nature of reality in Hesse's game.
- The interconnectedness of proteins in various biological processes, such as protein-protein interactions, which mirrors the interconnectedness of art, science, and philosophy in Hesse's game.
- The evolutionary adaptation of proteins, which can evolve and adapt to new environments, reflecting the dynamic and adaptive nature of reality in Hesse's game.

**Design Principles**:

The design principles that connect Herrman Hesse's Glass Bead Game with proteins are:

1. **Structural Complexity**: Proteins, like Hesse's glass beads, exhibit intricate structures influenced by chemical bonding and molecular interactions.
2. **Hierarchical Organization**: Proteins, like Hesse's glass beads, have hierarchical structures, from primary to quaternary levels, which are crucial for their functions.
3. **Dynamic Nature**: Proteins, like the game, can change conformation and function, reflecting the dynamic nature of reality.
4. **Interconnectedness**: Proteins, like the game, play interconnected roles in various biological processes, such as protein-protein interactions, mirroring the interconnectedness of art, science, and philosophy in the game.
5. **Evolutionary Adaptation**: Proteins, like the game, can evolve and adapt to new environments, reflecting the dynamic and adaptive nature of reality in the game.

**Material Properties**:

The material properties relevant to the connection between Herrman Hesse's Glass Bead Game and proteins are:

1. **Structural Complexity**: Proteins exhibit intricate, hierarchical structures, similar to the complex, interconnected structures of Hesse's glass beads.
2. **Hierarchical Organization**: Proteins have a hierarchical organization, from primary to quaternary levels, which is similar to the hierarchical organization of Hesse's glass beads.
3. **Dynamic Nature**: Proteins can change conformation and function, reflecting the dynamic nature of reality in Hesse's game.
4. **Interconnectedness**: Proteins play interconnected roles in various biological processes, mirroring the interconnectedness of art, science, and philosophy in Hesse's game.
5. **Evolutionary Adaptation**: Proteins can evolve and adapt to new environments, reflecting the dynamic and adaptive nature of reality in Hesse's game.

**Hypothesis**:

"Proteins, with their intricate structures and hierarchical organization, and their dynamic nature, are analogous to the interconnected, hierarchical, and dynamic elements of Herrman Hesse's Glass Bead Game."

**Additional Background**:

Herrman Hesse's Glass Bead Game is a philosophical and artistic exploration that delves into the nature of reality, focusing on the interconnectedness of art, science, and philosophy. It involves a complex, hierarchical structure of glass beads, each representing a different aspect of reality, which are interconnected and interdependent. This structure is reminiscent of the hierarchical organization found in biological systems, where complex structures are composed of simpler ones. Proteins, as biological molecules, also exhibit hierarchical structures, from primary to quaternary levels, which are crucial for their functions. The dynamic nature of proteins, which can change conformation and interact with other molecules, parallels the dynamic and adaptive nature of reality in Hesse's game. This dynamic nature allows proteins to evolve and adapt to new environments, much like the evolution of art and science in the game. The interconnectedness of these elements suggests that both the game and proteins reflect the intricate, interconnected, and dynamic nature of reality.

<|/thinking|>
Herrman Hesse's Glass Bead Game and proteins share several interconnected ideas that highlight the intricate and dynamic nature of reality. Here's a comprehensive review of the concepts and their relationships:

**Hesse's Glass Bead Game:**

1. **Interconnectedness of Art, Science, and Philosophy**: The game explores the interconnectedness of these three disciplines, suggesting that they are not separate entities but rather different aspects of a unified reality.
2. **Hierarchical Structure**: The game's structure is hierarchical, with each level representing a different aspect of reality. This mirrors the hierarchical organization found in biological systems, where complex structures are composed of simpler ones.
3. **Dynamic Nature**: The game's structure is dynamic, with elements changing and adapting to each other, reflecting the dynamic nature of reality.
4. **Evolutionary Adaptation**: The game's structure evolves over time, much like the evolution of art and science, reflecting the dynamic and adaptive nature of reality.

**Proteins:**

1. **Structural Complexity**: Proteins are complex biomolecules with intricate structures, influenced by chemical bonding and molecular interactions.
2. **Hierarchical Organization**: Proteins have a hierarchical organization, from primary to quaternary levels, which is crucial for their functions.
3. **Dynamic Nature**: Proteins can change conformation and function, reflecting the dynamic nature of reality.
4. **Interconnectedness**: Proteins play interconnected roles in various biological processes, such as protein-protein interactions, mirroring the interconnectedness of art, science, and philosophy in the game.
5. **Evolutionary Adaptation**: Proteins can evolve and adapt to new environments, reflecting the dynamic and adaptive nature of reality in the game.

**Relationships between Hesse's Glass Bead Game and Proteins:**

1. **Structural Complexity and Hierarchical Organization**: Both the game and proteins exhibit intricate, hierarchical structures, with proteins having a more complex and dynamic hierarchy.
2. **Dynamic Nature**: Both the game and proteins can change and adapt, reflecting the dynamic nature of reality.
3. **Interconnectedness**: Both the game and proteins play interconnected roles, with proteins interacting with each other and their environment, and the game's elements interacting with each other.
4. **Evolutionary Adaptation**: Both the game and proteins can evolve and adapt to new environments, reflecting the dynamic and adaptive nature of reality.
5. **Philosophical Implications**: The interconnectedness and dynamic nature of the game and proteins reflect the philosophical ideas of Hesse, highlighting the importance of understanding the interconnectedness of art, science, and philosophy in understanding reality.

In conclusion, the ideas of Hesse's Glass Bead Game and proteins are interconnected and reflect the dynamic, hierarchical, and interconnected nature of reality. While the game is a philosophical and artistic exploration, proteins are biological molecules that exhibit similar structural and dynamic properties, highlighting the universality of these concepts across different domains.
\end{lstlisting}
\end{LLMboxmultipage}

\clearpage

\textboxcaption{Example of a summarization and research proposal development task, reflecting a new task beyond the ones the model had been trained on, to test generalization of reasoning capability. }
\label{TexBox:thinking_and_reflection_Box_13}

\begin{LLMboxmultipage}{Model response to an unknown task featuring summarization and research proposal development.}
\begin{lstlisting}

(*@\bf\hlred{User:}@*) Consider this abstract of a paper:

Fungal mycelium, a living network of filamentous threads, thrives on lignocellulosic waste and exhibits rapid growth, hydrophobicity, and intrinsic regeneration, offering a potential means to create next-generation sustainable and functional composites. However, existing hybrid-living mycelium composites (myco-composites) are tremendously constrained by conventional mold-based manufacturing processes, which are only compatible with simple geometries and coarse biomass substrates that enable gas exchange. Here we introduce a class of structural myco-composites manufactured with a novel platform that harnesses high-resolution biocomposite additive manufacturing and robust mycelium colonization with indirect inoculation. We leverage principles of hierarchical composite design and selective nutritional provision to create a robust myco-composite that is scalable, tunable, and compatible with complex geometries. To illustrate the versatility of this platform, we characterize the impact of mycelium colonization on mechanical and surface properties of the composite. We found that our method yields the strongest mycelium composite reported to date with a modulus of 160 MPa and tensile strength of 0.72 MPa, which represents over a 15-fold improvement over typical mycelium composites, and further demonstrate unique applications with fabrication of foldable bio-welded containers and flexible mycelium textiles. This study bridges the gap between biocomposite and hybrid-living materials research, opening the door to advanced structural mycelium applications and demonstrating a novel platform for development of diverse hybrid-living materials.

Summarize the results and propose future research ideas.  Use <|thinking|>.

(*@\bf\hllightgreen{Assistant:}@*) <|thinking|>
**Reasoning Steps**:

1. The study introduces a novel platform for manufacturing structural myco-composites using high-resolution biocomposite additive manufacturing and robust mycelium colonization with indirect inoculation.
2. This platform leverages principles of hierarchical composite design and selective nutritional provision to create scalable, tunable, and complex-geometry compatible myco-composites.
3. The study characterizes the impact of mycelium colonization on mechanical and surface properties of the composite, achieving a modulus of 160 MPa and tensile strength of 0.72 MPa, a 15-fold improvement over typical mycelium composites.
4. The research demonstrates unique applications, such as foldable bio-welded containers and flexible mycelium textiles, showcasing the versatility of the novel platform.
5. The study bridges the gap between biocomposite and hybrid-living materials research, opening the door to advanced structural mycelium applications and novel hybrid-living materials development.

**Relevant Materials or Concepts**:

The relevant materials or concepts for this study are:
- High-resolution biocomposite additive manufacturing
- Robust mycelium colonization with indirect inoculation
- Hierarchical composite design
- Selective nutritional provision
- Mechanical properties (modulus, tensile strength)
- Surface properties
- Complex geometries
- Scalable, tunable, and versatile myco-composites
- Hybrid-living materials research
- Advanced structural mycelium applications

**Design Principles**:

The study introduces a novel platform for manufacturing structural myco-composites using high-resolution biocomposite additive manufacturing and robust mycelium colonization with indirect inoculation. This platform leverages principles of hierarchical composite design and selective nutritional provision to create scalable, tunable, and complex-geometry compatible myco-composites. The key design principles are:

1. **High-Resolution Additive Manufacturing**: Utilize high-resolution biocomposite additive manufacturing to create complex geometries and precise structures.
2. **Robust Mycelium Colonization**: Employ indirect inoculation to ensure robust mycelium colonization, enabling the creation of scalable and tunable myco-composites.
3. **Hierarchical Composite Design**: Leverage principles of hierarchical composite design to optimize the mechanical and surface properties of the myco-composite.
4. **Selective Nutritional Provision**: Control the nutritional environment to tailor the growth and properties of the mycelium, enhancing the composite's performance.
5. **Complex Geometry Compatibility**: Design the myco-composite to be compatible with complex geometries, enabling the creation of versatile and functional materials.

**Material Properties**:

The study reports a modulus of 160 MPa and tensile strength of 0.72 MPa for the myco-composite, which is a 15-fold improvement over typical mycelium composites.

**Hypothesis**:

The hypothesis is that the novel platform for manufacturing structural myco-composites, leveraging high-resolution biocomposite additive manufacturing and robust mycelium colonization, can create scalable, tunable, and complex-geometry compatible myco-composites with superior mechanical and surface properties.

**Additional Background**:

The study focuses on the development of structural myco-composites using a novel platform that combines high-resolution biocomposite additive manufacturing with robust mycelium colonization. This approach allows for the creation of complex geometries and scalable, tunable materials. The research explores the impact of mycelium colonization on mechanical and surface properties, achieving a modulus of 160 MPa and tensile strength of 0.72 MPa, which is a significant improvement over typical mycelium composites. The study also highlights the versatility of the platform in fabricating foldable bio-welded containers and flexible mycelium textiles, demonstrating its potential for advanced applications.

<|/thinking|>
The study introduces a novel platform for manufacturing structural myco-composites, which combines high-resolution biocomposite additive manufacturing with robust mycelium colonization using indirect inoculation. This platform leverages principles of hierarchical composite design and selective nutritional provision to create scalable, tunable, and complex-geometry compatible myco-composites. The research demonstrates the versatility of this platform by characterizing the impact of mycelium colonization on mechanical and surface properties of the composite, achieving a modulus of 160 MPa and tensile strength of 0.72 MPa, a 15-fold improvement over typical mycelium composites. The study also showcases unique applications, such as foldable bio-welded containers and flexible mycelium textiles, highlighting the potential of this novel platform for advanced structural mycelium applications and novel hybrid-living materials development.

Future research directions based on this study could include:

1. **Scaling Up and Down**: Investigate the scalability of the myco-composite manufacturing process to produce larger, more complex structures, and explore the feasibility of scaling down to smaller, more intricate components.

2. **Material Properties Enhancement**: Explore ways to further enhance the mechanical and surface properties of the myco-composite, such as by introducing additional materials, modifying the mycelium colonization process, or using different biocomposite additive manufacturing techniques.

3. **Multifunctional Composites**: Investigate the potential of the myco-composite platform to create multifunctional composites with integrated properties, such as self-healing, conductivity, or optical properties, to enable novel applications.

4. **Biodegradability and Sustainability**: Study the biodegradability and sustainability of the myco-composite materials, including their potential for use in biomedical applications, waste management, and other environmentally friendly contexts.

5. **Hybrid-Living Materials**: Continue to bridge the gap between biocomposite and hybrid-living materials research by exploring the integration of living organisms with synthetic materials, enabling the creation of novel, sustainable, and functional materials.

6. **Complex Geometry and Topology**: Investigate the impact of complex topologies and geometries on the mechanical and surface properties of the myco-composite, and explore ways to design and fabricate these complex structures using the novel platform.

7. **Inoculation Strategies**: Investigate different inoculation strategies to optimize mycelium colonization, including the use of different mycelium strains, inoculation methods, and nutrient provision techniques, to further enhance the mechanical and surface properties of the myco-composite.

8. **Biocomposite Additive Manufacturing**: Develop and optimize new biocomposite additive manufacturing techniques to improve the resolution, speed, and scalability of the manufacturing process, enabling the creation of more complex and intricate structures.

\end{lstlisting}
\end{LLMboxmultipage}

\clearpage

\textboxcaption{Example of integrated thinking and reflection phases, where the model uses both thinking and reflection tokens. }
\label{TexBox:thinking_and_reflection_Box_14}

%(*@\bf\hlblue{System:}@*) Your are a helpful and friendly assistant. You are an expert in materials science with special knowledge in biological materials. You provide concise but accurate responses. You are an expert in combining different concepts, often not obviously related, and provide accurate accounts of correct relationships or analogies. 

\begin{LLMboxmultipage}{Integrated thinking and reflection model.}
\begin{lstlisting}

(*@\bf\hlred{User:}@*) Tell me why hierarchical structures work so well. Use <|thinking|>.

(*@\bf\hllightgreen{Assistant:}@*) <|thinking|>
**Reasoning Steps**:

1. **Mechanical Properties**: Hierarchical structures have unique mechanical properties at different length scales, ranging from nanoscale to macroscopic scale.
2. **Material Organization**: These properties are a result of organized changes in the composition and structure of materials at different length scales.
3. **Anisotropic Nature**: The mechanical properties vary significantly with direction, leading to anisotropic behavior, which is a common characteristic of hierarchical structures.
4. **Functional Adaptation**: This anisotropic nature allows for the adaptation of materials to specific functional requirements in different directions, enabling them to perform multiple functions efficiently.

**Relevant Materials or Concepts**:

Hierarchical Structures, Mechanical Properties, Anisotropic Nature, Material Organization, Functional Adaptation.

**Design Principles**:

Design Principles:

1. **Hierarchical Organization**: Design materials and structures with organized changes at different length scales to create unique mechanical properties.
2. **Anisotropic Nature**: Incorporate anisotropic behavior to enable adaptation to different functional requirements in various directions.
3. **Mechanical Properties**: Utilize the unique mechanical properties at different length scales to improve the overall performance of the design.

**Material Properties or Features**:

Unique mechanical properties at different length scales, anisotropic behavior, organized changes in composition and structure, functional adaptation in different directions.

**Hypothesis**:

"Hierarchical structures work well due to their unique mechanical properties at different length scales, anisotropic nature, organized changes in composition and structure, and functional adaptation in various directions."

**Additional Background**:

Hierarchical structures work well due to the hierarchical organization of materials at different length scales. This organization allows for the creation of complex structures composed of multiple length scales, each with its own structure, composition, and properties. The combination of these scales, often with distinct materials or arrangements at different length scales, can lead to unique mechanical properties and behaviors. For example, in a hierarchical structure, the microstructure might be composed of a solid material with a nanostructure or vice versa. This combination of scales allows for the creation of complex structures that can exhibit unique mechanical properties, such as improved strength, toughness, or flexibility, compared to single-level structures. Additionally, the hierarchical organization of materials at different length scales can enable the creation of structures with complex, interconnected features that are not present in single-level structures. This is a result of the hierarchical organization of materials, which allows for the creation of structures with a high degree of complexity and interconnectivity, as seen in natural materials like bones, woods, and bones.

<|/thinking|>
<|reflect|>
**Improvements/Corrections**:

1. **Clarify the Role of Hierarchical Organization**: Emphasize that hierarchical structures work well due to the organized changes in material properties and structure at different length scales, which leads to unique mechanical properties and behaviors.
2. **Include Interplay of Factors**: Acknowledge that hierarchical structures often result from a combination of factors, such as structure, composition, and architecture, rather than a single dominant factor.
3. **Consider the Cost of Hierarchical Structures**: While hierarchical structures can offer unique benefits, they can also be more complex and expensive to design, manufacture, and maintain, especially at the molecular or nano level.
4. **Address Anisotropic Nature**: Clarify that the anisotropic nature of hierarchical structures is a result of the organized changes in material properties and structure, rather than a sole design principle.
<|/reflect|>

Hierarchical structures work well due to the unique mechanical properties they exhibit at different length scales. These structures, such as those composed of nanoscale to macroscopic scales, can be composed of a solid material with a nanostructure or vice versa. This combination of scales allows for the creation of complex structures that can exhibit unique mechanical properties and behaviors, such as improved strength, toughness, and flexibility. The anisotropic nature of these structures also contributes to their ability to adapt to specific functional requirements in different directions, enabling them to perform multiple functions efficiently. This is a result of the organized changes in the composition and structure of the materials at different length scales, leading to a high degree of adaptability and functionality.

\end{lstlisting}
\end{LLMboxmultipage}

\clearpage

\textboxcaption{Iterative improvements of responses, for $N=3$, obtained using a recursive algorithm using a model that produces both thinking and reflection tokens (see, Figure~\ref{fig:PreFlexOR_recursive} for an overview of the algorithm). A detailed analysis is shown in Table~\ref{tab:biological-materials-comparison} and Figure~\ref{fig:Fig_11}.}
\label{TexBox:thinking_and_reflection_Box_30}

% Start overall box around the entire process
\begin{OverallBox}{Iterative improvements of responses.}
\linespread{0.9}\selectfont % Adjust the line spread to reduce line

\begin{LLMboxSmall}{}
\begin{lstlisting}
(*@\bf\hlred{User:}@*) How do biological materials fail gracefully? Brief answer. Use <|thinking|>.
\end{lstlisting}

\end{LLMboxSmall}

\begin{LLMboxSmall}{Response Iteration 0}
\begin{lstlisting}
(*@\bf\hllightgreen{Assistant:}@*) Biological materials fail gracefully through a combination of hierarchical structures and periodic hierarchies that allow for energy dissipation and redistribution at multiple scales, preventing catastrophic failure.
\end{lstlisting}

\end{LLMboxSmall}

\downarrowbetweenboxes % Arrow between first and second iterations

\begin{LLMboxSmall}{Response Iteration 1}
\begin{lstlisting}
(*@\bf\hllightgreen{Assistant:}@*) Biological materials fail gracefully through a combination of hierarchical structures and periodic hierarchies that allow for energy dissipation and redistribution at multiple scales, preventing catastrophic failure. This is achieved by:

1. **Hierarchical Structures**: Biological materials are composed of multiple levels of organization, with each level contributing to the overall mechanical properties. This hierarchy allows for energy dissipation and toughening at different scales.

2. **Periodic Hierarchies**: The material properties and structures in biological materials are often periodic, with repeating patterns at different scales. This periodicity can enhance energy dissipation and toughening, preventing sudden failure.

3. **Energy Dissipation Mechanisms**: Biological materials often contain mechanisms that allow for energy dissipation and toughening, such as:

   - **Brittle Fracture**: While brittle materials can fracture suddenly, biological materials with helical fibers or sacrificial bonds can undergo brittle fracture, which can help dissipate energy and prevent catastrophic failure.

   - **Helical Fibers**: The presence of helical fibers in materials like bone or sea urchin spines can enhance energy dissipation and toughening, particularly under dynamic loading.

   - **Sacrificial Bonds**: Sacrificial bonds, such as the bonds between mineral bridges in nacre, can also contribute to energy dissipation and toughening, helping to prevent sudden failure.

These energy dissipation mechanisms can operate at different scales, from the molecular to the macroscopic level, further contributing to the toughening of biological materials and their ability to fail gracefully.
\end{lstlisting}
\end{LLMboxSmall}

\downarrowbetweenboxes % Arrow between first and second iterations

\begin{LLMboxSmall}{Response Iteration 2}
\begin{lstlisting}
(*@\bf\hllightgreen{Assistant:}@*) Biological materials fail gracefully through a combination of hierarchical structures and periodic hierarchies that allow for energy dissipation and redistribution at multiple scales. This includes:

1. **Hierarchical Structures**: Biological materials are composed of multiple levels of organization, with each level contributing to the overall mechanical properties. This hierarchy allows for energy dissipation and toughening at different scales.

2. **Periodic Hierarchies**: Many biological materials, such as animal shells, exhibit periodic hierarchies that allow for energy dissipation and toughening. These periodic hierarchies can be represented by a scale of operation ranging from 10 nm to 1 mm, which is typical for biological materials. This scale allows for the presence of features that enhance energy dissipation and toughness, such as helical fibers, sacrificial bonds, and sacrificial mineral bridges.

3. **Mechanical Properties**: Biological materials often exhibit mechanical properties that contribute to their toughness and energy dissipation. For example, nacre, a type of material found in seashells, exhibits a high strength-to-weight ratio due to its hierarchical structure and mechanical properties, such as helical fibers and sacrificial mineral bridges. These features allow for energy dissipation and toughening, preventing catastrophic failure under sudden and impact loading.

4. **Loading Conditions**: The effects of hierarchical structures and periodic hierarchies on energy dissipation and failure strength can vary depending on the loading conditions. For instance, under impact loading, the presence of hierarchical structures and periodic hierarchies can help to dissipate energy and prevent sudden failure. However, under tensile or compressive loading, the effects of these structures and hierarchies may be less pronounced due to the lower loading conditions.

In summary, the hierarchical structures and periodic hierarchies in biological materials contribute to their energy dissipation and toughness by allowing for the presence of features such as helical fibers, sacrificial bonds, and sacrificial mineral bridges at different scales. These features enhance the mechanical properties of biological materials, preventing catastrophic failure under sudden and impact loading, and ensuring a more gradual transfer of properties.
\end{lstlisting}
\end{LLMboxSmall}
\end{OverallBox}

\clearpage

\textboxcaption{Response of the non-fine tuned base model, \url{meta-llama/Llama-3.2-3B-Instruct}.}
\label{TexBox:thinking_and_reflection_Box_40_refmodel}

\begin{LLMboxmultipage}{Response of the non-fine tuned base model.}
\begin{lstlisting}
(*@\bf\hlred{User:}@*) Tell me why hierarchical structures work so well. 

(*@\bf\hllightgreen{Assistant:}@*) Hierarchical structures have been widely adopted in various fields, including business, government, and social organizations, due to their effectiveness in achieving goals and promoting efficiency. Here are some reasons why hierarchical structures work well:

1. **Clear Lines of Authority**: Hierarchical structures establish a clear chain of command, which helps to prevent confusion, overlapping work, and conflicting decisions. Each level of the hierarchy has a defined role and responsibility, ensuring that tasks are delegated and completed efficiently.
2. **Specialization and Expertise**: Hierarchies allow for specialization and expertise to be concentrated at different levels. This enables individuals to focus on specific tasks and develop deep knowledge in their areas of expertise, leading to improved performance and productivity.
3. **Efficient Communication**: Hierarchical structures facilitate communication by establishing a clear path for information to flow from top to bottom. This helps to prevent misunderstandings, ensures that decisions are made with input from relevant stakeholders, and enables swift action to be taken.
4. **Scalability**: Hierarchies can be easily scaled up or down to accommodate changing organizational needs. As the organization grows, new levels can be added, and existing ones can be modified to accommodate increased complexity.
5. **Accountability**: Hierarchical structures promote accountability by establishing clear lines of responsibility and consequences for non-compliance. This encourages individuals to take ownership of their work and strive for excellence.
6. **Decision-Making**: Hierarchies enable efficient decision-making by allowing for input from various stakeholders at different levels. This helps to ensure that decisions are informed, well-researched, and aligned with organizational goals.
7. **Motivation and Incentives**: Hierarchical structures can be designed to motivate employees by offering clear career paths, promotions, and incentives. This helps to attract and retain top talent, improve job satisfaction, and increase productivity.
8. **Flexibility**: While hierarchical structures can be rigid, they can also be flexible and adaptable to changing circumstances. Organizations can adjust their hierarchies to respond to new challenges, opportunities, or market conditions.
9. **Economies of Scale**: Hierarchies can take advantage of economies of scale by leveraging the expertise and resources of higher-level employees to support lower-level employees. This can lead to improved efficiency, reduced costs, and increased competitiveness.
10. **Cultural Fit**: Hierarchical structures can be designed to fit the culture and values of an organization. By aligning the hierarchy with the organization's culture, leaders can create a cohesive and effective team that is better equipped to achieve its goals.

While hierarchical structures have many advantages, they are not without limitations. Some potential drawbacks include:

* Inflexibility
* Bureaucratic red tape
* Limited opportunities for innovation and creativity
* Potential for power imbalances and corruption

To mitigate these limitations, organizations can adopt hybrid or flat organizational structures, which combine elements of hierarchy with more flexible and collaborative approaches. Ultimately, the effectiveness of a hierarchical structure depends on its design, implementation, and ongoing evaluation.
\end{lstlisting}
\end{LLMboxmultipage}

\end{document}